\documentclass{article}

\PassOptionsToPackage{numbers, compress}{natbib}
\usepackage[final]{neurips_2025}

\usepackage[utf8]{inputenc} 
\usepackage[T1]{fontenc}    
\usepackage{url}            
\usepackage{hyperref}
\usepackage{booktabs}       
\usepackage{amsfonts}       
\usepackage{nicefrac}       
\usepackage{microtype}      
\usepackage{xcolor}         

\usepackage{amsmath}
\usepackage{amssymb}
\usepackage{booktabs}
\usepackage{multirow}
\usepackage{multicol}
\usepackage{xcolor}
\usepackage{soul}

\usepackage{transparent}

\newcommand{\mathcolorbox}[2]{\colorbox{#1}{$\displaystyle #2$}}

\usepackage{algorithm}
\usepackage[noEnd,commentColor=black]{algpseudocodex}

\usepackage{amsmath}
\usepackage{amssymb}
\usepackage{mathtools}
\usepackage{amsthm}
\usepackage{cancel}
\usepackage{scalerel}

\usepackage[table]{xcolor}

\usepackage[capitalize,noabbrev]{cleveref}

\theoremstyle{plain}

\newtheorem{theorem}{Theorem}[section]

\newtheorem{proposition}[theorem]{Proposition}

\usepackage[textsize=tiny]{todonotes}

\usepackage[utf8]{inputenc} 
\usepackage[T1]{fontenc}    
\usepackage{url}            
\usepackage{booktabs}       
\usepackage{times}
\usepackage{xcolor} 
\usepackage{graphicx}
\usepackage{blindtext}
\usepackage[labelformat=empty, position=top]{subcaption}
\usepackage[export]{adjustbox}
\usepackage{wrapfig}
\usepackage{pifont}
\definecolor{mydarkblue}{rgb}{0,0.08,0.45}
\hypersetup{ %
    colorlinks=true,
    linkcolor=mydarkblue,
    citecolor=mydarkblue,
    filecolor=mydarkblue,
    urlcolor=mydarkblue,
}

\usepackage[shortlabels]{enumitem}

\usepackage{tikz}
\usetikzlibrary{decorations.pathreplacing,calc}

\usepackage{amsmath,amsfonts,bm}

\def\eqref#1{equation~\ref{#1}}

\def\1{\bm{1}}

\DeclareMathAlphabet{\mathsfit}{\encodingdefault}{\sfdefault}{m}{sl}
\SetMathAlphabet{\mathsfit}{bold}{\encodingdefault}{\sfdefault}{bx}{n}

\definecolor{darkblue}{rgb}{0,0.08,0.45}
\definecolor{cred}{HTML}{D62728}
\definecolor{cblue}{HTML}{1F77B4}
\definecolor{cgreen}{HTML}{79AB76}
\definecolor{cgrey}{rgb}{0.6,0.6,0.6}

\newcommand{\dd}[1]{\mathrm{d}#1}

\newcommand{\ourslong}[0]{Q-Chunking}
\newcommand{\hourpurple}[1]{\textbf{\color{ourpurple}{#1}}}
\newcommand{\hourblue}[1]{\textbf{\color{ourblue}{#1}}}
\newcommand{\hourmiddle}[1]{\textbf{\color{ourmiddle}{#1}}}

\newcommand{\rrc}[2]{\mathbf{r}_{#1:#2}}
\newcommand{\ac}[2]{\mathbf{a}_{#1:#2}}
\newcommand{\ssc}[2]{\mathbf{s}_{#1:#2}}

\renewcommand{\mathbf}{\boldsymbol}

\makeatletter

\title{Reinforcement Learning with Action Chunking}

\author{%
  Qiyang Li, Zhiyuan Zhou, Sergey Levine \\ UC Berkeley \\ \texttt{\{qcli,zhiyuan\_zhou,svlevine\}@eecs.berkeley.edu} 
}
\begin{document}

\definecolor{ourpurple}{HTML}{CB297B}
\definecolor{ourlightpurple}{HTML}{FDF7FA}
\definecolor{ourblue}{HTML}{0076BA}
\definecolor{ourlightblue}{HTML}{F5FAFE}
\definecolor{ourmiddle}{HTML}{66509B}
\definecolor{ourlightmiddle}{HTML}{F7F6F9}
\hypersetup{colorlinks=true, allcolors=ourblue}

\maketitle
\begin{abstract}
We present \hourpurple{Q-chunking}, a simple yet effective recipe for improving reinforcement learning (RL) algorithms for long-horizon, sparse-reward tasks. Our recipe is designed for the offline-to-online RL setting, where the goal is to leverage an offline prior dataset to maximize the sample-efficiency of online learning. Effective exploration and sample-efficient learning remain central challenges in this setting, as it is not obvious how the offline data should be utilized to acquire a good exploratory policy. 
Our key insight is that action chunking, a technique popularized in imitation learning where sequences of future actions are predicted rather than a single action at each timestep, can be applied to temporal difference (TD)-based RL methods to mitigate the exploration challenge.
Q-chunking adopts action chunking by directly running RL in a `chunked' action space, enabling the agent to (1) leverage temporally consistent behaviors from offline data for more effective online exploration and (2) use unbiased $n$-step backups for more stable and efficient TD learning. Our experimental results demonstrate that Q-chunking exhibits strong offline performance and online sample efficiency, outperforming prior best offline-to-online methods on a range of long-horizon, sparse-reward manipulation tasks. 
\begin{figure}[H]

    \centering
    \includegraphics[width=0.50\linewidth]{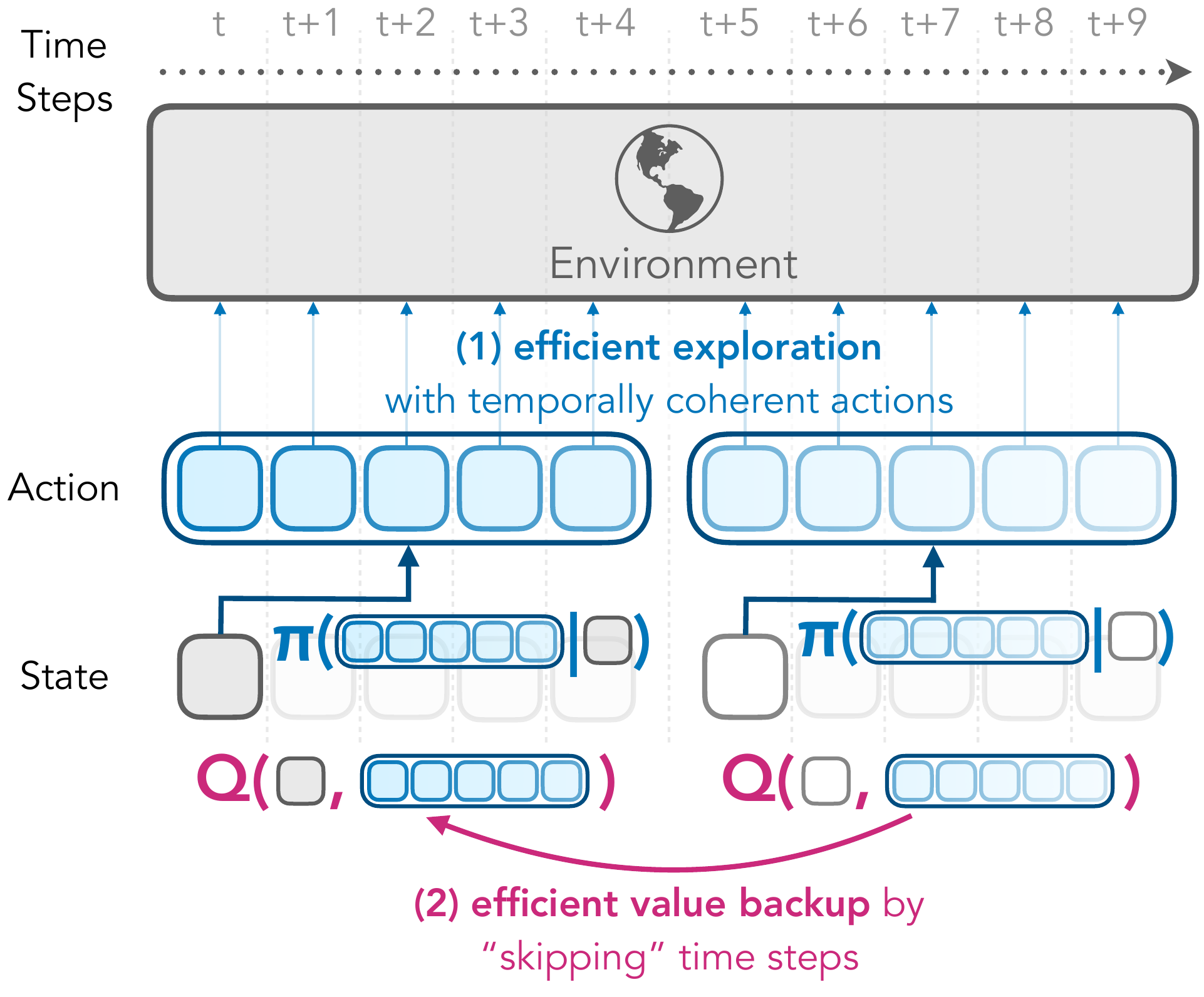}    
    \hfill
    \includegraphics[width=0.48\linewidth]{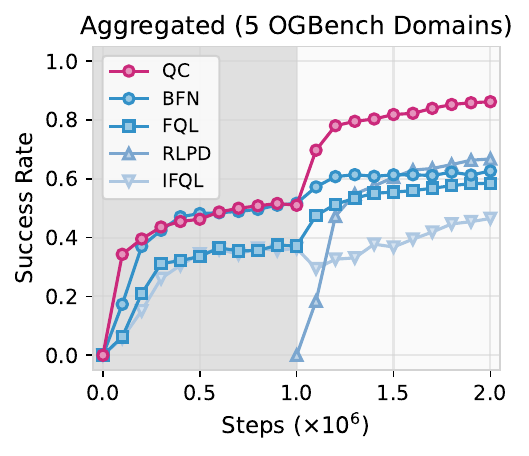}  
    \caption{\footnotesize \textbf{\hourpurple{Q-chunking} uses action chunking to enable fast value backups and effective exploration with temporally coherent actions.} \emph{left:} an overview of our approach: Q-chunking operates in a temporally extended action space that allows for (1) efficient value backups and (2) effective exploration via temporally coherent actions; \emph{right:} Our method (\hourpurple{QC}) first pre-trains on an offline dataset for 1M steps (grey) and then updates with online data for another 1M steps (white). Our method achieves strong aggregated performance over five challenging long-horizon sparse-reward domains in OGBench. \hourpurple{Code:} \href{http://github.com/ColinQiyangLi/qc}{\hourpurple{\texttt{github.com/ColinQiyangLi/qc}}}}.
    \label{fig:banner}
\end{figure}
\end{abstract}

\section{Introduction}
\label{sec:intro}
Reinforcement learning (RL) holds the promise of solving any given task based only on a reward function. However, this simple and direct formulation of the RL problem is often impractical: in complex environments, exploring entirely from scratch to learn an effective policy can be prohibitively expensive, as it requires the agent to successfully solve the task through random chance before learning a good policy. Indeed, even humans and animals rarely solve new tasks entirely from scratch, instead leveraging prior knowledge and skills from past experience. Inspired by this, a number of recent works have sought to incorporate prior offline data into online RL exploration~\citep{hu2023unsupervised, li2024accelerating, wilcoxson2024leveraging}.
But this poses a new set of challenges: the distribution of offline data might not match the policy that the agent should follow online, introducing distributional shift, and it is not obvious how the offline data should be leveraged to acquire a good online \emph{exploratory} policy.

In the adjacent field of imitation learning (IL), a widely used approach in recent years has been to employ \emph{action chunking}, where instead of training policies to predict a single action based on the state observation from prior data, the policy is instead trained to predict a short sequence of future actions (an ``action chunk'')~\citep{zhao2023learning, chi2023diffusion}. While a complete explanation for the effectiveness of action chunking in IL remains an open question, its effectiveness can be at least partially ascribed to better handling of non-Markovian behavior in the offline data, essentially providing a more powerful tool for modeling the kinds of complex distributions that might occur in (for example) human-provided demonstrations or mixtures of different behaviors~\citep{zhao2023learning}. Action chunking has not been used widely in RL, perhaps because optimal policies in fully observed MDPs are Markovian~\citep{sutton1998reinforcement}, and therefore chunking may appear unnecessary. 

We make the observation that, though we might desire a final optimal Markovian policy, the exploration problem can be better tackled with non-Markovian and temporally extended skills, and that action chunking offers a very simple and convenient recipe for obtaining this. Furthermore, action chunking provides a better way to leverage offline data (with a better handling of non-Markovian behavior in the data), and even improves the stability and efficiency of TD-based RL, by enabling unbiased $n$-step backups (where $n$ matches the length of the chunk). Thus, in combination with pretraining on offline data, action chunking offers a compelling and very simple way to mitigate the exploration challenge in RL. 

We present Q-learning with action chunking (or \hourpurple{Q-chunking} in short), a recipe for improving generic TD-based actor-critic RL algorithms in the offline-to-online RL setting (\cref{fig:banner}). The key idea is to run RL at an action sequence level --- (1) the policy predicts a sequence of actions for the next $h$ steps and executes them one-by-one open loop, and (2) the critic takes in the current state and a sequence of actions and estimates the value of carrying out the whole sequence rather than a single action. The benefits of operating RL on this extended action space are two-fold: (1) the policy can be optimized to generate temporally coherent actions by regularizing it towards some prior behavior data that exhibit such coherency, (2) the critic trained with a standard TD-backup loss is effectively performing $n$-step backups, with no off-policy bias (that typically occurs in na\"ive $n$-step return methods), since the critic takes the full action sequence into account. 

Our main contribution is \hourpurple{QC}, a practical offline-to-online RL algorithm that is instantiated from our \hourpurple{Q-chunking} recipe, essentially running RL with behavior regularization in the chunked action space. QC is simple to implement, requiring only training (1) an action chunking behavior policy using a standard flow-matching loss, and (2) a temporally extended critic with the standard TD-loss (\cref{algo:qc}). QC achieves strong performance on a range of six challenging long-horizon, sparse-reward domains, outperforming prior offline-to-online methods. Moreover, Q-chunking is a generic recipe that can be applied to existing offline-to-online algorithms with minimal modification. In this work, we demonstrate one such instantiation by applying it to \hourblue{FQL}~\citep{park2025flow}, resulting in \hourpurple{QC-FQL} (\cref{algo:qc-fql}), which shows significant improvements over the original method.

\section{Related Work}
\label{sec:related}

\textbf{Offline-to-online reinforcement learning} methods focus on leveraging prior offline data to accelerate reinforcement learning online~\citep{xie2021policy, song2022hybrid, lee2022offline, NEURIPS2022_ba1c5356, zhang2023policy, zheng2023adaptive, ball2023efficient, nakamoto2024cal, zhou2024efficient, li2024accelerating}. The simplest way to tackle offline-to-online RL is to use an existing offline RL algorithm to first pretrain on the offline data and then use the same offline optimization objective to continue training online using a growing dataset that combines the original offline data and the replay buffer data~\citep{nair2020awac, kumar2020conservative, kostrikov2021offline, tarasov2024revisiting, park2025flow, agarwal2022reincarnating, luo2023finetuning, lee2022offline}. While straightforward, this na\"ive approach often result in overly pessimistic that hinders exploration and consequently the online sample-efficiency. Several prior works have attempted to address this issue by adjusting the degree of pessimism online~\citep{zhou2024efficient, nakamoto2024cal, luo2023finetuning, lee2022offline, wang2023train}. 
However, these approaches can be difficult to tune and sometimes stills fall short in online sample efficiency compared to a simple, well-regularized online RL algorithm learning from scratch on both offline data and online replay buffer data~\citep{ball2023efficient}.
Our approach takes a step towards improving the sample efficiency of offline-to-online RL methods via value backup acceleration and temporally coherent exploration.

\textbf{Action chunking} is a technique popularized by roboticists for imitation learning (IL), where the policy predicts and executes a sequence of actions in an open-loop manner (``an action chunk'')~\citep{zhao2023learning}. Action chunking has been shown to improve policy robustness~\citep{zhao2023learning, george2023one, bharadhwaj2024roboagent}, and handle non-Markovian behavior in offline data~\citep{zhao2023learning}. 
Existing RL methods that incorporate action chunking typically focus on fine-tuning a policy pre-trained with imitation learning~\citep{ren2024diffusion}. \citet{tian2025chunking} propose to learn a critic on action chunks by integrating $n$-step returns with a transformer. However, their method only applies chunking to the critic, while still optimizing a single-step actor. 
\citet{li2024top} also observe that learning a critic over short action chunks removes the off-policy bias in $n$-step return backups, leading to more stable and effective value learning. There are key differences between their work and ours --- \citet{li2024top} operate in the online episodic RL setting~\citep{whitley1993genetic, kober2008policy} and use Gaussian policies to predict parameters of movement primitives (MP)~\citep{schaal2006dynamic, paraschos2013probabilistic}, which are then used to generate full action sequences at the beginning of each episode. In contrast, we operate in the conventional offline-to-online RL setting and leverage more expressive flow-matching-based policies (as we find Gaussian policies to be ineffective as shown in Figure \ref{fig:rlpd-all}) to predict short action sequences directly in the raw action space. 
\citet{seo2024reinforcement} also train critics on action chunks and impose a behavior cloning loss that aligns with the principles behind Q-chunking. The key distinction from our work lies in their use of a multi-level, factorized critic architecture~\citep{seo2024continuous}, which generates and refines action chunks from coarse to fine granularity via iterative discretization. At each level, the action space is discretized into bins, and the Q-function is modeled independently for each action dimension and timestep, conditioned on the whole action chunks predicted at the previous coarser level. While this factorized critic design enables tractable value-maximizing action sampling, it imposes strong structural assumptions on the action space, limiting policy expressiveness at each refinement level. In contrast, we make no such assumptions in our recipe which allows us to derive two general algorithms where both the critic and policy operate directly on action chunks without requiring factorization or iterative discretization/refinement.

\textbf{Exploration with temporally coherent actions.} 
Existing methods either rely on temporally correlated action noises~\citep{lillicrap2015continuous} that are constructed through heuristics; hierarchically structured policies (see the next paragraph), which are often tricky to stabilize during online training; or pre-trained frozen skill policies~\citep{pertsch2021accelerating, wilcoxson2024leveraging}, which are not amendable for fine-grained online fine-tuning. Our method uses a single network to represent the policy to generate temporally extended action chunk and it is trained using a single objective function that is stable to optimize. There is also no frozen, pretrained components in our approach, ensuring its online fine-tuning flexibility.

\textbf{Hierarchical reinforcement learning, options framework.} Learning temporally extended actions have also been widely studied in the hierarchical reinforcement learning (HRL) literature~\citep{dayan1992feudal, dietterich2000hierarchical, vezhnevets2016strategic, daniel2016hierarchical, kulkarni2016hierarchical, vezhnevets2017feudal, peng2017deeploco, riedmiller2018learning, nachum2018data, ajay2020opal, shankar2020learning, pertsch2021accelerating, gehring2021hierarchical, xie2021latent}. HRL methods typically train a space of low-level policies that can directly interact with the environment along with a high-level policy that selects among these low-level policies. These low-level policies can be hand-crafted~\citep{dalal2021accelerating}, automatically discovered online~\citep{dietterich2000hierarchical, kulkarni2016hierarchical, vezhnevets2016strategic, vezhnevets2017feudal, nachum2018data}, or pretrained using offline skill discovery methods~\citep{paraschos2013probabilistic,merel2018neural,shankar2020learning, ajay2020opal, singh2020parrot, pertsch2021accelerating, touati2022does, nasiriany2022learning, hu2023unsupervised, frans2024unsupervised, chen2024self, park2024foundation}. The options framework provides a slightly more sophisticated and more powerful formulation, where the low-level policy is additionally associated with learnable initiation condition and termination condition that makes utilization of the low-level policy more flexible~\citep{sutton1999between, menache2002q, chentanez2004intrinsically, mannor2004dynamic, csimcsek2004using, csimcsek2007betweenness, konidaris2011autonomous, daniel2016hierarchical, srinivas2016option, oh2017value, fox2017multi, bacon2017option, kim2019variational, bagaria2019option, bagaria2024effectively, de2025learning}. 
A long-lasting challenge in HRL is its bi-level optimization problem: when both low-level and high-level policies are updated during training, the high-level policies must optimize a moving objective function, which can lead to instability~\citep{nachum2018data}. To mitigate this, some methods keep the low-level policies frozen after initial pretraining~\citep{ajay2020opal, pertsch2021accelerating, wilcoxson2024leveraging} to improve stability during online training. Our approach is a special case of HRL where the low-level skill executes a sequence of actions open-loop. This design choice allows us to collapse the bi-level optimization problem into a standard RL objective in a temporally extended action space, while retaining many of the exploration benefits associated with HRL methods. While prior work in HRL has also explored such idea~\citep{mcgovern1998macro, durugkar2016deep}, they often leverage a discrete set of action sequences that is either heuristically extracted from the prior experience or given in advance. In contrast, we use an expressive flow-matching based policy to directly parameterize a continuous space of action sequences, and are directly trained and fine-tuned online with RL.

\textbf{Multi-step latent space planning and search} is a technique commonly used in model-based RL methods where they use a learned model to optimize a short-horizon action sequence towards high-return trajectories~\citep{oh2017value, schrittwieser2020mastering}. 
These approaches work by training a dynamics model on an encoded latent space, where the model takes in a latent state and an action to predict the next latent state and the associated reward value. This latent dynamics model, along with a value network on the latent state, can then provide an estimate of the $Q$-value on-the-fly for any given action sequence starting from a given latent state by simply simulating the action sequence in the latent dynamics model. In contrast, we do not learn a latent dynamics model and instead train a $Q$-network to directly estimate the value of the action sequence.
Lastly, these approaches operate in the purely online RL setting whereas we focus on the offline-to-online RL setting.

\section{Background}
\label{sec:preliminary}
\textbf{Offline-to-online RL.} In this paper, we consider an infinite-horizon, fully observable Markov decision process (MDP), $(\mathcal{S}, \mathcal{A}, \rho, T, r, \gamma)$, where $\mathcal{S}$ is the state space, $\mathcal{A}$ is the action space, $T(s'|s, a): \mathcal{S} \times \mathcal{A} \mapsto \Delta(\mathcal{S})$ is the transition kernel, $r(s, a): \mathcal{S}\times \mathcal{A}\mapsto \mathbb{R}$ is the reward function, $\rho: \Delta(\mathcal{S})$ is the initial state distribution and $\gamma \in [0, 1)$ is the discount factor. We also assume there is a prior offline dataset $\mathcal{D}$ that consists of transitions rollouts $\{(s, a, s', r)\}$ from $\mathcal{M}$. The goal of offline-to-online RL is to find a policy $\pi(a \mid s): \mathcal{S} \mapsto \Delta(\mathcal{A})$ that maximizes the expected discounted cumulative reward (or discounted return): $\eta(\pi) := \mathbb{E}_{s_{t+1} \sim T(s_t, a_t), a_t \sim \pi(\cdot \mid s_{t})}\left[\sum_{t=0}^{\infty} \gamma^t r(s_t, a_t)\right]$.
Oftentimes, offline-to-online RL algorithms operate in two distinct phases: an offline phase where a policy is pretrained on the offline data $\mathcal{D}$ and an online phase where the policy is further fine-tuned online with environment interactions. Our approach follows the same regime.

\textbf{Temporal difference and multi-step return.} TD-based RL algorithms typically learn $Q_\theta(s, a)$ to approximate the maximum expected discounted cumulative reward that a policy can receive starting from state $s$ and action $a$ by using a temporal difference (TD) loss~\citep{sutton1998reinforcement}:
\begin{align}
    L(\theta) = \left[Q_\theta(s_t, a_t) - \hat{V}\right]^2,
\end{align}
where $\hat{V}$ is an estimate of $Q(s_t, a_t)$ that is commonly chosen as $\hat{V}_{\mathrm{1\text{-}step}}$:
\begin{align}
    \hat{V}_{\mathrm{1\text{-}step}} := r_t + \gamma Q_{\bar \theta}(s_{t+1}, a_{t+1}), \quad a_{t+1} \sim \pi_\psi(\cdot| s_{t+1}),
\end{align}
and $s_t, a_t, s_{t+1}, r$ are sampled from some off-policy trajectories and $\bar \theta$ is a delayed version of $\theta$ that does not allow the gradient to pass through for learning stability. When the TD error is minimized, the $Q_\theta$ converges to the expected discounted value of the policy $\pi_\psi$. As the effective horizon $\tilde{H} = 1 / (1-\gamma)$ goes up, the learning slows down as the value only propagates 1 step backward (from $s_{t+1}$ to $s_t$). To speed-up long-horizon value backup, a common strategy is to sample a length-$n$ trajectory segment, ($s_t, a_t, s_{t+1}, \cdots, a_{t+n-1}, s_{t+n})$, and construct a $n$-step return from it~\citep{watkins1989learning, sutton1998reinforcement}:
\begin{align}
    \hat{V}_{\mathrm{n\text{-}step}} := \sum_{t'=t}^{t+n-1} \left[\gamma^{t'-t}r_{t'}\right] + Q_{\bar \theta}(s_{t+n}, a_{t+n}), \quad a_{t+n} \sim \pi_\psi(\cdot \mid s_{t+n}),
    \label{eq:nstep}
\end{align}
where again $r_t = r(s_t, a_t)$. This value estimate of $Q(s_t, a_t)$ allows for a $n$ times speed-up in terms of the number of time steps that the value can propagate back across. This estimator is sometimes referred to as the \emph{uncorrected $n$-step return estimator}~\citep{fedus2020revisiting, kozuno2021revisiting} because it is biased when the data collection policy is different from the current policy $\pi_\psi$. 
Nevertheless, due to the implementation simplicity of $n$-step return, it has been commonly adopted in large-scale RL systems~\citep{mnih2016asynchronous, hessel2018rainbow, kapturowski2018recurrent, wurman2022outracing}.

\section{\ourslong{}}
\label{sec:method}
In this section, we first describe two main design principles of \hourpurple{Q-chunking}: (1) Q-learning on a temporally extended action space (the space of chunks of actions), and (2) behavior constraint in this extended action space, followed by practical implementations of Q-chunking (\hourpurple{QC}, \hourpurple{QC-FQL}) as effective TD-based offline-to-online RL algorithms.

\subsection{Q-learning on a temporally extended action space} 
\label{sec:qc-v}
The first design principle of Q-chunking is to apply $Q$-learning on the temporally extended action space. Unlike normal 1-step TD-based actor-critic methods, which train a Q-function $Q(s_t, a_t)$ and a policy $\pi(a_t \mid s_t)$, we instead train both the critic and the actor with a span of $h$ consecutive actions:~\footnote{We use $\ac{t}{t+h}$ (or simply $\mathbf{a}_t$) to denote a concatenation of $h$ consecutive actions: $\begin{bmatrix} a_t & \cdots & a_{t+h-1} \end{bmatrix} \in \mathbb{R}^{Ah}$ for notation convenience. This is similar for $\ssc{t}{t+h}$ and $\rrc{t}{t+h}$.}
\begin{align*}
 &\text{\emph{Q-Chunking Policy: }} \pi_\psi({\color{ourpurple} \ac{t}{t+h}} \mid s_t) :=\pi_\psi( {\color{ourpurple} a_t, a_{t+1}, \cdots, a_{t+h-1}} \mid s_t) \\
 &\text{\emph{Q-Chunking Critic: }} Q_\theta(s_t, {\color{ourpurple} \ac{t}{t+h}}) := Q_\theta(s_t, {\color{ourpurple} a_t, a_{t+1}, \cdots, a_{t+h-1}})  
\end{align*}
In practice, this involves updating the critic and the actor on batches of transitions consisting of a random state $s_t$, an action sequence $\mathbf a_{t:t+h}$ followed by the state, and the state $h$ steps into the future, $s_{t+h}$. Specifically, we train $Q_\theta$ with the following TD loss,
\begin{align}
\label{eq:1}
    L(\theta) = \mathbb{E}_{s_t, {\color{ourpurple}\ac{t}{t+h}, s_{t+h}} \sim \mathcal{D}}
    \left[\left(Q_\theta(s_t, {\color{ourpurple}\ac{t}{t+h}}) - {\color{ourpurple}\sum_{t'=0}^{h-1} \gamma^{t'}r_{t+t'}} - 
    \gamma^{\color{ourpurple}h} Q_{\bar \theta}({\color{ourpurple}s_{t+h}},  {\color{ourpurple} \ac{t+h}{t+2h}})\right)^2\right]
\end{align}
with ${\color{ourpurple} \ac{t+h}{t+2h}} \sim \pi_\psi(\cdot \mid s_{t+h})$, and $\bar \theta$ being the target network parameters that are often an exponential moving average of $\theta$~\citep{haarnoja2018soft}.

The TD loss above shares striking similarity to the $n$-step return in Equation \ref{eq:nstep} (with $n$ matches $h$) but with a crucial difference --- the $Q$-function used in the $n$-step return backup takes in only one action (at time step $t$) whereas our $Q$-function takes in the whole action sequence. The implication of this difference can be best explained after we write out the TD backup equations for standard 1-step TD, $n$-step return, and Q-chunking:
\begin{align}
    Q(s_t, a_t) & \leftarrow r_t + \gamma Q(s_{t+1}, a_{t+1})  & \text{(standard 1-step TD)} \\
    Q(s_t, a_t) &\leftarrow \underbrace{\sum_{t'=t}^{t+h-1} \left[\gamma^{t'-t}r_{t'}\right]}_\text{\color{ourblue} biased} + \gamma^hQ(s_{t+h}, a_{t+h}), & \text{\color{ourblue}($n$-step return, $n=h$)} \\
    Q(s_t, {\color{ourpurple}\ac{t}{t+h}}) &\leftarrow \underbrace{\sum_{t'=t}^{t+h-1} \left[\gamma^{t'-t}r_{t'}\right]}_\text{\color{ourpurple} unbiased} + \gamma^h Q(s_{t+h}, {\color{ourpurple}\ac{t+h}{t+2h}}). & \text{\hourpurple{(Q-chunking)}}
\end{align}
For the standard 1-step TD, each backup step propagates the value back by only 1 time step. $n$-step return propagates the value back $h \times$ faster, but can suffer from a \emph{biased} value estimation issue when $\ssc{t}{t+h}$ and $\ac{t}{t+h}$ are off-policy~\citep{fedus2020revisiting}. This is because the discounted sum of the $n$-step rewards $\rrc{t}{t+h}$ from the dataset or replay buffer is no longer an unbiased estimate of the expected $n$-step rewards under the current policy $\pi$. Q-chunking value backup is similar to the $n$-step return where each step also propagates the value back by $h$ time steps, but \emph{does not} suffer from this biased estimation issue. Unlike $n$-step return where we are propagating the value to a 1-step $Q$-function, Q-chunking backup propagates the value back to a $h$-step $Q$-function that takes in the exact same actions that are taken to obtain the $n$-step rewards $\rrc{t}{t+h}$, eliminating the biased value estimation. We formalize this argument in \cref{thm:theory}. As a result, Q-chunking value backup enjoys the value propagation speedup while maintaining an unbiased value estimate.

\subsection{Behavior constraints for temporally coherent exploration}

\label{sec:qc-p}

The second design principle of Q-chunking addresses the action incoherency issues by leveraging a behavior constraint in the objective for the $\pi_\psi$:
\begin{align}
    L(\psi) = -\mathbb{E}_{\ac{t}{t+h} \sim \pi_\psi(\cdot \mid s_t)}\left[Q_\theta(s_t, \ac{t}{t+h})\right], \text{s.t. }  D(\pi_\psi(\ac{t}{t+h} \mid s_t), \pi_\beta(\ac{t}{t+h} \mid s_t)) \leq \varepsilon
    \label{eq:qc-p}
\end{align}
where we denote $\pi_\beta(\ac{t}{t+h} \mid s_t)$ as the behavior distribution in the offline data $\mathcal{D}$, and $D$ as some distance metric that measures how different the learned policy $\pi$ deviates from $\pi_\beta$. 

Intuitively, a behavior constraint on the temporally extended action sequence allows us to leverage temporally coherent action sequences in the offline dataset. This is particularly advantageous in the temporally extended action space compared to in the original action space because offline data often exhibit non-Markovian structure (\emph{e.g.}, from scripted policies~\citep{park2024ogbench}, human tele-operators~\citep{robomimic2021}, or noisy expert policies for sub-tasks~\citep{park2024ogbench, fu2020d4rl}) that cannot be well captured by a Markovian behavior constraint. 
Temporally coherent actions are desirable for online exploration because they resemble temporally extended skills (\emph{e.g.}, moving in a certain direction for navigation, jumping motions for going over obstacles) that help traverse the environment in a structured way rather than using random actions that often result in data that is localized near the initial states. 
Imposing behavior constraint for an action chunking policy is a very simple way to approximately extract skills without the need of training policy with bi-level structure as often necessitated by skill-based methods (see more discussion in Section \ref{sec:related}). We observe that Q-chunking, with such behavior constraints, can explore with temporally coherent actions (see Section \ref{sec:results-analysis}), mitigating the exploration challenge.

\begin{figure}[t]
\centering
\includegraphics[width=1.0\textwidth]{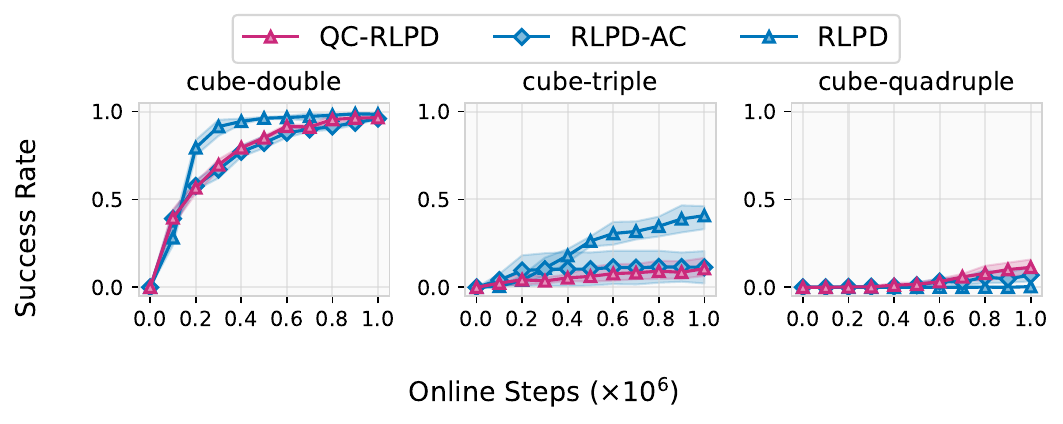}  %
\vspace{-0.7em}
\caption{\footnotesize \textbf{Na\"ively using action chunking for online RL with Gaussian policies leads to poor performance.} \emph{(1)} \hourblue{RLPD} runs online RL on both offline data and online replay buffer~\citep{ball2023efficient}. \emph{(2)} \hourblue{RLPD-AC} is the same algorithm as RLPD but operates in a temporally extended action space (action chunk size of 5). \emph{(3)} \hourpurple{QC-RLPD} additionally uses a behavior cloning loss on the actor (4 seeds). }
\vspace{-1em}
\label{fig:rlpd-all}
\end{figure}

\subsection{Practical implementations}
A key implementation challenge of Q-chunking is to enforce a good behavior constraint that captures the non-Markovian behavior at the action sequence level. One prerequisites of imposing a good behavior constraint is the ability of the policy to capture the complex behavior distribution (\emph{e.g.}, with a flow/diffusion policy). A Gaussian policy, a default choice in online RL algorithms, does not suffice. Indeed, if we na\"ively take an off-the-shelf online algorithm, \hourblue{RLPD}~\citep{ball2023efficient} for example, and apply Q-chunking with a behavior cloning loss, we find that it often performs poorly (\cref{fig:rlpd-all}). 

To enforce a good behavior constraint, we start by using flow-matching objective~\citep{liu2022flow} to train a behavior cloning flow policy to capture the behavior distribution. The flow policy is parameterized by a state-conditioned velocity field prediction model $f(s, \mathbf z, u): \mathcal{S} \times \mathbb{R}^{Ah} \times [0, 1] \mapsto \mathbb{R}^{Ah}$ and we denote $f_\xi(\cdot \mid s)$ as the action distribution that the flow policy parameterizes, which serves as an approximation of the true behavior distribution in the offline data ($f_\xi \approx \pi_\beta$). Now, we are ready to present our main method:

\hourpurple{QC}\textbf{: Q-chunking with implicit KL behavior constraint.}
We consider a KL constraint on our policy through the learned behavior distribution:
\begin{align}
    D_{\mathrm{KL}}(\pi_\psi \| f_\xi(\cdot \mid s)) \leq \varepsilon
    \label{eq:kl}
\end{align}
While it is possible include the KL as part of the loss, estimating the KL divergence or log probability for flow models is practically challenging. Instead, we use best-of-$N$ sampling~\citep{stiennon2020learning} to maximize $Q$-value while imposing this KL constraint implicitly altogether. Practically, this involves first sampling $N$ action chunks from the learned behavior policy $f_\xi(\cdot \mid s)$,
\begin{align*}
    \{\mathbf{a}^1, \mathbf{a}^2, \cdots, \mathbf{a}^N\} \sim f_\xi(\cdot \mid s),
\end{align*}
and then picking the action chunk sample that maximizes the temporally extended $Q$-function:
\begin{align*}
    \mathbf{a}^\star \leftarrow {\arg\max}_{\mathbf a \in \{\mathbf a^1, \mathbf a^2, \cdots, \mathbf a^N\}} Q(s, \mathbf{a})
\end{align*}
It has been shown in prior work that best-of-$N$ sampling admits a closed-form upper-bound on the KL divergence from the original distribution~\citep{hilton2023kl}:
\begin{align}
    D_{\mathrm{KL}}(\mathbf{a}^\star \| f_\xi(\cdot \mid s)) \leq \log N - \frac{N-1}{N},
\end{align}
which approximately satisfies KL constraint implicitly (Equation \ref{eq:kl}). Tuning the value of $N$ directly corresponds to the strength of the constraint.

Since we approximate the policy optimization (Equation \ref{eq:qc-p}) with the best-of-$N$ sampling, we can completely avoid separately parameterizing a policy $\pi_\psi$ and only sample from the behavioral policy $f_\xi(\cdot \mid s_t)$. In particular, we use the best-of-$N$ sampling to generate actions to both (1) interact with the environment, and (2) provide the action samples in the TD backup following \citet{ghasemipour2021emaq}. As a result, our algorithm has only one additional loss function:
\begin{align}
    L(\theta) = \mathbb{E}_{\substack{s_t, \mathbf{a}_t \sim D \\ \{\color{ourblue}{\mathbf{a}^{i}_{t+h}\}_{i=1}^{N} \sim f_\xi(\cdot|s_{t+h})}}} \left[\left(Q_\theta(s_t, \mathbf{a}_t) - \sum_{t'=0}^{h-1} \gamma^{t'}r_{t+t'} - \gamma^{h} Q_{\bar \theta}(s_{t+h}, {\color{ourpurple}\mathbf{a}^\star_{t+h}})\right)^2\right]
    \label{eq:qc-critic}
\end{align}
where again ${\color{ourpurple}\mathbf{a}^\star_{t+h} := {\arg\max}_{\mathbf a \in \{\mathbf a_{t+h}^{i}\}} Q(s, \mathbf{a})}$. 

While QC is simple and easy to implement, it does come with some additional computational costs associated with best-of-N sampling. In our experiments, we experiment with two other variants (\hourpurple{QC-FQL} and \hourpurple{QC-IFQL}) that leverage cheaper off-the-shelf offline/offline-to-online RL methods (FQL and IFQL respectively~\citep{park2025flow}). We present the FQL-version below as it performs better empirically. 

\hourpurple{QC-FQL}\textbf{: Q-chunking with 2-Wasserstein distance behavior constraint.} For this variant of our method, we leverage the optimal transport framework to impose a Wasserstein distance ($W_2$) constraint, again, through the learned behavior policy $f_\xi(\cdot \mid s)$:
\begin{align}
    W_2 (\pi_\psi, f_\xi(\cdot \mid s)) \leq \varepsilon
\end{align}
Following \hourblue{FQL}~\citep{park2025flow}, we parameterize the policy $\pi_\psi$ with a noise-conditioned action prediction model, $\mu_\psi(s, \mathbf z): \mathcal{S} \times \mathbb{R}^{Ah} \mapsto \mathbb{R}^{Ah}$, which directly outputs an action from Gaussian noise in one network forward pass. This noise-conditioned policy is trained to maximize the Q-chunking critic $Q_\theta(s_t, \ac{t}{t+h})$ while being regularized to be close to the behavioral cloning flow-matching policy via a BC loss that is shown to be an upper-bound on the squared 2-Wasserstein distance~\citep{park2025flow}:
\begin{align}
    L(\psi) &= \mathbb{E}_{s_t \sim \mathcal{D}, \mathbf z^0 \sim \mathcal{N}(0, \mathbf{I}_{Ah})}\left[{\color{ourblue}\alpha\left\|\mathbf z^1 - \mu_\psi(s_t, \mathbf z^0)\right\|_2^2} - Q(s_t, \mu_\psi(s_t, \mathbf z))\right] 
    \label{eq:fql-onestep}\\
    & \geq \mathbb{E}_{s_t \sim \mathcal{D}, \mathbf z^0 \sim \mathcal{N}(0, \mathbf{I}_{Ah})}\left[{\color{ourblue}\alpha W_2(\pi_\psi(\cdot \mid s_t), f_\xi(\cdot \mid s_t))^2} - Q(s_t, \mu_\psi(s_t, \mathbf z))\right],
\end{align}
where $\mathbf z^1$ is the ODE solution from $u=0$ to $u=1$ following $\dd \mathbf z^u = f_\xi(s_t, \mathbf z^u, u) \dd u$ (the initial value $\mathbf z^0$ is sampled from the unit Gaussian). The real-valued hyperparameter $\alpha$ directly controls the magnitude of the distillation loss. Finally, the TD loss remains the same as the previous section with the only difference in how we parameterize the policy:
\begin{align}
    L(\theta) = \mathbb{E}_{s_t, \mathbf a_t, s_{t+h} \sim \mathcal{D}, {\color{ourblue}\mathbf z}}\left[\left(Q_\theta(s_t, \mathbf a_t) - \sum_{t'=0}^{h-1}\gamma^{t'} r_{t+t'} - \gamma^h Q_{\bar \theta}(s_{t+h}, {\color{ourblue} \mu_\psi(s_{t+h}, \mathbf z)})\right)^2\right]
    \label{eq:fql-critic}
\end{align}
where again $\mathbf z \sim \mathcal{N}(0, \mathbf I_{Ah})$.

\textbf{Offline-to-online RL considerations.} Since both variants of our methods use behavior constraint (implicit KL for QC, explicit $W_2$ for QC-FQL), we can also directly run them for offline RL pre-training.
For both offline and online training, we use the same behavior constraint strength (\emph{e.g.}, $N$ for QC and $\alpha$ for QC-FQL). We use the same algorithm for both offline and online training, and the only difference is whether there is environment interactions.
See \cref{appendix:impl-details}, \cref{algo:qc} and \cref{algo:qc-fql} for an overview of QC and QC-FQL during online training. 

\definecolor{ddblue}{HTML}{4592d7}
\definecolor{lblue}{HTML}{cfe2f3}
\definecolor{lred}{rgb}{1.0, 0.9, 0.9}

\section{Experimental Results}
\label{sec:results}
We conduct extensive experiments to analyze the empirical effectiveness of our method on a range of long-horizon, sparse-reward domains. In particular, we are going to answer the following questions:
\begin{enumerate}[start=1,label={(\bfseries Q\arabic*)}]
    \item \emph{How well do Q-chunking methods perform compared to prior offline-to-online RL methods?}
    \item \emph{Why does action chunking helps online learning?}
    \item \emph{How does chunk length, critic ensemble size, and update-to-data ratio affect performance?}
\end{enumerate}

\subsection{Environments and Datasets}
We consider six sparse reward robotic manipulation domains with tasks of varying difficulties. This includes 5 domains (5 tasks each) from OGBench~\citep{ogbench_park2024}, \texttt{scene-sparse}, \texttt{puzzle-3x3-sparse}, \texttt{cube-double/triple/quadruple} and 3 tasks from robomimic~\citep{robomimic2021}. For OGBench, we use the default play-style datasets except for \texttt{cube-quadruple} where we use the larger 100M-size dataset. For robomimic, we use the multi-human datasets. See more details in Appendix \ref{appendix:domain}.

\begin{table}[t]
\centering

\renewcommand{\arraystretch}{1.5} 
\scalebox{0.6}{

\begin{tabular}{clcccccc}
\toprule
 &  & \texttt{puzzle-3x3-sparse} & \texttt{scene-sparse} & \texttt{cube-double} & \texttt{cube-triple} & \texttt{cube-quadruple} & \texttt{overall} \\
 &  & \texttt{(5 tasks)} & \texttt{(5 tasks)} & \texttt{(5 tasks)} & \texttt{(5 tasks)} & \texttt{(5 tasks)} & \texttt{(25 tasks)} \\
\midrule
\rowcolor{ourlightblue} & {\texttt{ RLPD}} & $\phantom{0}\scaleto{\text{--}}{1.7pt}\phantom{0}\rightarrow{{\color{ourblue}\mathbf{100}}}$ & $\phantom{0}\scaleto{\text{--}}{1.7pt}\phantom{0}\rightarrow{\phantom{0}{\color{gray!60}\mathbf{94}}}$ & $\phantom{0}\scaleto{\text{--}}{1.7pt}\phantom{0}\rightarrow{\phantom{0}{\color{ourblue}\mathbf{99}}}$ & $\phantom{0}\scaleto{\text{--}}{1.7pt}\phantom{0}\rightarrow{\phantom{0}{\color{gray!60}\mathbf{41}}}$ & $\phantom{0}\scaleto{\text{--}}{1.7pt}\phantom{0}\rightarrow{\phantom{0}\phantom{0}{\color{gray!60}\mathbf{0}}}$ & \cellcolor{ourblue!10}$\phantom{0}\scaleto{\text{--}}{1.7pt}\phantom{0}\rightarrow{\phantom{0}{\color{gray!60}\mathbf{67}}}$ \\
\rowcolor{ourlightblue} & {\texttt{ RLPD-AC}} & $\phantom{0}\scaleto{\text{--}}{1.7pt}\phantom{0}\rightarrow{{\color{ourblue}\mathbf{100}}}$ & $\phantom{0}\scaleto{\text{--}}{1.7pt}\phantom{0}\rightarrow{\phantom{0}{\color{gray!60}\mathbf{91}}}$ & $\phantom{0}\scaleto{\text{--}}{1.7pt}\phantom{0}\rightarrow{\phantom{0}{\color{gray!60}\mathbf{96}}}$ & $\phantom{0}\scaleto{\text{--}}{1.7pt}\phantom{0}\rightarrow{\phantom{0}{\color{gray!60}\mathbf{11}}}$ & $\phantom{0}\scaleto{\text{--}}{1.7pt}\phantom{0}\rightarrow{\phantom{0}\phantom{0}{\color{gray!60}\mathbf{7}}}$ & \cellcolor{ourblue!10}$\phantom{0}\scaleto{\text{--}}{1.7pt}\phantom{0}\rightarrow{\phantom{0}{\color{gray!60}\mathbf{61}}}$ \\
\rowcolor{ourlightblue}\multirow{-3}{*}{\texttt{From Scratch (1-step TD)}} & {\texttt{ SUPE-GT}} & $\phantom{0}\scaleto{\text{--}}{1.7pt}\phantom{0}\rightarrow{{\color{ourblue}\mathbf{100}}}$ & $\phantom{0}\scaleto{\text{--}}{1.7pt}\phantom{0}\rightarrow{\phantom{0}{\color{ourblue}\mathbf{92}}}$ & $\phantom{0}\scaleto{\text{--}}{1.7pt}\phantom{0}\rightarrow{\phantom{0}{\color{gray!60}\mathbf{67}}}$ & $\phantom{0}\scaleto{\text{--}}{1.7pt}\phantom{0}\rightarrow{\phantom{0}\phantom{0}{\color{gray!60}\mathbf{0}}}$ & $\phantom{0}\scaleto{\text{--}}{1.7pt}\phantom{0}\rightarrow{\phantom{0}\phantom{0}{\color{gray!60}\mathbf{0}}}$ & \cellcolor{ourblue!10}$\phantom{0}\scaleto{\text{--}}{1.7pt}\phantom{0}\rightarrow{\phantom{0}{\color{gray!60}\mathbf{52}}}$ \\
\midrule
\rowcolor{ourlightblue} & {\texttt{  IQL}} & ${\phantom{0}\phantom{0}{\color{gray!60}\mathbf{0}}}\rightarrow{\phantom{0}{\color{gray!60}\mathbf{20}}}$ & ${\phantom{0}\phantom{0}{\color{gray!60}\mathbf{0}}}\rightarrow{\phantom{0}{\color{gray!60}\mathbf{39}}}$ & ${\phantom{0}\phantom{0}{\color{gray!60}\mathbf{0}}}\rightarrow{\phantom{0}\phantom{0}{\color{gray!60}\mathbf{0}}}$ & ${\phantom{0}\phantom{0}{\color{gray!60}\mathbf{0}}}\rightarrow{\phantom{0}\phantom{0}{\color{gray!60}\mathbf{0}}}$ & ${\phantom{0}\phantom{0}{\color{gray!60}\mathbf{0}}}\rightarrow{\phantom{0}\phantom{0}{\color{gray!60}\mathbf{0}}}$ & \cellcolor{ourblue!10}${\phantom{0}\phantom{0}{\color{gray!60}\mathbf{0}}}\rightarrow{\phantom{0}{\color{gray!60}\mathbf{12}}}$ \\
\rowcolor{ourlightblue}\multirow{-2}{*}{\texttt{Gaussian (1-step TD)}} & {\texttt{  ReBRAC}} & ${\phantom{0}{\color{gray!60}\mathbf{55}}}\rightarrow{{\color{ourblue}\mathbf{100}}}$ & ${\phantom{0}{\color{gray!60}\mathbf{11}}}\rightarrow{\phantom{0}{\color{ourblue}\mathbf{99}}}$ & ${\phantom{0}\phantom{0}{\color{gray!60}\mathbf{3}}}\rightarrow{\phantom{0}{\color{gray!60}\mathbf{30}}}$ & ${\phantom{0}\phantom{0}{\color{gray!60}\mathbf{0}}}\rightarrow{\phantom{0}\phantom{0}{\color{gray!60}\mathbf{0}}}$ & ${\phantom{0}\phantom{0}{\color{gray!60}\mathbf{1}}}\rightarrow{\phantom{0}{\color{gray!60}\mathbf{20}}}$ & \cellcolor{ourblue!10}${\phantom{0}{\color{gray!60}\mathbf{14}}}\rightarrow{\phantom{0}{\color{gray!60}\mathbf{50}}}$ \\
\midrule
\rowcolor{ourlightblue} & {\texttt{ IFQL}} & ${\phantom{0}{\color{ourblue}\mathbf{98}}}\rightarrow{\phantom{0}{\color{gray!60}\mathbf{97}}}$ & ${\phantom{0}{\color{gray!60}\mathbf{68}}}\rightarrow{\phantom{0}{\color{gray!60}\mathbf{74}}}$ & ${\phantom{0}{\color{gray!60}\mathbf{11}}}\rightarrow{\phantom{0}{\color{gray!60}\mathbf{59}}}$ & ${\phantom{0}\phantom{0}{\color{gray!60}\mathbf{0}}}\rightarrow{\phantom{0}\phantom{0}{\color{gray!60}\mathbf{3}}}$ & ${\phantom{0}\phantom{0}{\color{gray!60}\mathbf{0}}}\rightarrow{\phantom{0}\phantom{0}{\color{gray!60}\mathbf{0}}}$ & \cellcolor{ourblue!10}${\phantom{0}{\color{gray!60}\mathbf{36}}}\rightarrow{\phantom{0}{\color{gray!60}\mathbf{47}}}$ \\
\rowcolor{ourlightblue} & {\texttt{ FQL}} & ${{\color{ourblue}\mathbf{100}}}\rightarrow{{\color{ourblue}\mathbf{100}}}$ & ${\phantom{0}{\color{gray!60}\mathbf{58}}}\rightarrow{\phantom{0}{\color{gray!60}\mathbf{95}}}$ & ${\phantom{0}{\color{gray!60}\mathbf{29}}}\rightarrow{\phantom{0}{\color{gray!60}\mathbf{76}}}$ & ${\phantom{0}\phantom{0}{\color{gray!60}\mathbf{1}}}\rightarrow{\phantom{0}{\color{gray!60}\mathbf{18}}}$ & ${\phantom{0}\phantom{0}{\color{gray!60}\mathbf{0}}}\rightarrow{\phantom{0}\phantom{0}{\color{gray!60}\mathbf{3}}}$ & \cellcolor{ourblue!10}${\phantom{0}{\color{gray!60}\mathbf{37}}}\rightarrow{\phantom{0}{\color{gray!60}\mathbf{58}}}$ \\
\rowcolor{ourlightblue}\multirow{-3}{*}{\texttt{Flow (1-step TD)}} & {\texttt{  BFN}} & ${\phantom{0}{\color{ourblue}\mathbf{98}}}\rightarrow{{\color{ourblue}\mathbf{100}}}$ & ${\phantom{0}{\color{ourblue}\mathbf{85}}}\rightarrow{\phantom{0}{\color{ourblue}\mathbf{99}}}$ & ${\phantom{0}{\color{ourblue}\mathbf{68}}}\rightarrow{\phantom{0}{\color{gray!60}\mathbf{79}}}$ & ${\phantom{0}\phantom{0}{\color{ourblue}\mathbf{4}}}\rightarrow{\phantom{0}{\color{gray!60}\mathbf{23}}}$ & ${\phantom{0}\phantom{0}{\color{gray!60}\mathbf{1}}}\rightarrow{\phantom{0}{\color{gray!60}\mathbf{12}}}$ & \cellcolor{ourblue!10}${\phantom{0}{\color{ourblue}\mathbf{51}}}\rightarrow{\phantom{0}{\color{gray!60}\mathbf{63}}}$ \\
\midrule
\rowcolor{ourlightmiddle} & {\texttt{ IFQL-n}} & ${\phantom{0}{\color{gray!60}\mathbf{94}}}\rightarrow{{\color{ourmiddle}\mathbf{100}}}$ & ${\phantom{0}{\color{gray!60}\mathbf{68}}}\rightarrow{\phantom{0}{\color{gray!60}\mathbf{92}}}$ & ${\phantom{0}\phantom{0}{\color{gray!60}\mathbf{5}}}\rightarrow{\phantom{0}{\color{gray!60}\mathbf{29}}}$ & ${\phantom{0}\phantom{0}{\color{gray!60}\mathbf{0}}}\rightarrow{\phantom{0}\phantom{0}{\color{gray!60}\mathbf{0}}}$ & ${\phantom{0}\phantom{0}{\color{gray!60}\mathbf{0}}}\rightarrow{\phantom{0}\phantom{0}{\color{gray!60}\mathbf{0}}}$ & \cellcolor{ourmiddle!10}${\phantom{0}{\color{gray!60}\mathbf{34}}}\rightarrow{\phantom{0}{\color{gray!60}\mathbf{44}}}$ \\
\rowcolor{ourlightmiddle} & {\texttt{ FQL-n}} & ${\phantom{0}{\color{ourmiddle}\mathbf{98}}}\rightarrow{{\color{ourmiddle}\mathbf{100}}}$ & ${\phantom{0}{\color{gray!60}\mathbf{18}}}\rightarrow{\phantom{0}{\color{gray!60}\mathbf{70}}}$ & ${\phantom{0}{\color{gray!60}\mathbf{11}}}\rightarrow{\phantom{0}{\color{gray!60}\mathbf{77}}}$ & ${\phantom{0}\phantom{0}{\color{gray!60}\mathbf{0}}}\rightarrow{\phantom{0}\phantom{0}{\color{gray!60}\mathbf{1}}}$ & ${\phantom{0}\phantom{0}{\color{ourmiddle}\mathbf{7}}}\rightarrow{\phantom{0}{\color{gray!60}\mathbf{36}}}$ & \cellcolor{ourmiddle!10}${\phantom{0}{\color{gray!60}\mathbf{27}}}\rightarrow{\phantom{0}{\color{gray!60}\mathbf{57}}}$ \\
\rowcolor{ourlightmiddle}\multirow{-3}{*}{\texttt{Flow (n-step TD)}} & {\texttt{ BFN-n}} & ${\phantom{0}{\color{gray!60}\mathbf{58}}}\rightarrow{\phantom{0}{\color{gray!60}\mathbf{90}}}$ & ${\phantom{0}{\color{gray!60}\mathbf{57}}}\rightarrow{\phantom{0}{\color{gray!60}\mathbf{97}}}$ & ${\phantom{0}{\color{gray!60}\mathbf{11}}}\rightarrow{\phantom{0}{\color{gray!60}\mathbf{65}}}$ & ${\phantom{0}\phantom{0}{\color{gray!60}\mathbf{0}}}\rightarrow{\phantom{0}\phantom{0}{\color{gray!60}\mathbf{0}}}$ & ${\phantom{0}\phantom{0}{\color{gray!60}\mathbf{0}}}\rightarrow{\phantom{0}\phantom{0}{\color{gray!60}\mathbf{0}}}$ & \cellcolor{ourmiddle!10}${\phantom{0}{\color{gray!60}\mathbf{25}}}\rightarrow{\phantom{0}{\color{gray!60}\mathbf{50}}}$ \\
\midrule
\rowcolor{ourlightpurple} & {\texttt{ \textbf{\color{ourpurple}QC-IFQL}}} & ${{\color{ourpurple}\mathbf{100}}}\rightarrow{{\color{ourpurple}\mathbf{100}}}$ & ${\phantom{0}{\color{ourpurple}\mathbf{83}}}\rightarrow{\phantom{0}{\color{gray!60}\mathbf{98}}}$ & ${\phantom{0}{\color{gray!60}\mathbf{12}}}\rightarrow{\phantom{0}{\color{gray!60}\mathbf{78}}}$ & ${\phantom{0}\phantom{0}{\color{gray!60}\mathbf{0}}}\rightarrow{\phantom{0}\phantom{0}{\color{gray!60}\mathbf{0}}}$ & ${\phantom{0}\phantom{0}{\color{gray!60}\mathbf{1}}}\rightarrow{\phantom{0}{\color{gray!60}\mathbf{19}}}$ & \cellcolor{ourpurple!10}${\phantom{0}{\color{gray!60}\mathbf{39}}}\rightarrow{\phantom{0}{\color{gray!60}\mathbf{59}}}$ \\
\rowcolor{ourlightpurple} & {\texttt{ \textbf{\color{ourpurple}QC-FQL}}} & ${\phantom{0}{\color{gray!60}\mathbf{63}}}\rightarrow{{\color{ourpurple}\mathbf{100}}}$ & ${\phantom{0}{\color{ourpurple}\mathbf{84}}}\rightarrow{\phantom{0}{\color{ourpurple}\mathbf{99}}}$ & ${\phantom{0}{\color{gray!60}\mathbf{39}}}\rightarrow{{\color{ourpurple}\mathbf{100}}}$ & ${\phantom{0}\phantom{0}{\color{ourpurple}\mathbf{4}}}\rightarrow{\phantom{0}{\color{ourpurple}\mathbf{53}}}$ & ${\phantom{0}\phantom{0}{\color{gray!60}\mathbf{1}}}\rightarrow{\phantom{0}{\color{ourpurple}\mathbf{77}}}$ & \cellcolor{ourpurple!10}${\phantom{0}{\color{gray!60}\mathbf{38}}}\rightarrow{\phantom{0}{\color{ourpurple}\mathbf{86}}}$ \\
\rowcolor{ourlightpurple}\multirow{-3}{*}{\texttt{\color{ourpurple}Q-chunking (Ours)}} & {\texttt{ \textbf{\color{ourpurple}QC}}} & ${{\color{ourpurple}\mathbf{100}}}\rightarrow{{\color{ourpurple}\mathbf{100}}}$ & ${\phantom{0}{\color{ourpurple}\mathbf{84}}}\rightarrow{\phantom{0}{\color{ourpurple}\mathbf{99}}}$ & ${\phantom{0}{\color{ourpurple}\mathbf{67}}}\rightarrow{\phantom{0}{\color{ourpurple}\mathbf{98}}}$ & ${\phantom{0}\phantom{0}{\color{ourpurple}\mathbf{6}}}\rightarrow{\phantom{0}{\color{ourpurple}\mathbf{64}}}$ & ${\phantom{0}\phantom{0}{\color{ourpurple}\mathbf{4}}}\rightarrow{\phantom{0}{\color{gray!60}\mathbf{73}}}$ & \cellcolor{ourpurple!10}${\phantom{0}{\color{ourpurple}\mathbf{52}}}\rightarrow{\phantom{0}{\color{ourpurple}\mathbf{86}}}$ \\
\bottomrule
\end{tabular}
}
\vspace{2mm}
\caption{\footnotesize \textbf{Summary table for OGBench offline-to-online RL results.} For each cell, we report the offline performance after 1M of training steps and then the online performance after 1M of additional online steps. The best method(s) for each column is highlighted in bold and color. Q-chunking methods outperform all prior methods at the end of the online training. See the full results in \cref{tab:full-main-o2o} (complete table) and \cref{fig:main-all-individual} (individual plot).}
\label{tab:main-o2o}
\end{table}

\begin{figure}[t]
    \centering
    \includegraphics[width=\linewidth]{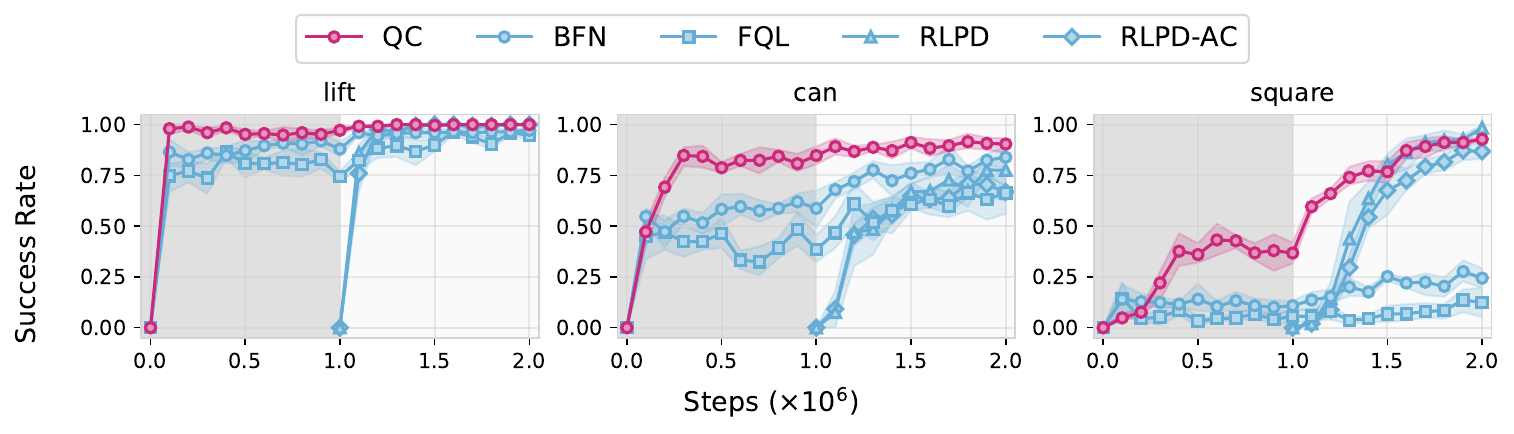} %
    \caption{\footnotesize \textbf{Robomimic results.} \hourpurple{QC} achieves strong performance across all three robomimic tasks. The first 1M steps are offline and the next 1M steps are online with one environment step per training step (5 seeds).}
    \label{fig:main-robomimic}
\end{figure}

\subsection{Comparisons}
We compare with prior methods that speedup value backup as well as the previous best offline-to-online RL methods. We include a brief description of them below with  more  details in \cref{appendix:impl-details}.

\hourblue{BFN (best-of-$N$)} is a baseline that we propose to combine the expected-max $Q$ operator~\citep{ghasemipour2021emaq} with an expressive behavior flow policy. BFN operates in the original action space and uses best-of-$N$ sampling to pick the action (out of $N$) that maximizes the current $Q$-value. This baseline is an ablation to isolate the benefit of Q-chunking in \hourpurple{QC}.

\hourblue{FQL}~\citep{park2025flow} is a recently proposed offline RL method that achieves strong offline and offline-to-online RL performance. 
This baseline is an ablation to isolate the benefit of Q-chunking in \hourpurple{QC-FQL}.

\hourmiddle{BFN-n/FQL-n.} These baselines are the same as BFN/FQL but uses $n$-step backup with $n>1$ (Equation \ref{eq:nstep}) instead of the standard 1-step TD backup. This baseline enjoys the benefits of value backup speedup, but does not use chunked critic or actor, and potentially suffer from the bias issue.

\hourblue{RLPD}~\citep{ball2023efficient}, \hourblue{RLPD-AC}. RLPD is a sample-efficient RL algorithm that treats offline data as additional off-policy data and learn from scratch online. 
RLPD-AC is the same as RLPD but operates on the temporally extended action space. Both of them do not use a behavior constraint.

We also compare with \hourblue{SUPE-GT}~\citep{wilcoxson2024leveraging}, a recently proposed skill-based that is designed for offline-to-online RL. The original method is designed to deal with unlabeled dataset by learning a reward model
with reward bonuses. We adapt it to our setting by directly using the ground truth rewards.

Finally, we compare with additional baselines: \hourblue{IQL}~\citep{kostrikov2021offline}, \hourblue{ReBRAC}~\citep{tarasov2023revisiting}, \hourblue{IFQL} (a baseline implemented in \citet{park2025flow} that combines IQL with rejection sampling policy extraction), and \hourmiddle{IFQL-n} (IFQL with $n$-step return backup).

\begin{wrapfigure}{r}{0.4\textwidth}
    \centering
    \hfill
    \includegraphics[width=\linewidth]{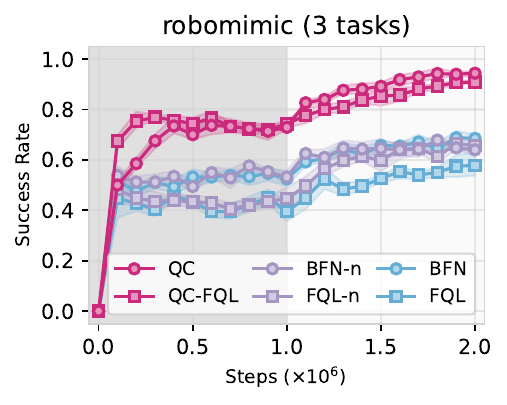} %
    \caption{\footnotesize \textbf{$n$-step return ablations on robomimic.} Both Q-chunking methods consistently outperform their $n$-step and 1-step TD counterparts  (5 seeds).
    }
    \vspace{-2em}
    \label{fig:ablation}
\end{wrapfigure}

\subsection{How well does our method compare to prior offline-to-online RL methods?}
\label{sec:results-analysis}
We report the performance of all Q-chunking methods and the baselines in \cref{tab:main-o2o} (OGBench) and a selective (strong) subset in \cref{fig:main-robomimic} (robomimic). \hourpurple{QC} achieves competitive performance offline, often matching or sometimes outperforming best prior methods. In the online phase, \hourpurple{QC} shows strong sample-efficiency, especially on the two hardest OGBench domains (\texttt{cube-triple/quadruple}), where it outperforms all prior methods (especially on \texttt{cube-quadruple}) by a large margin.
In addition, on OGBench domains, all Q-chunking methods (QC, QC-FQL, QC-IFQL) outperforms both their corresponding $1$-step TD counterpart (BFN, FQL, IFQL) and their correspdoning $n$-step return counterpart (BFN-n, FQL-n, IFQL-n). 
A similar trend continues on robomimic tasks as shown in Figure \ref{fig:ablation}. Across OGBench and robomimic, $n$-step return baselines, which do not use chunked critics or policies, perform significantly worse than Q-chunking methods.

\begin{figure}[t]
    \centering
\includegraphics[width=0.52\linewidth]{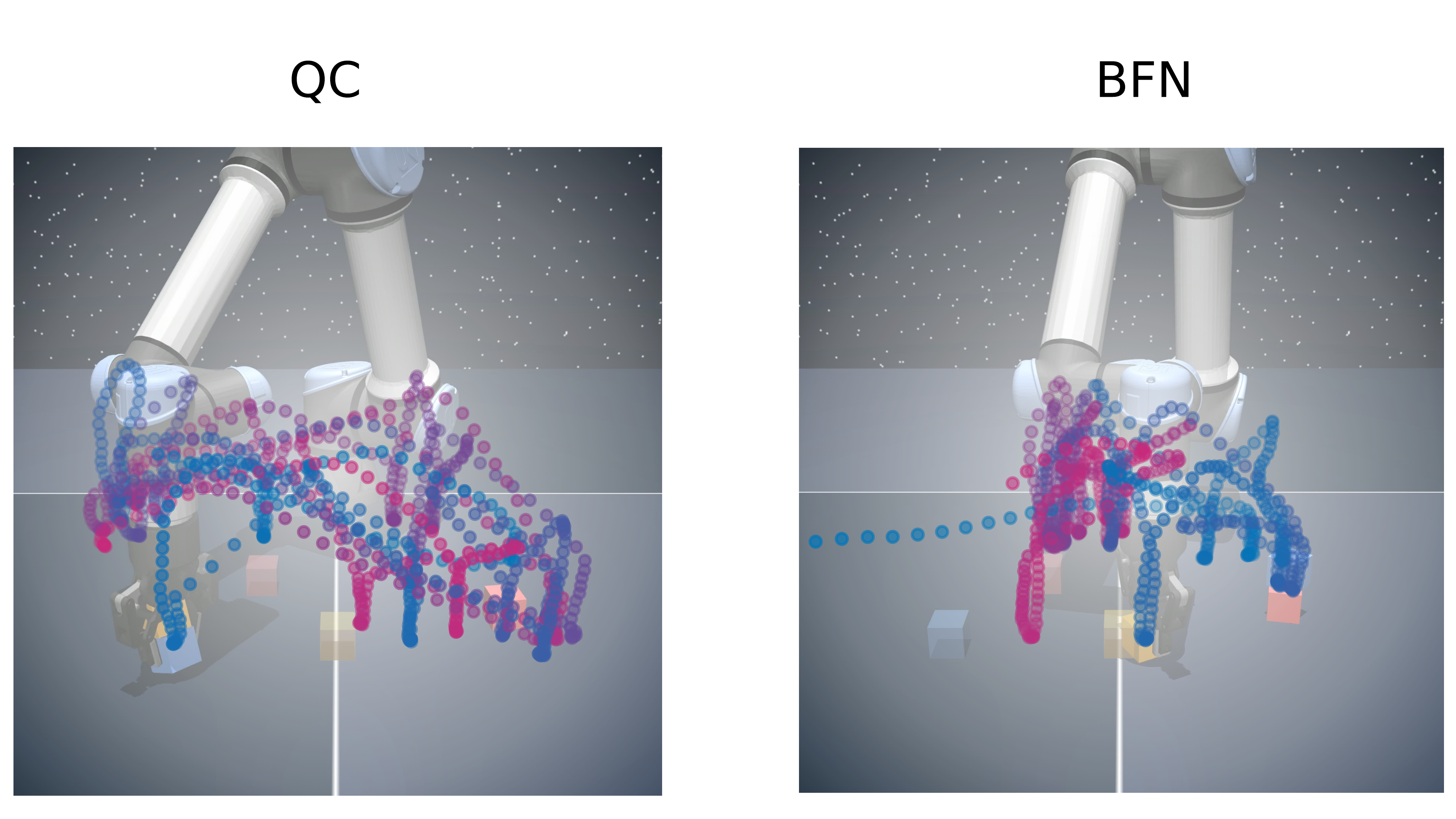}
\includegraphics[width=0.47\linewidth]{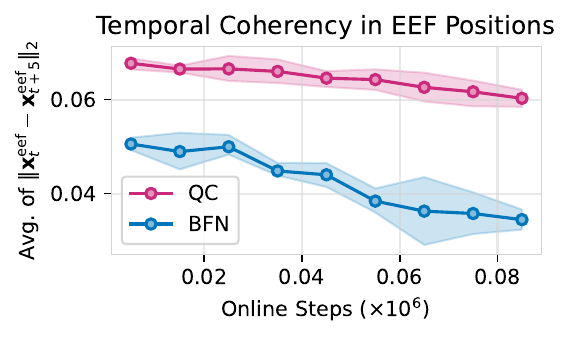} %

\caption{\footnotesize \textbf{End-effector movements {\color{gray}early} in the training and temporal coherency analysis on \texttt{cube-triple-task3}.} \emph{Left:} \hourpurple{QC} covers a more diverse set of states compared to \hourblue{BFN} in the first 1000 environment steps. \emph{Right:} \hourpurple{QC} exhibits a higher temporal coherency in end-effector compared to \hourblue{BFN}.}
\label{fig:eef}
\end{figure}

\subsection{Why does action chunking help exploration?}
We hypothesize in Section \ref{sec:qc-p} that action chunking policy produce more temporally coherent actions and thus lead to better state coverage and exploration. In this section, we study to what degree that holds empirically. We first visualize the end-effector movements early in the training for \hourpurple{QC} and \hourblue{BFN} (Figure \ref{fig:eef}, left). \hourblue{BFN}'s trajectory contains many pauses (as indicated by a very big and dense cluster near the center of the visualization), especially when the end-effector is being lowered to pickup a cube. In contrast, \hourpurple{QC} has fewer pauses (fewer and shallower clusters) and a more diverse state coverage in the end-effector space. We include additional examples in Appendix \ref{appendix:results}, Figure \ref{fig:traj-viz-early} and Figure \ref{fig:traj-viz-late}. To get a quantitative measure of the temporal coherency in actions, we record the 3-D end-effector position throughout training every 5 time steps: $\{\mathbf{x}^{\mathrm{eef}}_0, \mathbf{x}^{\mathrm{eef}}_5, \cdots\}$ and compute the average $L_2$ norm of the difference vector of two adjacent end-effector positions. This average norm would be small if there are any pauses or jittery motions, making a good proxy for measuring the temporal coherency in actions. 
As shown in Figure \ref{fig:eef} (right), \hourpurple{QC} exhibits a higher action temporal coherency throughout training compared to \hourblue{BFN}. This suggests that Q-chunking improves temporal coherency in actions, which explains the improved sample-efficiency that Q-chunking brings.

\begin{figure}[t]
    \includegraphics[width=0.33\linewidth]{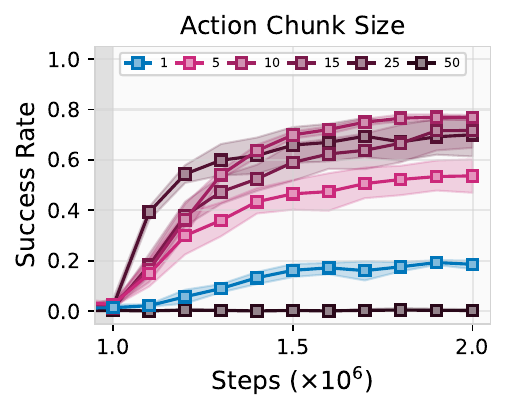} %
    \includegraphics[width=0.33\linewidth]{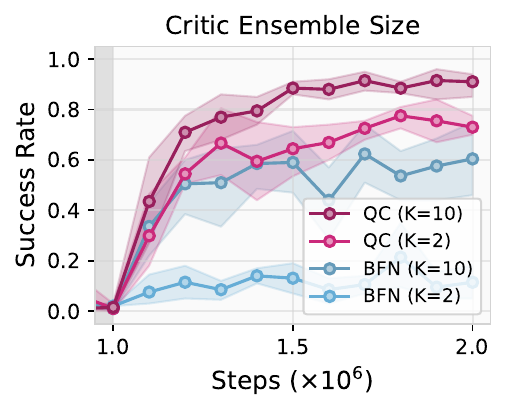} 
    \includegraphics[width=0.33\linewidth]{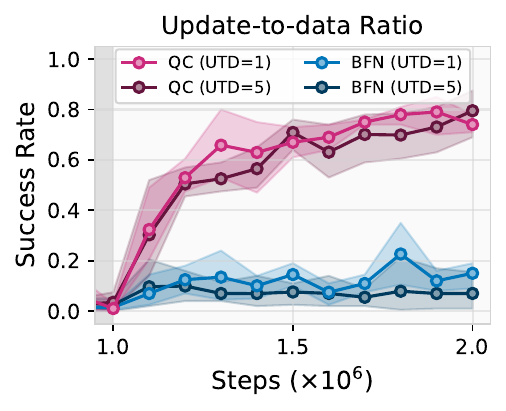} %
    \caption{\footnotesize \textbf{Sensitivity analysis: action chunk size ($h$), critic ensemble size ($K$), and update-to-data ratio (UTD).} \emph{Left:} \hourpurple{QC-FQL} with different $h$ on all 5 \texttt{cube-triple} tasks (5 seeds). QC-FQL with $h=1$ is equivalent to \hourblue{FQL}. \emph{Center:} Increasing the ensemble size to $K=10$ improves performance of both \hourpurple{QC} and \hourblue{BFN} on \texttt{cube-triple-task3} (5 seeds). \emph{Right:} \hourpurple{QC} with UTD of 5 on \texttt{cube-triple-task3} (5 seeds). We report only the online phase results, as all methods achieve near-zero success rates during the offline phase.}
    \label{fig:sensitivity}
\end{figure}
\vspace{-2mm}

\subsection{How does action chunk length, critic ensemble size, and UTD ratio affect performance?}
In Figure \ref{fig:sensitivity} (left), we examine the performance of \hourpurple{QC-FQL} across different action chunk lengths ($h \in \{1, 5, 10, 25, 50\}$) on the \texttt{cube-triple} domain. Increasing the chunk length helps up to $h=10$, after which the asymptotic performance starts to drop. Although $h=25$ shows faster early learning, it fails to achieve the same performance as $h=10$ at the end of online fine-tuning. An even larger chunk length ($h=50$) fails to achieve any success. We suspect that overly large chunk sizes either hurt policy reactivity too much or make policy learning too difficult, as the network must predict a much longer action sequence at once. We use $h=5$ in all our other experiments as $h=5$ generally performs well and is cheaper to run. We also include the individual task breakdown in \cref{fig:c3-ac-ablations} and additional ablation results on \texttt{cube-quadruple} in \cref{fig:c4-ac-ablations}. In Figure \ref{fig:sensitivity} (center), we study how the critic ensemble size affects the performance of our method. Using 10 critics improves both \hourpurple{QC} and \hourblue{BFN}. We use $K=2$ in our other experiments as it is cheap to run. Using $K=10$ could potentially make Q-chunking perform much better on the benchmark tasks we consider. Finally, increasing the update-to-data ratio (UTD) does not improve the sample efficiency of \hourpurple{QC} (Figure \ref{fig:sensitivity}, right). 

\vspace{-1mm}
\section{Discussions}
\vspace{-1mm}
\label{sec:discussions}
We demonstrate how action chunking can be integrated into an offline-to-online RL agent with a simple recipe. Our approach speeds up value backup and explores more effectively online with temporally coherent actions. As a result, it outperforms prior offline-to-online methods on a range of challenging long-horizon tasks. Our work serves as a step towards training non-Markovian policy for effective online exploration from prior offline data. Several challenges remain, opening promising directions for future research. First, our approach use a fixed action chunk, but it is unclear how to choose this size other than task-specific hyperparameter tuning. A natural next step would be to develop mechanisms that automatically determine chunk boundaries. Second, action chunking represents only a limited subclass of non-Markovian policies and may perform poorly in settings where a high-frequency control feedback loop is essential. Developing practical techniques for training more general non-Markovian policies for online exploration would further improve the online sample efficiency of offline-to-online RL algorithms.

\section*{Acknowledgments}
This research was partially supported by RAI, and ONR N00014-22-1-2773. This research used the Savio computational cluster resource provided by the Berkeley Research Computing program at UC Berkeley. We would like to thank Seohong Park for providing the code infrastructure and the 100M dataset for \texttt{cube-quadruple}. We would also like to thank Ameesh Shah, Chuer Pan, Oleg Rybkin, Andrew Wagenmaker, Seohong Park, Yifei Zhou, Fangchen Liu, William Chen for discussions on the method and feedback on the early draft of the paper. We would also like to thank Chongyi Zheng and Hyunwoo Park for spotting typos in the paper. Finally, we would like to thank NeurIPS review committee for thoughtful comments and suggestions.

\bibliography{main}

@article{de2025learning,
  title={Learning Transferable Sub-Goals by Hypothesizing Generalizing Features},
  author={de Mello Koch, Anita and Bagaria, Akhil and Huo, Bingnan and Zhou, Zhiyuan and Allen, Cameron and Konidaris, George},
  year={2025}
}

@article{kober2008policy,
  title={Policy search for motor primitives in robotics},
  author={Kober, Jens and Peters, Jan},
  journal={Advances in neural information processing systems},
  volume={21},
  year={2008}
}

@incollection{schaal2006dynamic,
  title={Dynamic movement primitives-a framework for motor control in humans and humanoid robotics},
  author={Schaal, Stefan},
  booktitle={Adaptive motion of animals and machines},
  pages={261--280},
  year={2006},
  publisher={Springer}
}

@article{whitley1993genetic,
  title={Genetic reinforcement learning for neurocontrol problems},
  author={Whitley, Darrell and Dominic, Stephen and Das, Rajarshi and Anderson, Charles W},
  journal={Machine Learning},
  volume={13},
  number={2},
  pages={259--284},
  year={1993},
  publisher={Springer}
}

@article{wurman2022outracing,
  title={Outracing champion Gran Turismo drivers with deep reinforcement learning},
  author={Wurman, Peter R and Barrett, Samuel and Kawamoto, Kenta and MacGlashan, James and Subramanian, Kaushik and Walsh, Thomas J and Capobianco, Roberto and Devlic, Alisa and Eckert, Franziska and Fuchs, Florian and others},
  journal={Nature},
  volume={602},
  number={7896},
  pages={223--228},
  year={2022},
  publisher={Nature Publishing Group UK London}
}

@inproceedings{kozuno2021revisiting,
  title={Revisiting Peng’s Q ($\lambda$) for Modern Reinforcement Learning},
  author={Kozuno, Tadashi and Tang, Yunhao and Rowland, Mark and Munos, R{\'e}mi and Kapturowski, Steven and Dabney, Will and Valko, Michal and Abel, David},
  booktitle={International Conference on Machine Learning},
  pages={5794--5804},
  year={2021},
  organization={PMLR}
}

@inproceedings{kapturowski2018recurrent,
  title={Recurrent experience replay in distributed reinforcement learning},
  author={Kapturowski, Steven and Ostrovski, Georg and Quan, John and Munos, Remi and Dabney, Will},
  booktitle={International conference on learning representations},
  year={2018}
}

@inproceedings{mnih2016asynchronous,
  title={Asynchronous methods for deep reinforcement learning},
  author={Mnih, Volodymyr and Badia, Adria Puigdomenech and Mirza, Mehdi and Graves, Alex and Lillicrap, Timothy and Harley, Tim and Silver, David and Kavukcuoglu, Koray},
  booktitle={International conference on machine learning},
  pages={1928--1937},
  year={2016},
  organization={PmLR}
}

@article{watkins1989learning,
  title={Learning from delayed rewards},
  author={Watkins, Christopher John Cornish Hellaby and others},
  year={1989},
  publisher={King's College, Cambridge United Kingdom}
}

@inproceedings{hessel2018rainbow,
  title={Rainbow: Combining improvements in deep reinforcement learning},
  author={Hessel, Matteo and Modayil, Joseph and Van Hasselt, Hado and Schaul, Tom and Ostrovski, Georg and Dabney, Will and Horgan, Dan and Piot, Bilal and Azar, Mohammad and Silver, David},
  booktitle={Proceedings of the AAAI conference on artificial intelligence},
  volume={32},
  number={1},
  year={2018}
}

@book{sutton1998reinforcement,
  title={Reinforcement learning: An introduction},
  author={Sutton, Richard S and Barto, Andrew G and others},
  volume={1},
  number={1},
  year={1998},
  publisher={MIT press Cambridge}
}

@inproceedings{fedus2020revisiting,
  title={Revisiting fundamentals of experience replay},
  author={Fedus, William and Ramachandran, Prajit and Agarwal, Rishabh and Bengio, Yoshua and Larochelle, Hugo and Rowland, Mark and Dabney, Will},
  booktitle={International conference on machine learning},
  pages={3061--3071},
  year={2020},
  organization={PMLR}
}

@article{liu2022flow,
  title={Flow straight and fast: Learning to generate and transfer data with rectified flow},
  author={Liu, Xingchao and Gong, Chengyue and Liu, Qiang},
  journal={arXiv preprint arXiv:2209.03003},
  year={2022}
}

@inproceedings{ghasemipour2021emaq,
  title={Emaq: Expected-max q-learning operator for simple yet effective offline and online rl},
  author={Ghasemipour, Seyed Kamyar Seyed and Schuurmans, Dale and Gu, Shixiang Shane},
  booktitle={International Conference on Machine Learning},
  pages={3682--3691},
  year={2021},
  organization={PMLR}
}

@article{nair2020awac,
  title={Awac: Accelerating online reinforcement learning with offline datasets},
  author={Nair, Ashvin and Gupta, Abhishek and Dalal, Murtaza and Levine, Sergey},
  journal={arXiv preprint arXiv:2006.09359},
  year={2020}
}

@article{tian2025chunking,
  title={Chunking the Critic: A Transformer-based Soft Actor-Critic with N-Step Returns},
  author={Tian, Dong and Li, Ge and Zhou, Hongyi and Celik, Onur and Neumann, Gerhard},
  journal={arXiv preprint arXiv:2503.03660},
  year={2025}
}

@inproceedings{
seo2024continuous,
title={Continuous Control with Coarse-to-fine Reinforcement Learning},
author={Younggyo Seo and Jafar Uru{\c{c}} and Stephen James},
booktitle={8th Annual Conference on Robot Learning},
year={2024},
url={https://openreview.net/forum?id=WjDR48cL3O}
}

@article{durugkar2016deep,
  title={Deep reinforcement learning with macro-actions},
  author={Durugkar, Ishan P and Rosenbaum, Clemens and Dernbach, Stefan and Mahadevan, Sridhar},
  journal={arXiv preprint arXiv:1606.04615},
  year={2016}
}

@article{mcgovern1998macro,
  title={Macro-actions in reinforcement learning: An empirical analysis},
  author={McGovern, Amy and Sutton, Richard S},
  year={1998}
}

@article{tarasov2023revisiting,
  title={Revisiting the minimalist approach to offline reinforcement learning},
  author={Tarasov, Denis and Kurenkov, Vladislav and Nikulin, Alexander and Kolesnikov, Sergey},
  journal={Advances in Neural Information Processing Systems},
  volume={36},
  pages={11592--11620},
  year={2023}
}

@inproceedings{
li2024top,
title={{TOP}-{ERL}: Transformer-based Off-Policy Episodic Reinforcement Learning},
author={Ge Li and Dong Tian and Hongyi Zhou and Xinkai Jiang and Rudolf Lioutikov and Gerhard Neumann},
booktitle={The Thirteenth International Conference on Learning Representations},
year={2025},
url={https://openreview.net/forum?id=N4NhVN30ph}
}

@inproceedings{bharadhwaj2024roboagent,
  title={Roboagent: Generalization and efficiency in robot manipulation via semantic augmentations and action chunking},
  author={Bharadhwaj, Homanga and Vakil, Jay and Sharma, Mohit and Gupta, Abhinav and Tulsiani, Shubham and Kumar, Vikash},
  booktitle={2024 IEEE International Conference on Robotics and Automation (ICRA)},
  pages={4788--4795},
  year={2024},
  organization={IEEE}
}

@article{george2023one,
  title={One act play: Single demonstration behavior cloning with action chunking transformers},
  author={George, Abraham and Farimani, Amir Barati},
  journal={arXiv preprint arXiv:2309.10175},
  year={2023}
}

@article{wang2023train,
  title={Train once, get a family: State-adaptive balances for offline-to-online reinforcement learning},
  author={Wang, Shenzhi and Yang, Qisen and Gao, Jiawei and Lin, Matthieu and Chen, Hao and Wu, Liwei and Jia, Ning and Song, Shiji and Huang, Gao},
  journal={Advances in Neural Information Processing Systems},
  volume={36},
  pages={47081--47104},
  year={2023}
}

@article{luo2023finetuning,
  title={Finetuning from offline reinforcement learning: Challenges, trade-offs and practical solutions},
  author={Luo, Yicheng and Kay, Jackie and Grefenstette, Edward and Deisenroth, Marc Peter},
  journal={arXiv preprint arXiv:2303.17396},
  year={2023}
}

@article{agarwal2022reincarnating,
  title={Reincarnating reinforcement learning: Reusing prior computation to accelerate progress},
  author={Agarwal, Rishabh and Schwarzer, Max and Castro, Pablo Samuel and Courville, Aaron C and Bellemare, Marc},
  journal={Advances in neural information processing systems},
  volume={35},
  pages={28955--28971},
  year={2022}
}

@article{zhou2024efficient,
  title={Efficient online reinforcement learning fine-tuning need not retain offline data},
  author={Zhou, Zhiyuan and Peng, Andy and Li, Qiyang and Levine, Sergey and Kumar, Aviral},
  journal={arXiv preprint arXiv:2412.07762},
  year={2024}
}

@article{wilcoxson2024leveraging,
  title={Leveraging Skills from Unlabeled Prior Data for Efficient Online Exploration},
  author={Wilcoxson, Max and Li, Qiyang and Frans, Kevin and Levine, Sergey},
  journal={arXiv preprint arXiv:2410.18076},
  year={2024}
}

@article{ren2024diffusion,
  title={Diffusion policy policy optimization},
  author={Ren, Allen Z and Lidard, Justin and Ankile, Lars L and Simeonov, Anthony and Agrawal, Pulkit and Majumdar, Anirudha and Burchfiel, Benjamin and Dai, Hongkai and Simchowitz, Max},
  journal={arXiv preprint arXiv:2409.00588},
  year={2024}
}

@inproceedings{robomimic2021,
  title={What Matters in Learning from Offline Human Demonstrations for Robot Manipulation},
  author={Ajay Mandlekar and Danfei Xu and Josiah Wong and Soroush Nasiriany and Chen Wang and Rohun Kulkarni and Li Fei-Fei and Silvio Savarese and Yuke Zhu and Roberto Mart\'{i}n-Mart\'{i}n},
  booktitle={arXiv preprint arXiv:2108.03298},
  year={2021}
}

@article{park2024ogbench,
  title={Ogbench: Benchmarking offline goal-conditioned rl},
  author={Park, Seohong and Frans, Kevin and Eysenbach, Benjamin and Levine, Sergey},
  journal={arXiv preprint arXiv:2410.20092},
  year={2024}
}

@inproceedings{touati2022does,
  title={Does Zero-Shot Reinforcement Learning Exist?},
  author={Touati, Ahmed and Rapin, J{\'e}r{\'e}my and Ollivier, Yann},
  booktitle={The Eleventh International Conference on Learning Representations},
  year={2022}
}

@inproceedings{bagaria2019option,
  title={Option discovery using deep skill chaining},
  author={Bagaria, Akhil and Konidaris, George},
  booktitle={International Conference on Learning Representations},
  year={2019}
}

@techreport{csimcsek2007betweenness,
title = "Betweenness Centrality as a Basis for Forming Skills",
author = "{\"O}zg{\"u}r {\c S}im{\c s}ek and Barto, {Andrew G.}",
year = "2007",
month = apr,
day = "12",
language = "English",
publisher = "University of Massachusetts Amherst",
type = "WorkingPaper",
institution = "University of Massachusetts Amherst",
}

@inproceedings{NEURIPS2022_ba1c5356,
 author = {Agarwal, Rishabh and Schwarzer, Max and Castro, Pablo Samuel and Courville, Aaron C and Bellemare, Marc},
 booktitle = {Advances in Neural Information Processing Systems},
 editor = {S. Koyejo and S. Mohamed and A. Agarwal and D. Belgrave and K. Cho and A. Oh},
 pages = {28955--28971},
 publisher = {Curran Associates, Inc.},
 title = {Reincarnating Reinforcement Learning: Reusing Prior Computation to Accelerate Progress},
 volume = {35},
 year = {2022}
}

@inproceedings{
ajay2020opal,
title={{OPAL}: Offline Primitive Discovery for Accelerating Offline Reinforcement Learning},
author={Anurag Ajay and Aviral Kumar and Pulkit Agrawal and Sergey Levine and Ofir Nachum},
booktitle={International Conference on Learning Representations},
year={2021},
url={https://openreview.net/forum?id=V69LGwJ0lIN}
}

@inproceedings{bacon2017option,
  title={The option-critic architecture},
  author={Bacon, Pierre-Luc and Harb, Jean and Precup, Doina},
  booktitle={Proceedings of the AAAI conference on artificial intelligence},
  volume={31},
  number={1},
  year={2017}
}

@article{srinivas2016option,
  title={Option discovery in hierarchical reinforcement learning using spatio-temporal clustering},
  author={Srinivas, Aravind and Krishnamurthy, Ramnandan and Kumar, Peeyush and Ravindran, Balaraman},
  journal={arXiv preprint arXiv:1605.05359},
  year={2016}
}

@article{bagaria2024effectively,
  title={Effectively learning initiation sets in hierarchical reinforcement learning},
  author={Bagaria, Akhil and Abbatematteo, Ben and Gottesman, Omer and Corsaro, Matt and Rammohan, Sreehari and Konidaris, George},
  journal={Advances in Neural Information Processing Systems},
  volume={36},
  year={2024}
}

@inproceedings{ball2023efficient,
  title={Efficient online reinforcement learning with offline data},
  author={Ball, Philip J and Smith, Laura and Kostrikov, Ilya and Levine, Sergey},
  booktitle={International Conference on Machine Learning},
  pages={1577--1594},
  year={2023},
  organization={PMLR}
}

@article{chen2024self,
  title={Self-supervised reinforcement learning that transfers using random features},
  author={Chen, Boyuan and Zhu, Chuning and Agrawal, Pulkit and Zhang, Kaiqing and Gupta, Abhishek},
  journal={Advances in Neural Information Processing Systems},
  volume={36},
  year={2024}
}

@article{daniel2016hierarchical,
  title={Hierarchical relative entropy policy search},
  author={Daniel, Christian and Neumann, Gerhard and Kroemer, Oliver and Peters, Jan},
  journal={Journal of Machine Learning Research},
  volume={17},
  number={93},
  pages={1--50},
  year={2016}
}

@article{fox2017multi,
  title={Multi-level discovery of deep options},
  author={Fox, Roy and Krishnan, Sanjay and Stoica, Ion and Goldberg, Ken},
  journal={arXiv preprint arXiv:1703.08294},
  year={2017}
}

@InProceedings{frans2024unsupervised,
  title = 	 {Unsupervised Zero-Shot Reinforcement Learning via Functional Reward Encodings},
  author =       {Frans, Kevin and Park, Seohong and Abbeel, Pieter and Levine, Sergey},
  booktitle = 	 {Proceedings of the 41st International Conference on Machine Learning},
  pages = 	 {13927--13942},
  year = 	 {2024},
  editor = 	 {Salakhutdinov, Ruslan and Kolter, Zico and Heller, Katherine and Weller, Adrian and Oliver, Nuria and Scarlett, Jonathan and Berkenkamp, Felix},
  volume = 	 {235},
  series = 	 {Proceedings of Machine Learning Research},
  month = 	 {21--27 Jul},
  publisher =    {PMLR},
  url = 	 {https://proceedings.mlr.press/v235/frans24a.html}
}

@article{fu2020d4rl,
  title={{D4RL}: Datasets for deep data-driven reinforcement learning},
  author={Fu, Justin and Kumar, Aviral and Nachum, Ofir and Tucker, George and Levine, Sergey},
  journal={arXiv preprint arXiv:2004.07219},
  year={2020}
}

@inproceedings{haarnoja2018soft,
  title={Soft actor-critic: Off-policy maximum entropy deep reinforcement learning with a stochastic actor},
  author={Haarnoja, Tuomas and Zhou, Aurick and Abbeel, Pieter and Levine, Sergey},
  booktitle={International conference on machine learning},
  pages={1861--1870},
  year={2018},
  organization={PMLR}
}

@inproceedings{
hu2023unsupervised,
title={Unsupervised Behavior Extraction via Random Intent Priors},
author={Hao Hu and Yiqin Yang and Jianing Ye and Ziqing Mai and Chongjie Zhang},
booktitle={Thirty-seventh Conference on Neural Information Processing Systems},
year={2023},
url={https://openreview.net/forum?id=4vGVQVz5KG}
}

@article{kim2019variational,
  title={Variational temporal abstraction},
  author={Kim, Taesup and Ahn, Sungjin and Bengio, Yoshua},
  journal={Advances in Neural Information Processing Systems},
  volume={32},
  year={2019}
}

@book{konidaris2011autonomous,
  title={Autonomous robot skill acquisition},
  author={Konidaris, George Dimitri},
  year={2011},
  publisher={University of Massachusetts Amherst}
}

@article{kostrikov2021offline,
  title={Offline reinforcement learning with implicit {Q}-learning},
  author={Kostrikov, Ilya and Nair, Ashvin and Levine, Sergey},
  journal={arXiv preprint arXiv:2110.06169},
  year={2021}
}

@article{kumar2020conservative,
  title={Conservative {Q}-learning for offline reinforcement learning},
  author={Kumar, Aviral and Zhou, Aurick and Tucker, George and Levine, Sergey},
  journal={Advances in Neural Information Processing Systems},
  volume={33},
  pages={1179--1191},
  year={2020}
}

@inproceedings{lee2022offline,
  title={Offline-to-online reinforcement learning via balanced replay and pessimistic {Q}-ensemble},
  author={Lee, Seunghyun and Seo, Younggyo and Lee, Kimin and Abbeel, Pieter and Shin, Jinwoo},
  booktitle={Conference on Robot Learning},
  pages={1702--1712},
  year={2022},
  organization={PMLR}
}

@article{li2024accelerating,
  title={Accelerating exploration with unlabeled prior data},
  author={Li, Qiyang and Zhang, Jason and Ghosh, Dibya and Zhang, Amy and Levine, Sergey},
  journal={Advances in Neural Information Processing Systems},
  volume={36},
  year={2024}
}

@inproceedings{mannor2004dynamic,
  title={Dynamic abstraction in reinforcement learning via clustering},
  author={Mannor, Shie and Menache, Ishai and Hoze, Amit and Klein, Uri},
  booktitle={Proceedings of the twenty-first international conference on Machine learning},
  pages={71},
  year={2004}
}

@inproceedings{menache2002q,
  title={{Q}-cut—dynamic discovery of sub-goals in reinforcement learning},
  author={Menache, Ishai and Mannor, Shie and Shimkin, Nahum},
  booktitle={Machine Learning: ECML 2002: 13th European Conference on Machine Learning Helsinki, Finland, August 19--23, 2002 Proceedings 13},
  pages={295--306},
  year={2002},
  organization={Springer}
}

@article{nakamoto2024cal,
  title={{Cal-QL}: Calibrated offline {RL} pre-training for efficient online fine-tuning},
  author={Nakamoto, Mitsuhiko and Zhai, Simon and Singh, Anikait and Sobol Mark, Max and Ma, Yi and Finn, Chelsea and Kumar, Aviral and Levine, Sergey},
  journal={Advances in Neural Information Processing Systems},
  volume={36},
  year={2024}
}

@inproceedings{nasiriany2022learning,
  title={Learning and Retrieval from Prior Data for Skill-based Imitation Learning},
  author={Nasiriany, Soroush and Gao, Tian and Mandlekar, Ajay and Zhu, Yuke},
  booktitle={Conference on Robot Learning},
  year={2022}
}

@inproceedings{
    park2024foundation,
    title={Foundation Policies with Hilbert Representations},
    author={Seohong Park and Tobias Kreiman and Sergey Levine},
    booktitle={Forty-first International Conference on Machine Learning},
    year={2024},
    url={https://openreview.net/forum?id=LhNsSaAKub}
}

@inproceedings{pertsch2021accelerating,
  title={Accelerating reinforcement learning with learned skill priors},
  author={Pertsch, Karl and Lee, Youngwoon and Lim, Joseph},
  booktitle={Conference on robot learning},
  pages={188--204},
  year={2021},
  organization={PMLR}
}

@inproceedings{shankar2020learning,
  title={Learning robot skills with temporal variational inference},
  author={Shankar, Tanmay and Gupta, Abhinav},
  booktitle={International Conference on Machine Learning},
  pages={8624--8633},
  year={2020},
  organization={PMLR}
}

@inproceedings{csimcsek2004using,
  title={Using relative novelty to identify useful temporal abstractions in reinforcement learning},
  author={{\c{S}}im{\c{s}}ek, {\"O}zg{\"u}r and Barto, Andrew G},
  booktitle={Proceedings of the twenty-first international conference on Machine learning},
  pages={95},
  year={2004}
}

@inproceedings{
    singh2020parrot,
    title={Parrot: Data-Driven Behavioral Priors for Reinforcement Learning},
    author={Avi Singh and Huihan Liu and Gaoyue Zhou and Albert Yu and Nicholas Rhinehart and Sergey Levine},
    booktitle={International Conference on Learning Representations},
    year={2021},
    url={https://openreview.net/forum?id=Ysuv-WOFeKR}
}

@inproceedings{
    song2022hybrid,
    title={Hybrid {RL}: Using both offline and online data can make {RL} efficient},
    author={Yuda Song and Yifei Zhou and Ayush Sekhari and Drew Bagnell and Akshay Krishnamurthy and Wen Sun},
    booktitle={The Eleventh International Conference on Learning Representations },
    year={2023},
    url={https://openreview.net/forum?id=yyBis80iUuU}
}

@article{sutton1999between,
  title={Between {MDPs} and {semi-MDPs}: A framework for temporal abstraction in reinforcement learning},
  author={Sutton, Richard S and Precup, Doina and Singh, Satinder},
  journal={Artificial intelligence},
  volume={112},
  number={1-2},
  pages={181--211},
  year={1999},
  publisher={Elsevier}
}

@article{tarasov2024revisiting,
  title={Revisiting the minimalist approach to offline reinforcement learning},
  author={Tarasov, Denis and Kurenkov, Vladislav and Nikulin, Alexander and Kolesnikov, Sergey},
  journal={Advances in Neural Information Processing Systems},
  volume={36},
  year={2024}
}

@article{xie2021policy,
  title={Policy finetuning: Bridging sample-efficient offline and online reinforcement learning},
  author={Xie, Tengyang and Jiang, Nan and Wang, Huan and Xiong, Caiming and Bai, Yu},
  journal={Advances in neural information processing systems},
  volume={34},
  pages={27395--27407},
  year={2021}
}

@inproceedings{
zhang2023policy,
title={Policy Expansion for Bridging Offline-to-Online Reinforcement Learning},
author={Haichao Zhang and Wei Xu and Haonan Yu},
booktitle={The Eleventh International Conference on Learning Representations },
year={2023},
url={https://openreview.net/forum?id=-Y34L45JR6z}
}

@inproceedings{zheng2023adaptive,
  title={Adaptive policy learning for offline-to-online reinforcement learning},
  author={Zheng, Han and Luo, Xufang and Wei, Pengfei and Song, Xuan and Li, Dongsheng and Jiang, Jing},
  booktitle={Proceedings of the AAAI Conference on Artificial Intelligence},
  volume={37},
  number={9},
  pages={11372--11380},
  year={2023}
}

@article{ogbench_park2024,
  title={OGBench: Benchmarking Offline Goal-Conditioned RL},
  author={Seohong Park and Kevin Frans and Benjamin Eysenbach and Sergey Levine},
  journal={ArXiv},
  year={2024}
}

@article{dalal2021accelerating,
  title={Accelerating robotic reinforcement learning via parameterized action primitives},
  author={Dalal, Murtaza and Pathak, Deepak and Salakhutdinov, Russ R},
  journal={Advances in Neural Information Processing Systems},
  volume={34},
  pages={21847--21859},
  year={2021}
}

@article{kulkarni2016hierarchical,
  title={Hierarchical deep reinforcement learning: Integrating temporal abstraction and intrinsic motivation},
  author={Kulkarni, Tejas D and Narasimhan, Karthik and Saeedi, Ardavan and Tenenbaum, Josh},
  journal={Advances in neural information processing systems},
  volume={29},
  year={2016}
}

@article{nachum2018data,
  title={Data-efficient hierarchical reinforcement learning},
  author={Nachum, Ofir and Gu, Shixiang Shane and Lee, Honglak and Levine, Sergey},
  journal={Advances in neural information processing systems},
  volume={31},
  year={2018}
}

@article{gehring2021hierarchical,
  title={Hierarchical skills for efficient exploration},
  author={Gehring, Jonas and Synnaeve, Gabriel and Krause, Andreas and Usunier, Nicolas},
  journal={Advances in Neural Information Processing Systems},
  volume={34},
  pages={11553--11564},
  year={2021}
}

@inproceedings{
xie2021latent,
title={Latent Skill Planning for Exploration and Transfer},
author={Kevin Xie and Homanga Bharadhwaj and Danijar Hafner and Animesh Garg and Florian Shkurti},
booktitle={International Conference on Learning Representations},
year={2021},
url={https://openreview.net/forum?id=jXe91kq3jAq}
}

@article{paraschos2013probabilistic,
  title={Probabilistic movement primitives},
  author={Paraschos, Alexandros and Daniel, Christian and Peters, Jan R and Neumann, Gerhard},
  journal={Advances in neural information processing systems},
  volume={26},
  year={2013}
}

@article{dietterich2000hierarchical,
  title={Hierarchical reinforcement learning with the MAXQ value function decomposition},
  author={Dietterich, Thomas G},
  journal={Journal of artificial intelligence research},
  volume={13},
  pages={227--303},
  year={2000}
}

@article{peng2017deeploco,
  title={Deeploco: Dynamic locomotion skills using hierarchical deep reinforcement learning},
  author={Peng, Xue Bin and Berseth, Glen and Yin, KangKang and Van De Panne, Michiel},
  journal={Acm transactions on graphics (tog)},
  volume={36},
  number={4},
  pages={1--13},
  year={2017},
  publisher={ACM New York, NY, USA}
}

@article{merel2018neural,
  title={Neural probabilistic motor primitives for humanoid control},
  author={Merel, Josh and Hasenclever, Leonard and Galashov, Alexandre and Ahuja, Arun and Pham, Vu and Wayne, Greg and Teh, Yee Whye and Heess, Nicolas},
  journal={arXiv preprint arXiv:1811.11711},
  year={2018}
}

@article{dayan1992feudal,
  title={Feudal reinforcement learning},
  author={Dayan, Peter and Hinton, Geoffrey E},
  journal={Advances in neural information processing systems},
  volume={5},
  year={1992}
}

@inproceedings{vezhnevets2017feudal,
  title={Feudal networks for hierarchical reinforcement learning},
  author={Vezhnevets, Alexander Sasha and Osindero, Simon and Schaul, Tom and Heess, Nicolas and Jaderberg, Max and Silver, David and Kavukcuoglu, Koray},
  booktitle={International conference on machine learning},
  pages={3540--3549},
  year={2017},
  organization={PMLR}
}

@inproceedings{riedmiller2018learning,
  title={Learning by playing solving sparse reward tasks from scratch},
  author={Riedmiller, Martin and Hafner, Roland and Lampe, Thomas and Neunert, Michael and Degrave, Jonas and Wiele, Tom and Mnih, Vlad and Heess, Nicolas and Springenberg, Jost Tobias},
  booktitle={International conference on machine learning},
  pages={4344--4353},
  year={2018},
  organization={PMLR}
}

@article{chentanez2004intrinsically,
  title={Intrinsically motivated reinforcement learning},
  author={Chentanez, Nuttapong and Barto, Andrew and Singh, Satinder},
  journal={Advances in neural information processing systems},
  volume={17},
  year={2004}
}

@article{oh2017value,
  title={Value prediction network},
  author={Oh, Junhyuk and Singh, Satinder and Lee, Honglak},
  journal={Advances in neural information processing systems},
  volume={30},
  year={2017}
}

@article{zhao2023learning,
  title={Learning fine-grained bimanual manipulation with low-cost hardware},
  author={Zhao, Tony Z and Kumar, Vikash and Levine, Sergey and Finn, Chelsea},
  journal={arXiv preprint arXiv:2304.13705},
  year={2023}
}

@article{seo2024reinforcement,
  title={Reinforcement Learning with Action Sequence for Data-Efficient Robot Learning},
  author={Seo, Younggyo and Abbeel, Pieter},
  journal={arXiv preprint arXiv:2411.12155},
  year={2024}
}

@article{vezhnevets2016strategic,
  title={Strategic attentive writer for learning macro-actions},
  author={Vezhnevets, Alexander and Mnih, Volodymyr and Osindero, Simon and Graves, Alex and Vinyals, Oriol and Agapiou, John and others},
  journal={Advances in neural information processing systems},
  volume={29},
  year={2016}
}

@article{park2025flow,
  title={Flow {Q}-Learning},
  author={Park, Seohong and Li, Qiyang and Levine, Sergey},
  journal={arXiv preprint arXiv:2502.02538},
  year={2025}
}

@article{stiennon2020learning,
  title={Learning to summarize with human feedback},
  author={Stiennon, Nisan and Ouyang, Long and Wu, Jeffrey and Ziegler, Daniel and Lowe, Ryan and Voss, Chelsea and Radford, Alec and Amodei, Dario and Christiano, Paul F},
  journal={Advances in neural information processing systems},
  volume={33},
  pages={3008--3021},
  year={2020}
}

@article{schrittwieser2020mastering,
  title={Mastering atari, go, chess and shogi by planning with a learned model},
  author={Schrittwieser, Julian and Antonoglou, Ioannis and Hubert, Thomas and Simonyan, Karen and Sifre, Laurent and Schmitt, Simon and Guez, Arthur and Lockhart, Edward and Hassabis, Demis and Graepel, Thore and others},
  journal={Nature},
  volume={588},
  number={7839},
  pages={604--609},
  year={2020},
  publisher={Nature Publishing Group}
}

@article{chi2023diffusion,
  title={Diffusion policy: Visuomotor policy learning via action diffusion},
  author={Chi, Cheng and Xu, Zhenjia and Feng, Siyuan and Cousineau, Eric and Du, Yilun and Burchfiel, Benjamin and Tedrake, Russ and Song, Shuran},
  journal={The International Journal of Robotics Research},
  pages={02783649241273668},
  year={2023},
  publisher={SAGE Publications Sage UK: London, England}
}

@misc{hilton2023kl,
  author = {Jacob Hilton},
  title = {KL divergence of max-of-n},
  year = {2023},
  url = {https://www.jacobh.co.uk/bon_kl.pdf}
}

@article{lillicrap2015continuous,
  title={Continuous control with deep reinforcement learning},
  author={Lillicrap, Timothy P and Hunt, Jonathan J and Pritzel, Alexander and Heess, Nicolas and Erez, Tom and Tassa, Yuval and Silver, David and Wierstra, Daan},
  journal={arXiv preprint arXiv:1509.02971},
  year={2015}
}
\bibliographystyle{plainnat}


\newpage

\appendix

\section{Theoretical Justification}
\label{appendix:theory}
\begin{proposition}[Q-chunking performs unbiased $n$-step return backup]
\label{thm:theory}
Let $s_t, a_t, \cdots, s_{t+n}$ be a trajectory segment generated by following a data collection policy $\pi_\beta(a_t, \cdots, a_{t+n} | s_t)$ (\emph{i.e.}, $s_{t+k} \sim T(\cdot \mid s_{t+k-1}, a_{t+k-1}), \forall k\in\{1, \cdots, n\}$, and $r_t, r_{t+1}, \cdots r_{t+n-1}$ be the reward received at each corresponding time step (\emph{i.e.}, $r_{t+k} = r(s_{t+k}, a_{t+k})$). Let $V^\pi(s_t)$ be the value for an action chunking policy $\pi(a_t, \cdots, a_{t+n} | s_t)$ starting from state $s_t$. Let $Q^\pi(s_t, a_t, a_{t+1}, \cdots, a_{t+n-1})$ be the Q-value for $\pi$ starting from state $s_t$ and executing the action sequence $a_t, a_{t+1}, \cdots, a_{t+n-1}$.

The $n$-step return estimate $\hat{V}_{\mathrm{n\text{-}step}} := \sum_{t'=t}^{t+n-1}\left[\gamma^{t'-t}r_{t'}\right] + \hat{V}(s_{t+n})$ under the trajectory segment distribution described above is unbiased for $Q^\pi(s_t, a_{t}, \cdots, a_{t+n-1})$ as long as $\hat{V}(s_{t+n})$ is unbiased for $V^\pi(s_{t+n})$.
\end{proposition}

\begin{proof}
From the definition of $Q^\pi(s_t, a_t, a_{t+1}, \cdots, a_{t+n-1})$, we can write it as an expectation: 

\begin{align}
    Q^\pi(s_t, a_t, a_{t+1}, \cdots, a_{t+n-1}) 
    &= \mathbb{E}_{s_{t'} \sim T(\cdot | s_{t'}, a_{t'})} \left[\sum_{t'=t}^{t+n-1}\left[\gamma^{t'-t}r(s_{t'}, a_{t'})\right] + V^\pi(s_{t+n})\right] \\
    &= \mathbb{E}_{s_{t+n}, r_{t+1}, r_{t+2}, \cdots, r_{t+n-1}}\left[\sum_{t'=t}^{t+n-1}\left[\gamma^{t'-t}r_{t'}\right] + V^\pi(s_{t+n})\right] \\
    & = \mathbb{E}_{s_{t+n}, r_{t+1}, r_{t+2}, \cdots, r_{t+n-1}}\left[\sum_{t'=t}^{t+n-1}\left[\gamma^{t'-t}r_{t'}\right] + \hat{V}(s_{t+n})\right] \\
    & = \mathbb{E}\left[\hat{V}_{\mathrm{n\text{-}step}}\right].
\end{align}
The second line uses the fact that the rewards $r_t, r_{t+1}, \cdots, r_{t+h-1}$ are associated with the trajectory segment $s_t, a_t, \cdots, s_{t+n}$. The third line uses the fact that $\hat{V}(s_{t+n})$ is unbiased for $V^\pi(s_{t+n})$. The fourth line uses the definition of $\hat{V}_{\mathrm{n\text{-}step}}$.
\end{proof}

The implication of this theorem is that if we use the $n$-step return backup for the chunked Q-function $Q(s_t, a_t, a_{t+1}, \cdots, a_{t+n-1})$ (with a value estimate of $\hat{V}_{s} \leftarrow Q(s_t, a_t, a_{t+1}, \cdots, a_{t+n-1})$), it converges to the ground truth $Q$-value under the same policy $Q^\pi(s_t, a_t, a_{t+1}, \cdots, a_{t+n-1})$. In contrast, using $n$-step return backup in the original action space does not converge to the correct $Q$-value (\emph{i.e.}, $Q^\pi(s_{t}, a_{t}, a_{t+1}, \cdots, a_{t+n-1})$).

\section{Domain Details}
\label{appendix:domain}
\begin{figure*}[t]
    \centering
    \begin{minipage}{0.24\textwidth}
        \begin{subfigure}{\textwidth}
            \centering
            \includegraphics[width=0.98\textwidth]{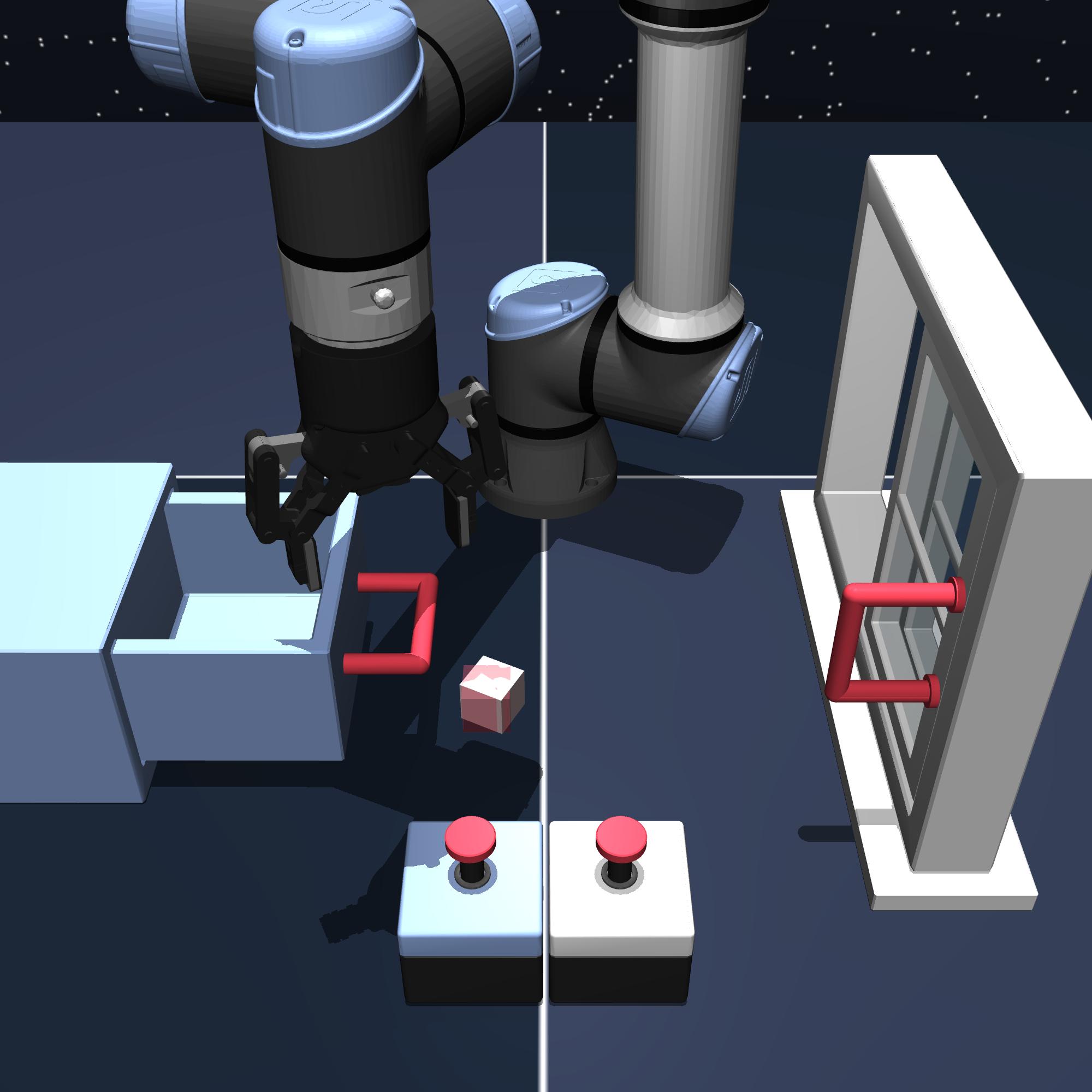}
            \caption{\footnotesize a) \texttt{scene}}
            \label{fig:scene-viz}
        \end{subfigure}
    \end{minipage}\hfill
    \begin{minipage}{0.24\textwidth}
        \begin{subfigure}{\textwidth}
            \centering
            \includegraphics[width=0.98\linewidth]{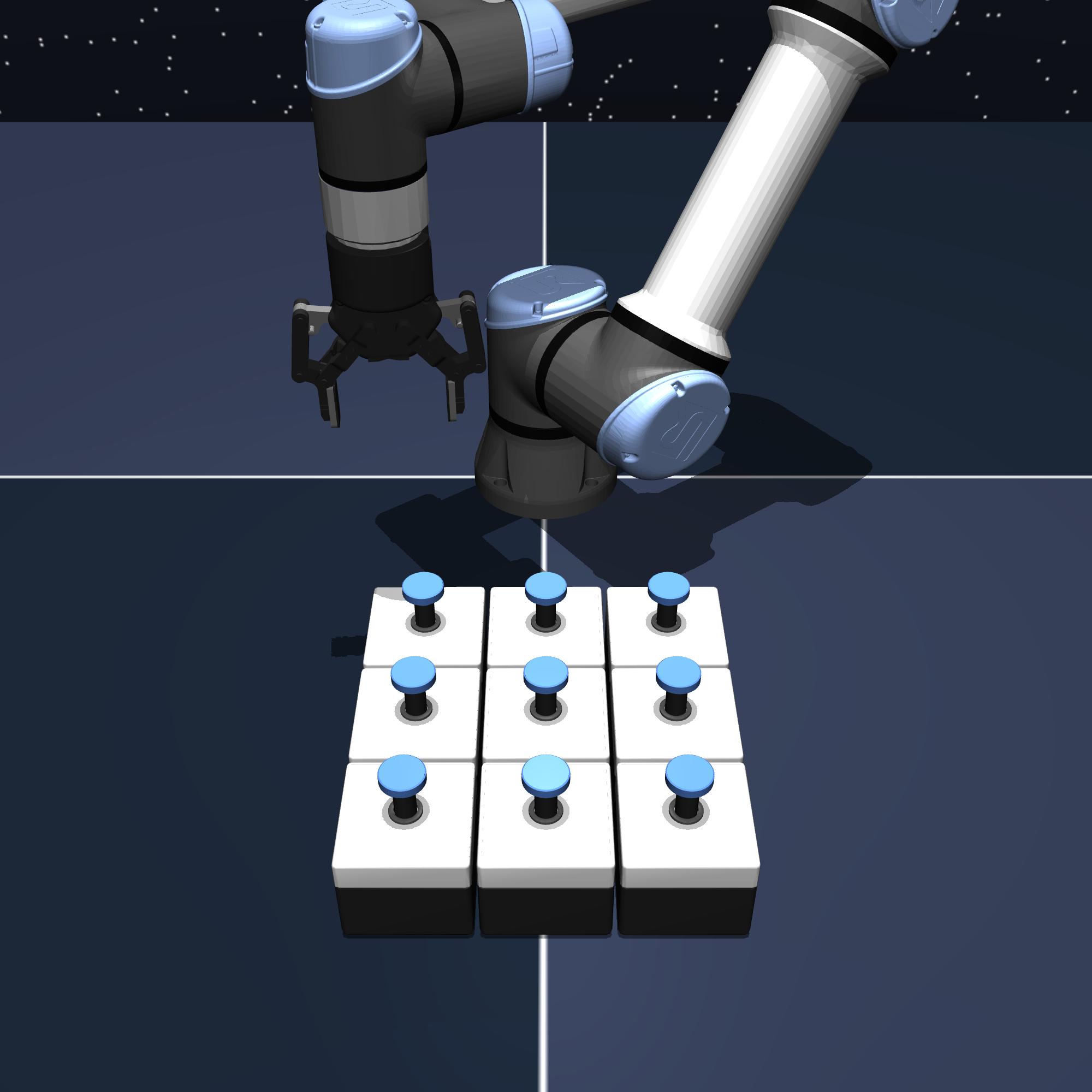}
            \caption{\footnotesize b) \texttt{puzzle-3x3}}
            \label{fig:puzzle-viz}
        \end{subfigure}
    \end{minipage}\hfill    
    \begin{minipage}{0.24\textwidth}
        \begin{subfigure}{\textwidth}
            \centering
            \includegraphics[width=0.98\textwidth]{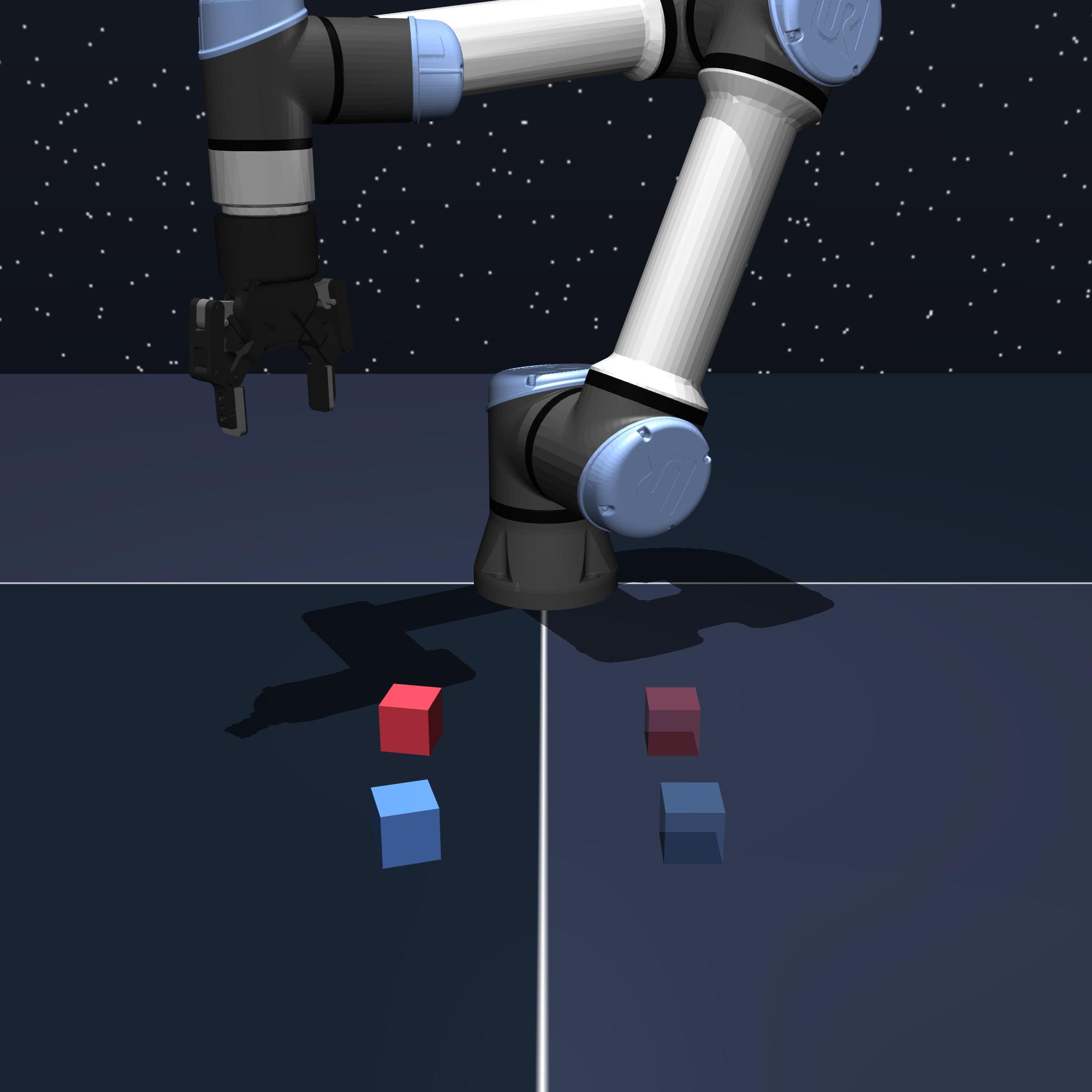}
            \caption{\footnotesize c) \texttt{cube-double}}
            \label{fig:cube-double-viz}
        \end{subfigure}
    \end{minipage}\hfill
    \begin{minipage}{0.24\textwidth}
        \begin{subfigure}{\textwidth}
            \centering
            \includegraphics[width=0.98\linewidth]{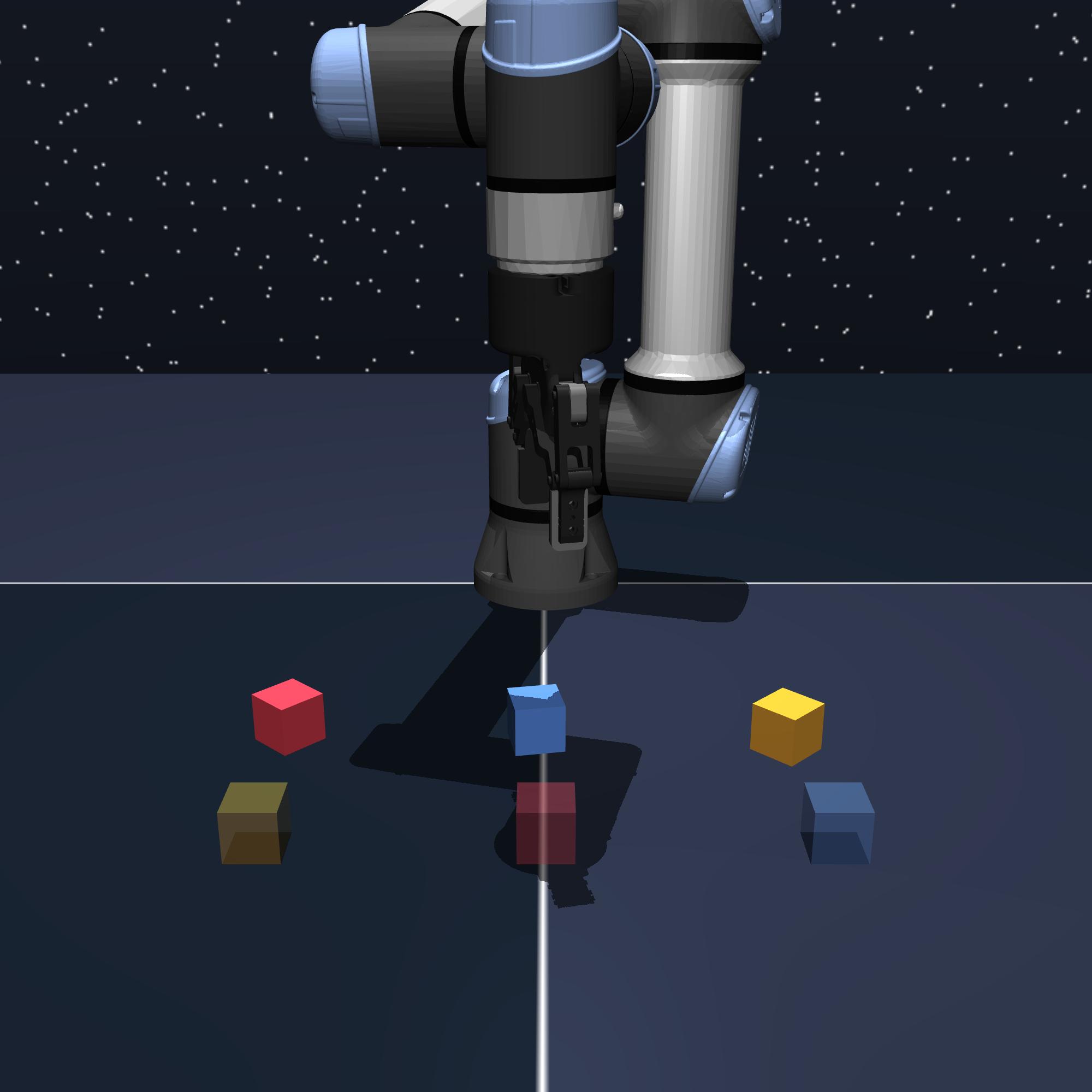}
            \caption{\footnotesize d) \texttt{cube-triple}}
            \label{fig:cube-triple-viz}
        \end{subfigure}
    \end{minipage}
    
    \begin{minipage}{0.24\textwidth}
        \begin{subfigure}{\textwidth}
            \centering
            \includegraphics[width=0.98\linewidth]{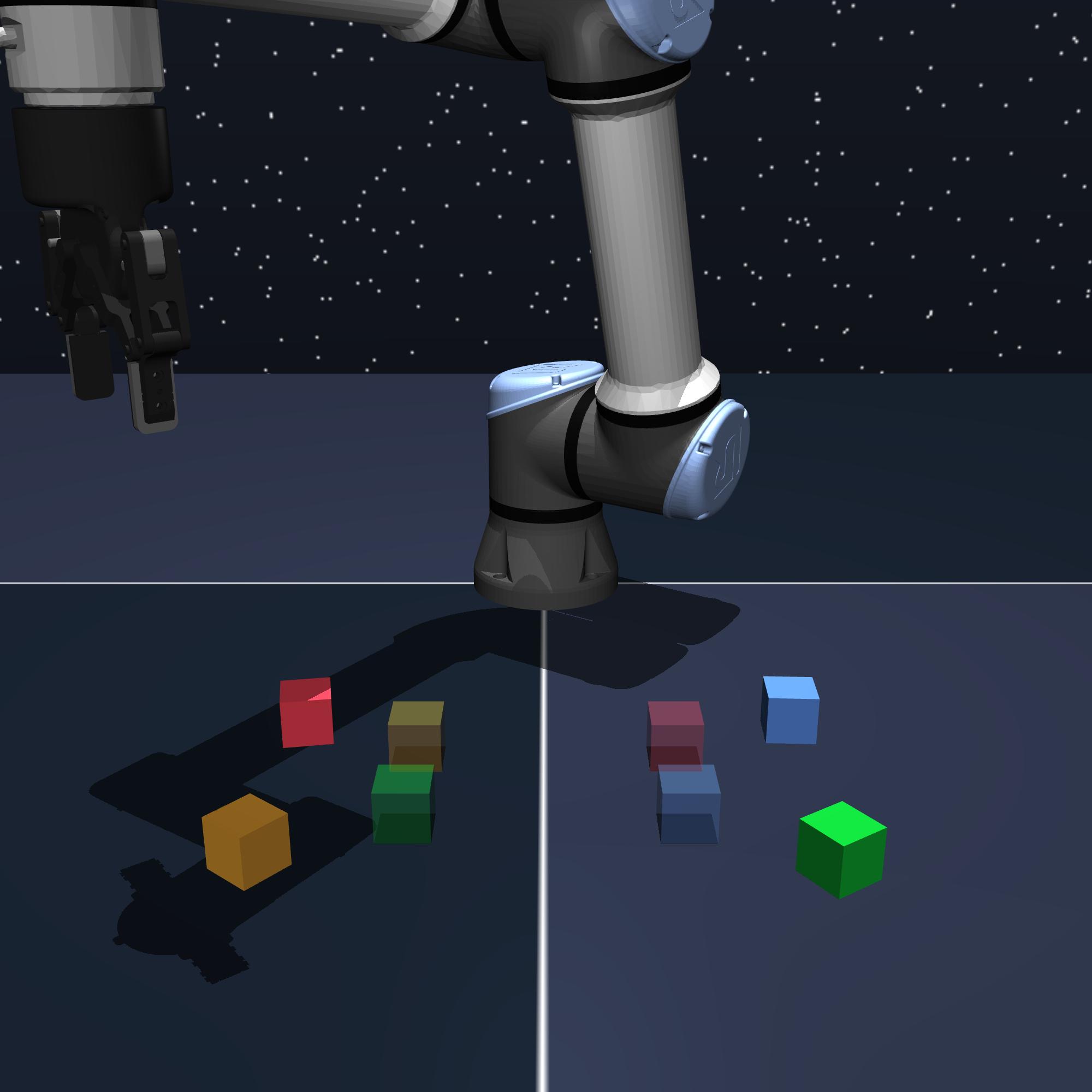}
            \caption{\footnotesize e) \texttt{cube-quadruple}}
            \label{fig:cube-quadruple-viz}
        \end{subfigure}
    \end{minipage}\hfill
    \begin{minipage}{0.24\textwidth}
        \begin{subfigure}{\textwidth}
            \centering
            \includegraphics[width=\linewidth]{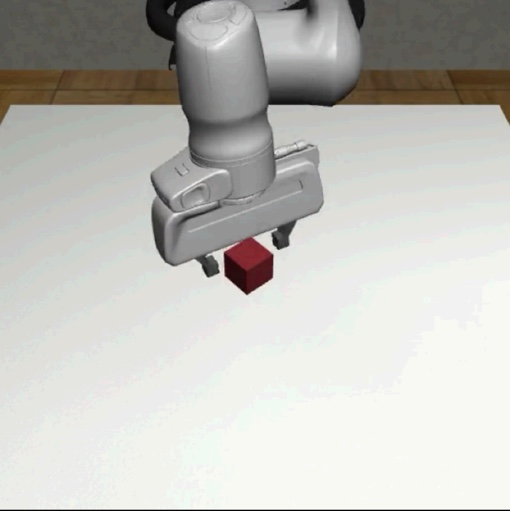}
            \caption{\footnotesize f) \texttt{lift}}
        \end{subfigure}
    \end{minipage}\hfill
    \begin{minipage}{0.24\textwidth}
        \begin{subfigure}{\textwidth}
            \centering
            \includegraphics[width=\linewidth]{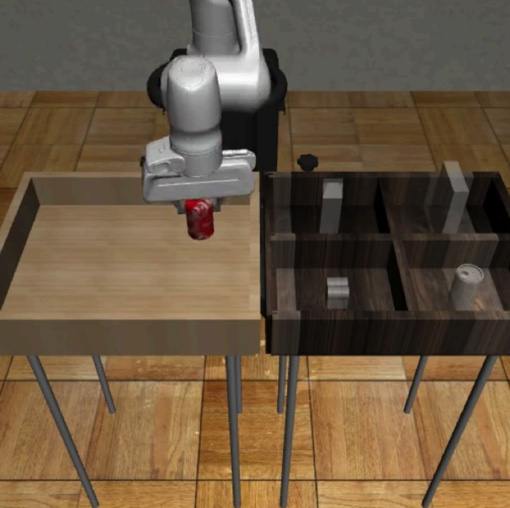}
            \caption{\footnotesize g) \texttt{can}}
        \end{subfigure}
    \end{minipage}\hfill
    \begin{minipage}{0.24\textwidth}
        \begin{subfigure}{\textwidth}
            \centering
            \includegraphics[width=\linewidth]{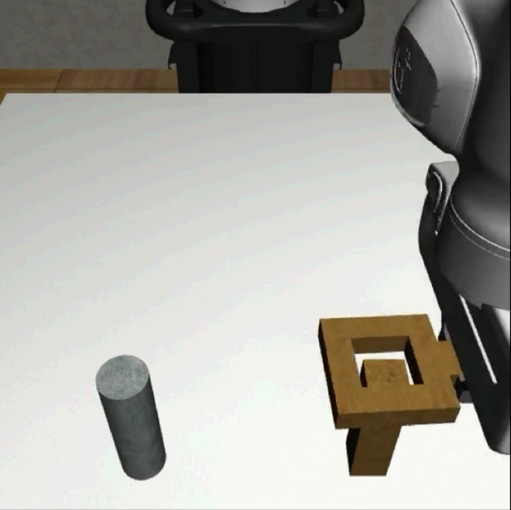}
            \caption{\footnotesize h) \texttt{square}}
        \end{subfigure}
    \end{minipage}\hfill
    \caption{\footnotesize \textbf{We experiment on several challenging long-horizon, sparse-reward domains.} See detailed task description for each domain in Appendix \ref{appendix:domain}. The rendered images of the robomimic tasks above are taken from \citet{robomimic2021}.}
    \label{fig:env_renders}
\end{figure*}

See an overview of the six domains we use in our experiments in \cref{fig:env_renders}. 
We also include the dataset size, episode length and the action dimension in Table \ref{tab:metadata}. In the following sections, we describe each domain in details.

\subsection{OGBench environments.} We consider five manipulation domains from OGBench~\citep{ogbench_park2024} and take the publicly available single-task versions of it in our experiments. For \texttt{scene-sparse} and \texttt{puzzle-3x3-sparse}, we sparsify the reward function such that the reward values are $-1$ when the task is incomplete and $0$ when the task is completed. For \texttt{cube-double/triple/quadruple}, the RL agent needs to command an UR-5 arm to pick and place two/three/four cubes to target locations. In particular, \texttt{cube-triple} and \texttt{cube-quadruple} are extremely difficult to solve from offline data only, and often achieve zero success rate. The RL agent must explore efficiently online in these domains to solve the tasks. The \texttt{cube-*} domains provide a great test ground for sample-efficiency of offline-to-online RL algorithms which we primarily focus on. For \texttt{cube-quadruple}, we use the 100M-size dataset. The dataset is too big to fit our CPU memory, so we periodically (after every 1000 gradient steps) load in a $1$M-size chunk of the dataset for offline training. For online training of \hourblue{RLPD}, \hourpurple{QC-RLPD}, we use the same strategy where we load in a $1$M-size chunk of the dataset as the offline data and perform 50/50 sampling (e.g., 50\% of the data comes from the $1$M-chunk of the offline data, 50\% of the data comes from the online replay buffer). For online fine-tuning of \hourpurple{QC-*}, \hourblue{FQL}, \hourblue{FQL-n}, \hourblue{BFN}, and \hourblue{BFN-n}, we keep a fixed $1$M-size chunk of the offline dataset as the initialization of $\mathcal{D}$ and adds new data to $\mathcal{D}$ directly. The remaining $99$M transitions in the offline data are not being used online. We now describe each of the five domains in details:

\texttt{scene-sparse}: This domain involves a drawer, a window, a cube and two button locks that control whether the drawer and the window can be opened. These tasks typically involve a sequence of actions. For example, \texttt{scene-task2} requires the robotic arm to unlock both locks, move the drawer and the window to the desired position, and then lock both locks. \texttt{scene-task4} requires the robotic arm to unlock the drawer, open the drawer, put the cube into the drawer, close the drawer. The reward is binary: $-1$ if the desired configuration is not yet reached and $0$ if the desired configuration is reached (and the episode terminates).

\texttt{puzzle-3x3-sparse}: This domain contains a $3\times3$ grid of buttons. Each button has two states represented by its color (blue or red). Pressing any button causes its color and the color of all its adjacent buttons to flip (red $\rightarrow$ blue and blue $\rightarrow$ red). The goal is to achieve a pre-specified configuration of colors. \texttt{puzzle-3x3-task2} starts with all buttons to be blue, and the goal is to flip exactly one button (the top-left one) to be red. \texttt{puzzle-3x3-task4} starts with four buttons (top-center, bottom-center, left-center, right-center) to be blue, and the goal is to turn all the buttons to be blue. The reward is binary: $-1$ if the desired configuration is not yet reached and $0$ if the desired configuration is reached (and the episode terminates).

\texttt{cube-double/triple/quadruple}: These three domains contain 2/3/4 cubes respectively. The tasks in the three domains all involve moving the cubes to their desired locations. 
The reward is $-n_{\mathrm{wrong}}$ where $n_{\mathrm{wrong}}$ is the number of the cubes that are at the wrong position. The episode terminates when all cubes are at the correct position (reward is 0).

\begin{table}[t]
    \centering
    \begin{tabular}{@{}cccc@{}}
        \toprule
        \textbf{Tasks} & \textbf{Dataset Size}  & \textbf{Episode Length} & \textbf{Action Dimension ($A$)} \\
        \midrule
        \texttt{scene-sparse-*}       &  $1$M & $750$ & $5$\\
        \texttt{puzzle-3x3-sparse-*}  &  $1$M & $500$ & $5$\\
        \texttt{cube-double-*} &  $1$M & $500$ & $5$\\
        \texttt{cube-triple-*} &  $3$M & $1000$ & $5$\\
        \texttt{cube-quadruple-100M-*} &  $100$M & $1000$ & $5$\\
        \texttt{lift}        &  $31\,127$& $500$ & $7$\\
        \texttt{can}        &  $62\,756$ & $500$ & $7$\\
        \texttt{square}        &  $80\,731$ & $500$ & $7$\\
    \bottomrule
    \end{tabular}
    \vspace{2mm}
    \caption{\footnotesize \textbf{Domain metadata.} Dataset size (number of transitions), episode length, and the action dimension. For OGBench tasks, the action dimension is 5 ($x$ position, $y$ position, $z$ position, gripper yaw and gripper opening). For robomimic tasks, the action dimension is 7 for square to control one arm (3 degree of freedoms (DoF) for translation, 3 DoF for rotation, and one final DoF for the gripper opening).}
    
    \label{tab:metadata}
\end{table}

\subsection{Robomimic environments.} We use three challenging tasks from the robomimic domain~\citep{robomimic2021}.
We use the multi-human datasets that were collected by six human operators. Each dataset contains 300 successful trajectories. The three tasks are as described as follows.
\begin{itemize}
    \item \texttt{lift}: This task requires the robot arm to pick a small cube. This is the simplest task of the benchmark.
    \item \texttt{can}: This task requires the robot arm to pick up a coke can and place in a smaller container bin.
    \item \texttt{square}: This task requires the robot arm to pick a square nut and place it on a rod. The nut is slightly bigger than the rod and requires the arm to move precisely to complete the task successfully.
\end{itemize}

All of the three robomimic tasks use binary task completion rewards where the agent receives $-1$ reward when the task is not completed and $0$ reward when the task is completed.

\section{Implementation Details}
\label{appendix:impl-details}
In this section, we provide more implementation details on both of our Q-chunking methods (QC, QC-FQL) and our baselines used in our experiments. For IFQL, ReBRAC, IQL, we directly use the implementation from \citet{park2025flow}.

\begin{minipage}{0.48\textwidth}
\begin{algorithm}[H]
\caption{{\color{ourpurple}\texttt{QC}}}
\label{algo:qc}
\begin{algorithmic}

\small
\State \textbf{Input:} Behavior policy and critic: $f_\xi(\ac{t}{t+h} | s)$ and $Q_\theta(s_t, \ac{t}{t+h})$.
\State $\mathcal{D} \leftarrow$ offline prior data.
\For{every environment step $t$}
    \If{$t \mod h \equiv 0$}
        \State $\ac{t}{t+h}^1 \cdots \ac{t}{t+h}^N \sim f_\xi(\cdot | s_t)$
        \State $\ac{t}{t+h}^\star \leftarrow \arg\max_{\ac{t}{t+h}^i} Q_\theta(s, \ac{t}{t+h}^i)$
    \EndIf
    \State Act with $a^\star_t$ and receive $s_{t+1}, r_t$.
    \State $\mathcal{D} \leftarrow \mathcal{D} \cup \{(s_t, a^\star_t, s_{t+1}, r_t)\}$
    \State Update $f_\xi$ via flow-matching loss using $\mathcal{D}$ .
    \State Update $Q_\theta$ via Eq (\ref{eq:qc-critic}) using $\mathcal{D}$.
\EndFor
\State \textbf{Output:} $f_\xi, Q_\theta$.
\end{algorithmic}
\end{algorithm}
\end{minipage}
\hfill
\begin{minipage}{0.48\textwidth}
\begin{algorithm}[H]
\caption{{\color{ourpurple}\texttt{QC-FQL}}}
\label{algo:qc-fql}
\begin{algorithmic}
\small
\State \textbf{Input:} Behavior policy, critic, one-step policy: $f_\xi(\ac{t}{t+h} | s)$, $Q_\theta(s_t, \ac{t}{t+h})$, $\mu_\psi(s, \mathbf z)$.
\State $\mathcal{D} \leftarrow$ offline prior data.
\For{every environment step $t$}
    \If{$t \mod h  \equiv 0$}
        \State $\mathbf z \sim \mathcal{N}(0, \mathbf I_{Ah})$
        \State $\ac{t}{t+h} \leftarrow \mu_\psi(s_t, \mathbf z)$
    \EndIf
    \State Act with $a_t$ and receive $s_{t+1}, r_t$.
    \State $\mathcal{D} \leftarrow \mathcal{D} \cup \{(s_t, a_t, s_{t+1}, r_t)\}$
    \State Update $f_\xi$ via flow-matching loss using $\mathcal{D}$.
    \State Update $\mu_{\psi}$ and $Q_\theta$ via Eq. (\ref{eq:fql-onestep}, \ref{eq:fql-critic}) using $\mathcal{D}$.
\EndFor
\State \textbf{Output:} $f_\xi, Q_\theta, \mu_\psi(s, \mathbf z)$.
\end{algorithmic}
\end{algorithm}
\end{minipage}

\subsection{\hourpurple{QC-FQL}}
We build the implementation of our method on top of \hourblue{FQL}~\citep{park2025flow}, a recently proposed offline RL/offline-to-online RL method that uses TD3+BC-style objective. It is implemented with a one-step noise-conditioned policy (instead of a Gaussian policy that is commonly used in RL) and it uses a distillation loss from a behavior flow-matching policy as the BC loss. To adapt this method to use action chunking, we simply apply \hourblue{FQL} on the temporally extended action space -- the behavior flow-matching policy generates a sequence of actions, the one-step noise-conditioned policy predicts a sequence of actions, and the $Q$-network also takes in a state and a sequence of actions. More concretely, we train three networks: 
\begin{enumerate}
    \item $Q_\theta(s, a_1, \cdots a_h): \mathcal{S}\times \mathcal{A}^h \mapsto \mathbb{R}$ --- the value function that takes in a state and a sequence of actions (action chunk). In practice, we train an ensemble of $Q$ networks. We denote the weight of the ensemble element as $\theta = (\theta_1, \cdots, \theta_K)$.
    \item $\mu_\psi(s, z): \mathcal{S}\times \mathbb{R}^{Ah} \mapsto \mathbb{R}^{Ah}$ --- the one-step noise-conditioned policy that takes in a state and a noise, and outputs a sequence of actions conditioned on them.
    \item $f_\xi(s, m, u): \mathcal{S} \times \mathbb{R}^{Ah} \times [0, 1] \mapsto \mathbb{R}^{Ah}$ --- the flow-matching behavior policy parameterized by a velocity prediction network. The network predicts takes in a state, an intermediate state of the flow and a time, and outputs the velocity direction that the intermediate action sequence should move in at the specified time. See \cref{algo:flow-step} for more details on how this velocity prediction network is used to generate an action from a noise vector.
\end{enumerate}
We denote our policy as $\pi_\psi(\cdot|s)$. It is implemented by first sampling a Gaussian noise $z \sim \mathcal{N}(0, I_{Ah})$ and run it through the one-step noise-conditioned policy $\begin{bmatrix} a_1 & \cdots & a_h \end{bmatrix} \leftarrow \mu_\psi(s, z)$. To train these three networks, we sample a high-level transition, $w = (s_t, a_t, a_{t+1}, \cdots, a_{t+h-1}, s_{t+h}, R^h_t) \sim \mathcal{D}$ where $R^h_t = \sum_{t'=0}^{h-1}\gamma^{t'}r_{t+t'}$, to construct the following losses:

\textbf{(1) Critic loss:}
\begin{align}
    L(\theta_k, w) = \left(Q_{\theta_k}(s_t, a_t, \cdots, a_{t+h-1}) - R^h_t - \frac{\gamma^h}{K}\sum_{k'=1}^{K} Q_{\bar \theta_{k'}}(s_{t+h}, a_{t+h}, \cdots, a_{t+2h-1})\right)^2, 
\end{align}
where $\begin{bmatrix} a_{t+h} & \cdots & a_{t+2h-1} \end{bmatrix} \sim \pi_\psi(\cdot | s_{t+h}), k \in \{1, 2, \cdots, K\}$.

\textbf{(2) Actor loss:}
\begin{align}
    L(\psi, w) = -Q_\theta(s_t, \mu_\psi(s_t, z_t)) + \alpha \left\|\mu_\psi(s_t, z_t) - \begin{bmatrix} a^\xi_t & \cdots & a^\xi_{t+h-1} \end{bmatrix}\right\|_2^2,
\end{align}
where $z_t \sim \mathcal{N}(0, I_{Ah})$ and  $\begin{bmatrix} a^\xi_t & \cdots & a^\xi_{t+h-1} \end{bmatrix}$ is the result of running the behavior policy $f_\xi(s, m, t)$ with \cref{algo:flow-step} from $z_t$, and $\alpha \in \mathbb{R}$ is a tunable parameter that controls the strength of the behavior regularization (higher $\alpha$ leads to stronger behavior regularization).

\textbf{(3) Flow-matching behavior policy loss:}
\begin{align}
    L(\xi, w) = \|f_\xi(s_t, u\begin{bmatrix} a_t & \cdots & a_{t+h-1} \end{bmatrix} + (1-u)z_t, u) - (\begin{bmatrix} a_t & \cdots & a_{t+h-1} \end{bmatrix} - z_t)\|_2^2,
    \label{eq:fql-flow}
\end{align}
where $u \sim U([0, 1])$, $z_t \sim \mathcal{N}(0, I_{Ah})$.

Practically, we sample a batch of transitions $\{w_1, w_2, \cdots, w_M\}$ and optimize the average loss for each network: $\mathcal{L}(\theta) = \frac{1}{M}\sum_{i=1}^M \sum_{k=1}^K \mathcal{L}(\theta_k, w_i), \mathcal{L}(\psi) = \frac{1}{N}\sum_{i=1}^M  \mathcal{L}(\psi, w_i), \mathcal{L}(\xi) = \frac{1}{N}\sum_{i=1}^M \mathcal{L}(\xi, w_i)$.

\subsection{\hourblue{FQL}}
\hourblue{FQL}~\citep{park2025flow} is a recently proposed offline RL/offline-to-online RL method that uses TD3+BC-style objective. It is equivalent to our method with $h=1$. For completeness, we write out the objectives for a transition sample $w = (s_t, a_t, r_t, s_{t+1})$:
\begin{align}
    L(\theta_k, w) &= \left(Q_{\theta_k}(s_t, a_t) - r_t - \frac{\gamma}{K}\sum_{k'=1}^{K} Q_{\bar \theta_{k'}}(s_{t+1}, \mu_\psi(s_{t+1}, z_t^{k'}))\right)^2, z^{k'}_t \sim \mathcal{N}(0, I_A), \\
    L(\psi, w) &= -Q_\theta(s_t, \mu_\psi(s_t, z_t)) + \alpha \left\|\mu_\psi(s_t, z_t) - a^\xi_t\right\|_2^2, \\
    & a_t^\xi \leftarrow \mathtt{FlowODE\_Euler}(s_t, z_t, f_\xi, T), z_t \sim\mathcal{N}(0, I_A), \\
    L(\xi, w) &= \|f_\xi(s_t, ua_t + (1-u)z_t, u) - (a_t - z_t)\|_2^2, z_t \sim \mathcal{N}(0, I_A), u \sim U([0, 1]).
\end{align}
In practice, we sample a batch of transitions $\{w_1, w_2, \cdots, w_N\}$ and optimize the average loss for each network: $\mathcal{L}(\theta) = \frac{1}{N}\sum_{i=1}^N \sum_{k=1}^K \mathcal{L}(\theta_k, w_i), \mathcal{L}(\psi) = \frac{1}{N}\sum_{i=1}^N  \mathcal{L}(\psi, w_i), \mathcal{L}(\xi) = \frac{1}{N}\sum_{i=1}^N \mathcal{L}(\xi, w_i)$.

\subsection{\hourblue{FQL-n}}
To implement the $n$-step return baseline, we take \hourblue{FQL} and replace the 1-step TD update with the $h$-step TD update:
\begin{align}
    L(\theta_k, w) &= \left(Q_{\theta_k}(s_t, a_t) - \sum_{t'=0}^{h-1}(\gamma^{t'}r_{t+t'}) - \frac{\gamma^h}{K}\sum_{k'=1}^{K} Q_{\bar \theta_{k'}}(s_{t+h}, \mu_\psi(s_{t+h}, z_t^{k'}))\right)^2,
\end{align}
where $z^{k'}_t \sim \mathcal{N}(0, I_A)$ for all $k' \in \{1, 2, \cdots, K\}$. The actor loss and flow-matching loss remain the same as FQL.

\subsection{\hourpurple{QC}}
The flow-matching behavior policy is trained with the same loss as used in \hourpurple{QC-FQL} (\cref{eq:fql-flow}). On top of the flow-matching behavior policy, we simply parameterize the policy $\pi$ implicitly by sampling multiple action chunks from the behavior policy and pick the one that maximizes the $Q$-value. Specifically, let $\{z_t^1, z^2_t, \cdots, z^N_t\} \sim \mathcal{N}(0, I_{Ah})$ and
\begin{align*}
    \begin{bmatrix}a_t^i & \cdots & a_{t+h-1}^{i}  \end{bmatrix} = \mathtt{FlowODE\_Euler}(s_t, z^i_t, f_\xi, T)
\end{align*}
The policy outputs the one action chunk out of $N$ that maximizes the $Q$-value, $\begin{bmatrix}a_t^{i^\star} & \cdots & a_{t+h-1}^{i^\star}  \end{bmatrix}$, where
\begin{align*}
     i^\star \leftarrow \arg\max_{i \in [N]} Q(s, \begin{bmatrix}a_t^i & \cdots & a_{t+h-1}^{i}  \end{bmatrix}).
\end{align*}
Finally, we directly use this implicitly parameterize policy to generate actions for computing the TD target for our TD loss:
\begin{align}
    L(\theta_k, w) = \left(Q_{\theta_k}(s_t, a_t, \cdots, a_{t+h-1}) - R^h_t - \frac{\gamma^h}{K}\sum_{k'=1}^{K} Q_{\bar \theta_{k'}}(s_{t+h}, a^{i^\star}_{t+h}, \cdots, a^{i^\star}_{t+2h-1})\right)^2
    \label{eq:bfn-critic}
\end{align}
where $a^{i^\star}_{t+h}, \cdots, a^{i^\star}_{t+2h-1} \sim \pi(\cdot | s_{t+h})$.

The baselines \hourblue{BFN-n} and \hourblue{BFN} are implemented similarly to \hourblue{FQL-n} and \hourblue{FQL} by operating in the original action space.

\begin{algorithm}[t]
  \caption{$\mathtt{FlowODE\_Euler}(s_t, z_t, f_\xi, T)$: generate actions from the behavior flow policy $f_\xi(s, m, u)$ with Euler's method.}\label{algo:flow-step}
  \begin{algorithmic}
\State \textbf{Input:} State $s_t$, noise $z_t$ and flow model $f_\xi(s, m, u)$, number of flow steps $T$.
\State $m^0 \leftarrow z_t$
\For{$i \in \{1, \cdots,  T\}$}
    \State $m^i \leftarrow f_\xi(s_t, m^{i-1}, (i-1)/T)$
\EndFor
\State \textbf{Output:} $m^T$.
\end{algorithmic}
\end{algorithm}

\subsection{\hourblue{RLPD}, \hourblue{RLPD-AC}, \hourpurple{QC-RLPD}}
All the \hourblue{RLPD} baseline results are obtained by running the official codebase (as linked in \citet{ball2023efficient}) with additional modification to incorporate action chunking and behavior cloning. This baseline runs online RL from scratch using off-policy transitions where 50\% of them come from the offline dataset and the other 50\% come from the online replay buffer. It essentially up-weights the online data more, allowing the online RL agent to learn more quickly. This is different from how \hourpurple{QC-*}, \hourblue{BFN}, \hourblue{BFN-n}, \hourblue{FQL}, \hourblue{FQL-n} samples off-policy transitions (where we sample from the dataset that combines the offline dataset and online replay buffer data with no weighting). \hourblue{RLPD} baselines all use Gaussian policy. This is also different from our method as our method uses noise-conditioned policy that can represent a wider range of distributions. For \hourblue{RLPD-AC}, we change all the actor and critic networks such that they work with an action chunk rather than a single action. The baseline is exactly the same as our method except that actor and the critic are updated the same as how \hourblue{RLPD} updates its actor and critic. For \hourpurple{QC-RLPD}, we add a behavior cloning loss in the actor loss as follows (highlighted in red below):
\definecolor{rlpdred}{HTML}{FFDFDD}
\begin{align}
    \mathcal{L}(\psi) = \mathbb{E}_{a'_t \sim  \pi_\psi(\cdot | s_t)}\left[-\frac{1}{K}\sum_{k=1}^{K}Q_{\theta_k}(s_t, a'_t) \mathcolorbox{rlpdred}{- \alpha \log \pi_\psi(a_t | s_t)}\right].
\end{align}

\subsection{\hourblue{SUPE-GT}}
For this baseline, we adapt from a recently proposed skill-based offline-to-online RL method~\citep{wilcoxson2024leveraging}.
The original method has additional modules such reward models and random distillation networks to
deal with the problem setting where offline dataset is unlabeled (\emph{e.g.}, reward information is missing).
We remove these two modules as we have ground truth reward information in our offline dataset.
The method has two stages: (1) skill-pretraining and (2) online RL with skills. In the first stage, the
baseline uses a trajectory VAE to learn latent-conditioned skill policy. In the second stage, the baseline
uses RLPD to learn a high-level policy that outputs latent vector to command the latent-conditioned
skill policy after every $h$ steps. Such a skill-based method in theory also allows temporally coherent
actions for exploration and value backup speedup.

\section{Experiment Details}
\label{appendix:experiment-details}

\subsection{Computational resources}
We use NVIDIA RTX-A5000 GPU to run all our experiments. Each complete offline-to-online experiment run takes around 4-7 hours. To reproduce all our results in \cref{tab:full-main-o2o}, we estimate that it would take around $\underbrace{6}_{\text{hours per single run}} \times \underbrace{15}_{\text{\# of methods}} \times \underbrace{25}_{\text{\# of tasks}} \times \underbrace{4}_{\text{\# of seeds}} = 9\ 000 \text{ GPU hours}$. Each robomimic experiment run takes longer (around 8-12 hours). While each single run takes longer, we only run a subset of strong methods/baselines on robomimic tasks. We estimate it would take around $\underbrace{10}_{\text{hours per single run}} \times \underbrace{8}_{\text{\# of methods}} \times \underbrace{3}_{\text{\# of tasks}} \times \underbrace{5}_{\text{\# of seeds}} = 1\ 350 \text{ GPU hours}$. In total, we estimate that it would take around $10\ 350$ GPU hours to reproduce all the main results in our paper.

\subsection{Evaluation protocol}
Unless specified otherwise, for all methods, we run 4 seeds on each OGBench task and 5 seeds on each robomimic task. All plots use 95\% confidence interval with stratified sampling (5000 samples). The success rate is computed by running the policy in the environment for 50 episodes and record the number of times that the policy succeeds at solving the task (and divide it by 50).

\subsection{Hyperparameter tuning}
\begin{table}[ht]
    \centering
    \begin{tabular}{@{}cc@{}}
        \toprule
        \textbf{Parameter} & \textbf{Value} \\
        \midrule
        Batch size ($M$) &  $256$ \\
        Discount factor ($\gamma$) & $0.99$ \\
        Optimizer & Adam \\
        Learning rate & $3 \times 10^{-4}$ \\
        Target network update rate ($\tau$) & $5 \times 10^{-3}$ \\
        \multirow{2}{*}{Critic ensemble size ($K$)} & 10 for \hourblue{RLPD}, \hourblue{RLPD-AC}, \hourpurple{QC-RLPD}, and \hourblue{SUPE-GT} \\& 2 for \hourpurple{QC-FQL}, \hourblue{FQL}, \hourblue{FQL-n}, \hourpurple{QC-BFN}, \hourblue{BFN}, \hourblue{BFN-n} \\
        UTD Ratio & $1$ \\
        Number of flow steps ($T$) & $10$ \\
        Number of offline training steps & $10^6$ except \hourblue{RLPD}-based approaches ($0$) \\
        Number of online environment steps & $1 \times 10^6$ \\
        Network width & $512$ \\
        Network depth & $4$ hidden layers \\
    \bottomrule
    \end{tabular}
    \vspace{2mm}
    \caption{\footnotesize \textbf{Common hyperparameters.}}
    \label{tab:rl-hyperparams}
\end{table}

\begin{table}[h]
    \centering
    \begin{tabular}{@{}ccccc@{}}
    \toprule
            Environments                   & \hourblue{FQL} & \hourblue{FQL-n} & \hourpurple{QC-FQL} & \hourblue{ReBRAC} \\
    \midrule
        \texttt{scene-sparse-*}               & 300 & 100 & 300 & 0.1\\
        \texttt{puzzle-3x3-sparse-*}          & 100 & 100 & 300 & 0.1 \\
        \texttt{cube-double-*}         & 300 & 100 & 300 & 0.1 \\
        \texttt{cube-triple-*}         & 300 & 100 & 100 & 0.1 \\
        \texttt{cube-quadruple-100M-*} & 300 & 100 & 100 & 0.1 \\
        \texttt{lift}              & 10000& 10000& 10000 & -\\
        \texttt{can}              & 10000& 10000& 10000 & - \\
        \texttt{square}              & 10000 & 10000& 10000 & -\\
    \bottomrule
    \end{tabular}
    \vspace{2mm}
    \caption{\footnotesize \textbf{Behavior regularization coefficient ($\alpha$).} }
    \label{tab:alpha}
\end{table}

\begin{table}[h]
    \centering
    \scalebox{1.0}{
    \begin{tabular}{@{}cccc@{}}
    \toprule
            Environments                   & BFN & BFN-n & QC-BFN \\
    \midrule
        \texttt{scene-sparse*}                 & 4 & 4 & 32 \\
        \texttt{puzzle-3x3-sparse-*}           & 4 & 4 & 64 \\
        \texttt{cube-*}                        & 4 & 4 & 32 \\
        \texttt{lift, can, square}             & 4& 4& 16\\
    \bottomrule
    \end{tabular}}
    \vspace{2mm}
\caption{\footnotesize \textbf{Number of actions sampled for the expected-max $Q$ operator ($N$) for BFN methods.} }
    \label{tab:emaq}
\end{table}
\hourpurple{QC}, \hourblue{BFN}, \hourblue{BFN-n}. We tune the number of actions sampled, $N$, for the expected-max $Q$ operator. On OGBench domains, we sweep over $\{2, 4, 8, 16, 32, 64, 128\}$ and select the best parameter for each domain and for each method on \texttt{task2}. We report the performance of each method with the best $\alpha$ in \cref{tab:main-o2o} and \cref{fig:banner} (on all tasks). \cref{tab:emaq} summarizes the $\alpha$ value we use for each task. 

\hourpurple{QC-FQL}, \hourblue{FQL}, \hourblue{FQL-n}. We tune the behavior regularization coefficient $\alpha$. On OGBench domains, we take the default hyperparameter of \hourblue{FQL} for each domain $\alpha_{\mathrm{default}}$ and tune all methods on \texttt{task2} of each domain with three choices of $\alpha$: $\{\alpha_{\mathrm{default}} / 3, \alpha_{\mathrm{default}}, 3\alpha_{\mathrm{default}} \}$ (our $\alpha_{\mathrm{default}}$ comes from Table 6 in \citet{park2025flow}). On robomimic domain, we sweep over much large $\alpha$ values: $\{100, 1000, 10000\}$. We report the performance of each method with the best $\alpha$ in  in \cref{tab:main-o2o} and \cref{fig:banner} (on all tasks). \cref{tab:alpha} summarizes the $\alpha$ value we use for each task. 

\hourblue{RLPD}, \hourblue{RLPD-AC}, \hourpurple{QC-RLPD}. We sweep over (1) whether or not to use clipped double Q-learning (CDQ), and (2) whether or not to use entropy backup. We find that not using CDQ and not using entropy backup to perform the best for all of the RLPD baselines and use that across all domains. Even though our method and the other FQL baselines use $K=2$ critic ensemble size, we use $K=10$ critic ensemble size for \hourblue{RLPD} to keep it the same as the hyperparameter in the original paper~\citep{ball2023efficient}. 
For \hourpurple{QC-RLPD}, we sweep over behavior regularization coefficient $\alpha \in \{0.001, 0.01, 0.1\}$ and pick $0.01$ since it works the best.

\hourblue{SUPE-GT}. We tune the KL coefficient for the VAE skill-pretraining from $\{0.001,0.003,0.01,0.03,0.1\}$ and pick $0.003$ as it works the best.

\hourblue{ReBRAC}. We tune the behavior regularization coefficient from $\{0.01,0.03,0.1,0.3,1.0\}$ and pick $0.1$ as it works the best.

\hourblue{IFQL}. We directly use the default hyperparameter from \citet{park2025flow} ($N=32, \tau=0.9$) for this baseline. The first parameter, $N=32$, means that for both online policy rollout and policy evaluation, we sample 32 actions and pick the action that has the highest Q-value. The second parameter, $\tau =0.9$ is the expectile coefficient.

\hourblue{IQL}. We directly use the default hyperparameter from \citet{park2025flow} for this baseline. For \texttt{cube-*}, we use $\alpha=0.3$. For \texttt{scene} and \texttt{puzzle-3x3}, we use $\alpha=10.0$. For the expectile coefficient, we use $\tau = 0.9$.

\section{Full Results}
\label{appendix:results}

\subsection{End-effector visualization}
\begin{figure}[H]
    \centering
\includegraphics[width=0.32\linewidth]{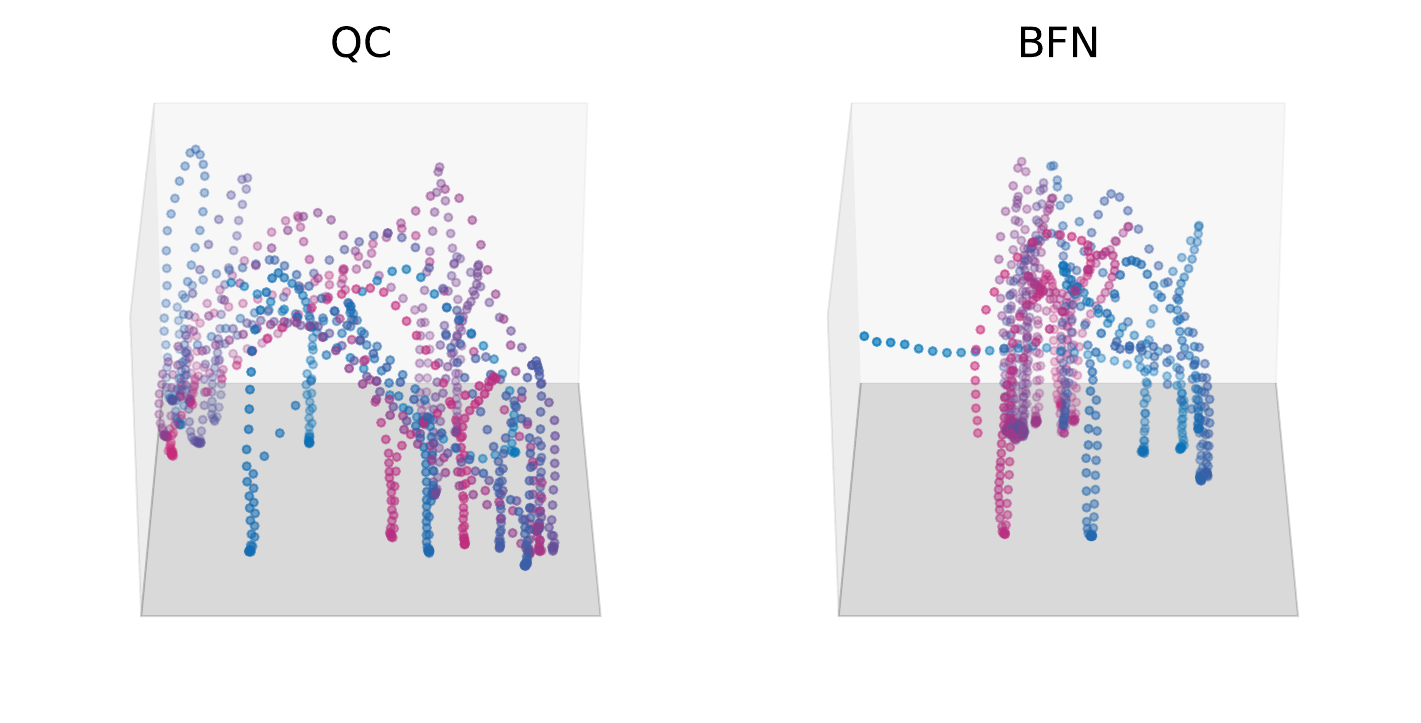}
\includegraphics[width=0.32\linewidth]{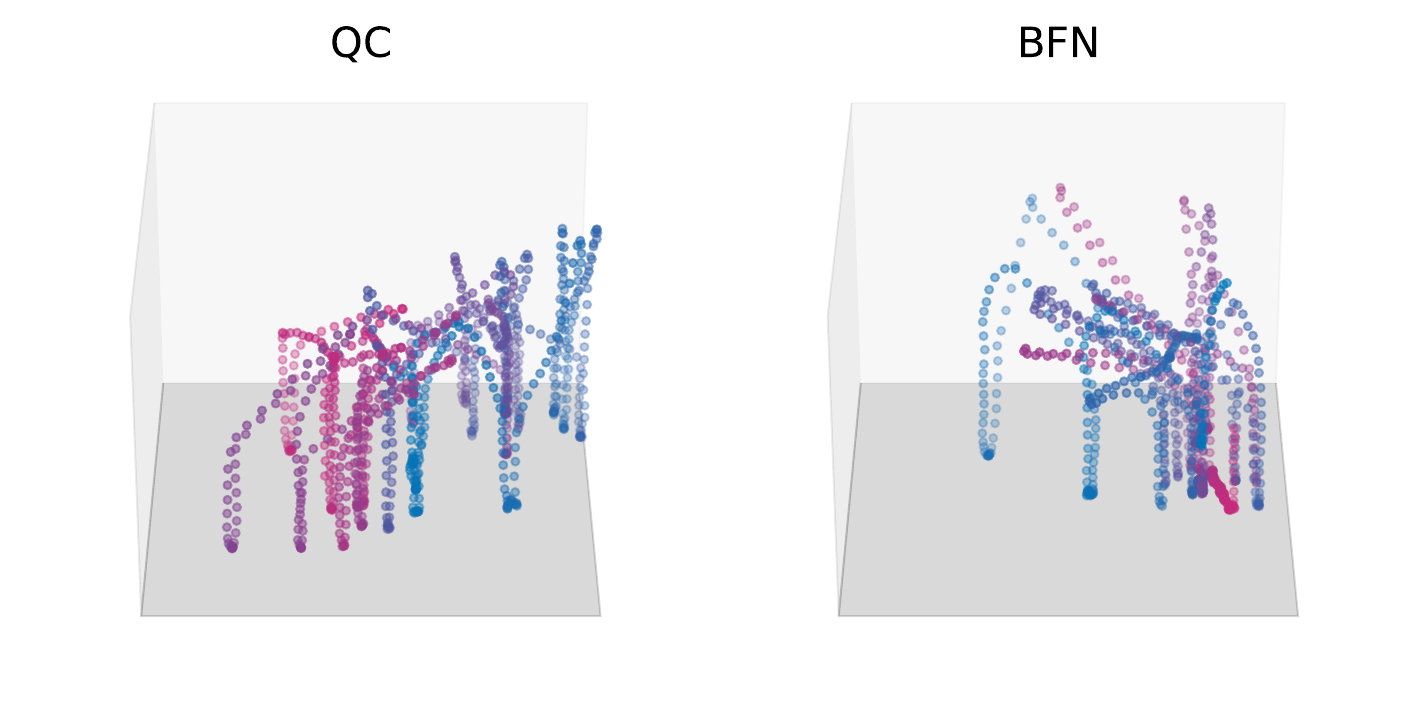}
\includegraphics[width=0.32\linewidth]{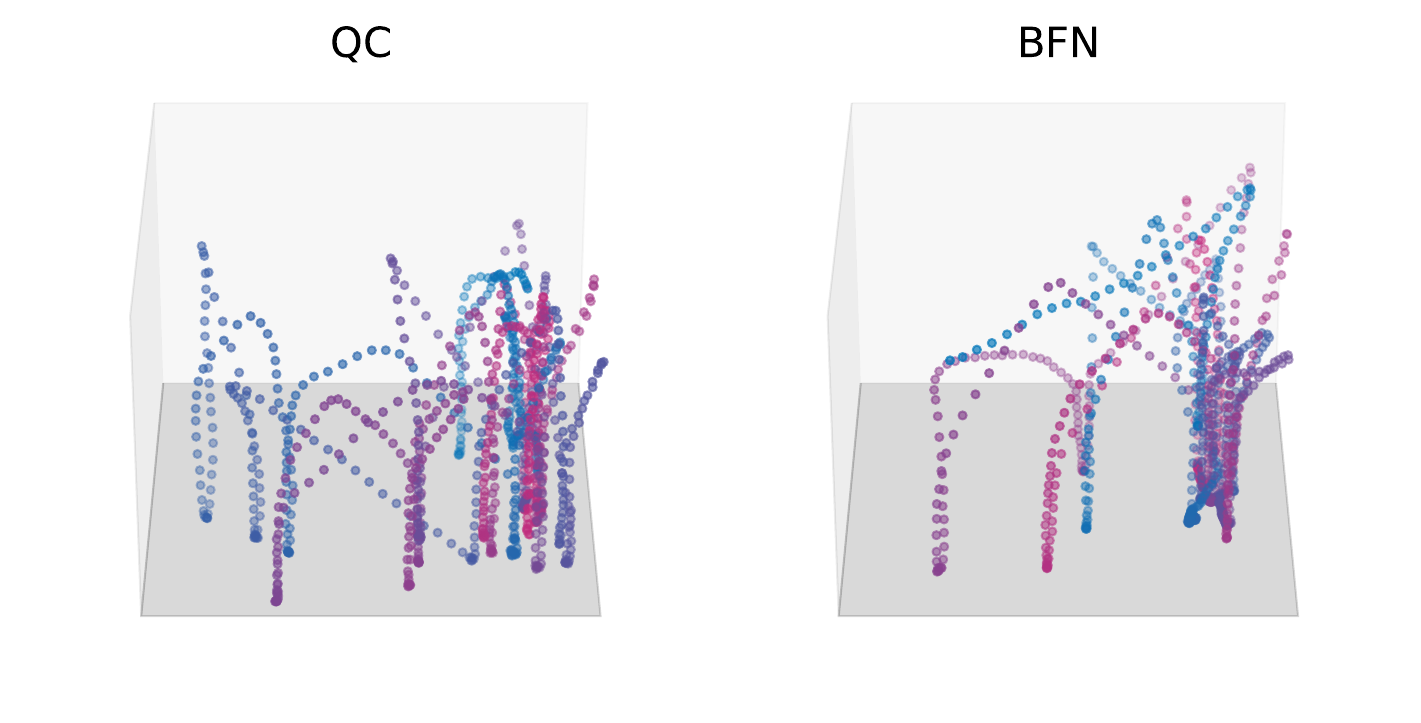}
\includegraphics[width=0.32\linewidth]{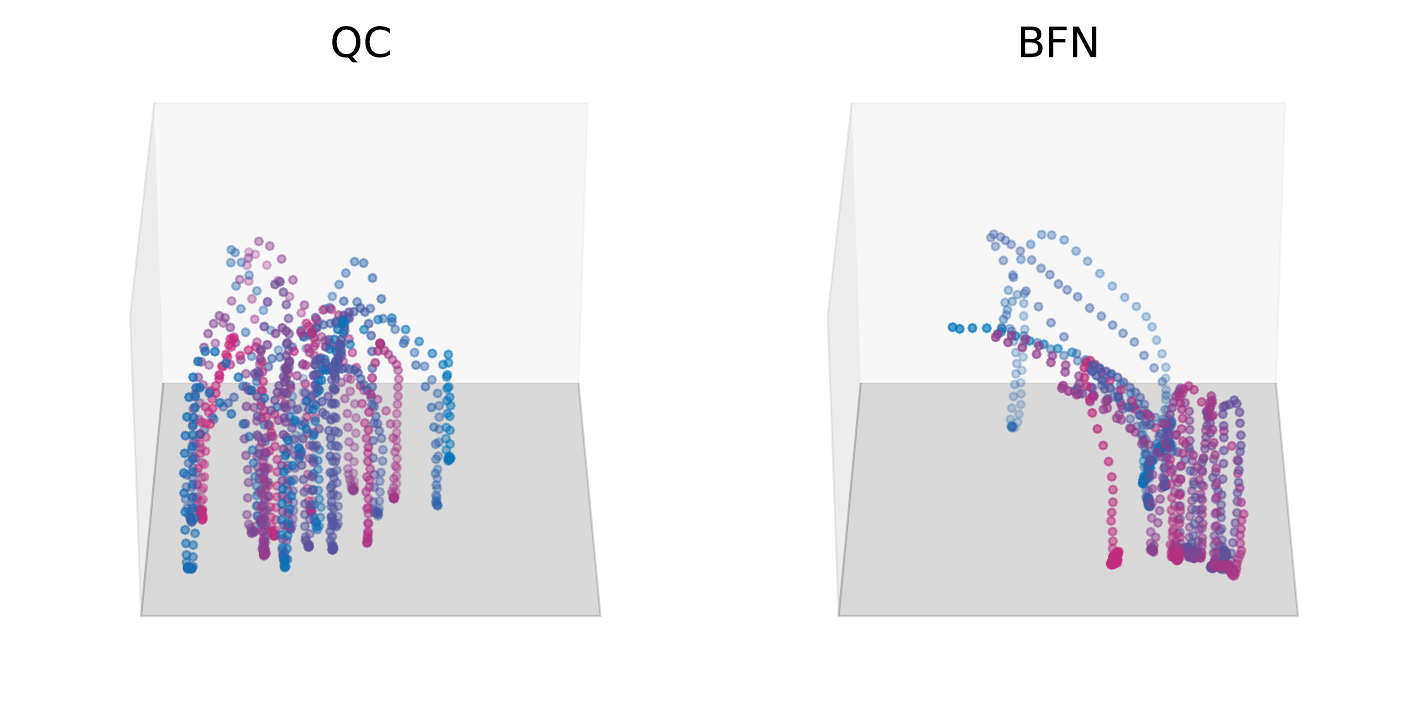}
\includegraphics[width=0.32\linewidth]{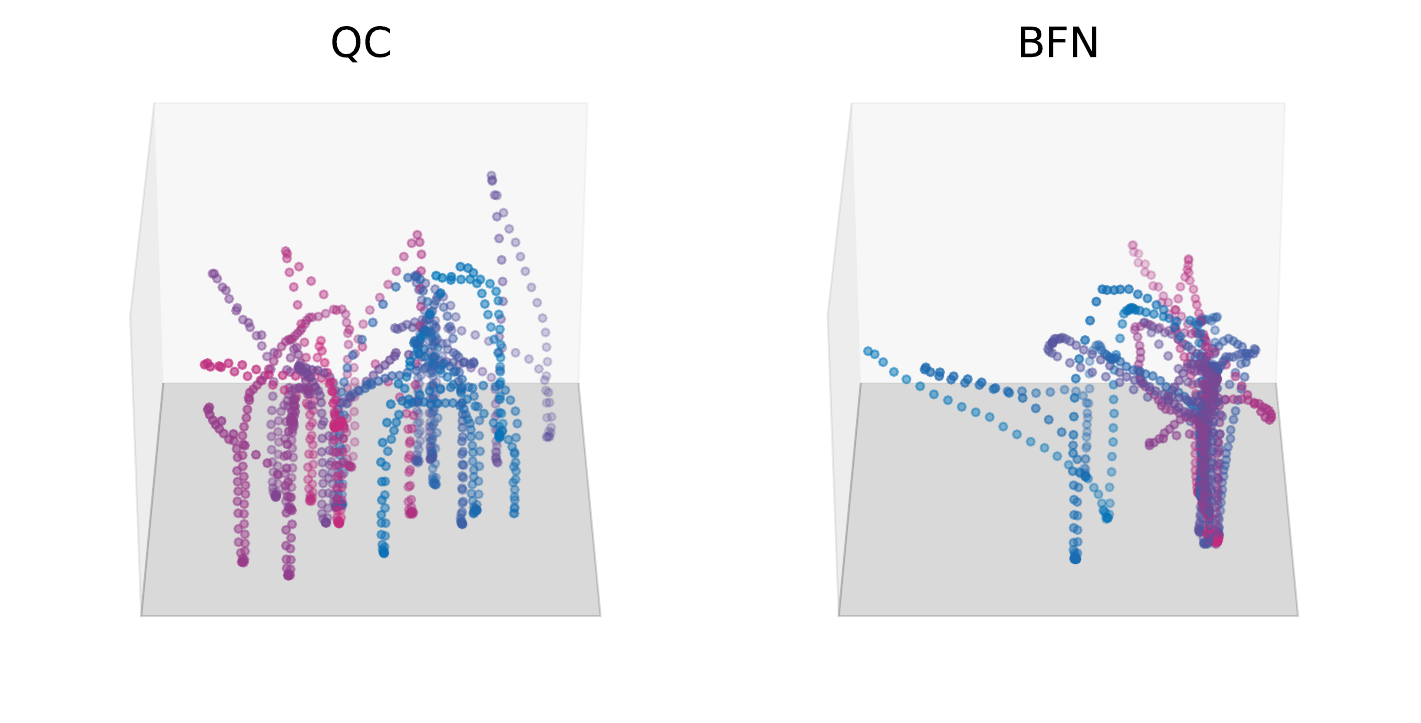}
\includegraphics[width=0.32\linewidth]{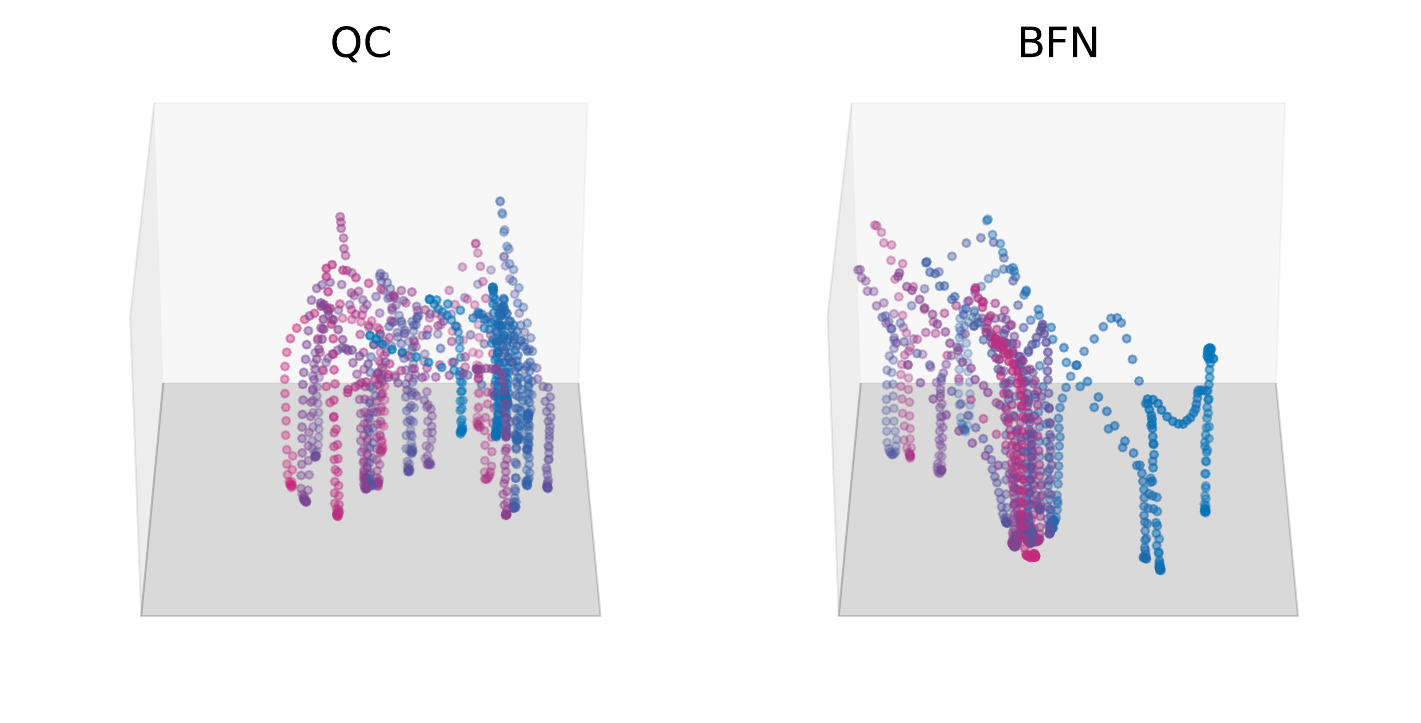}
\includegraphics[width=0.32\linewidth]{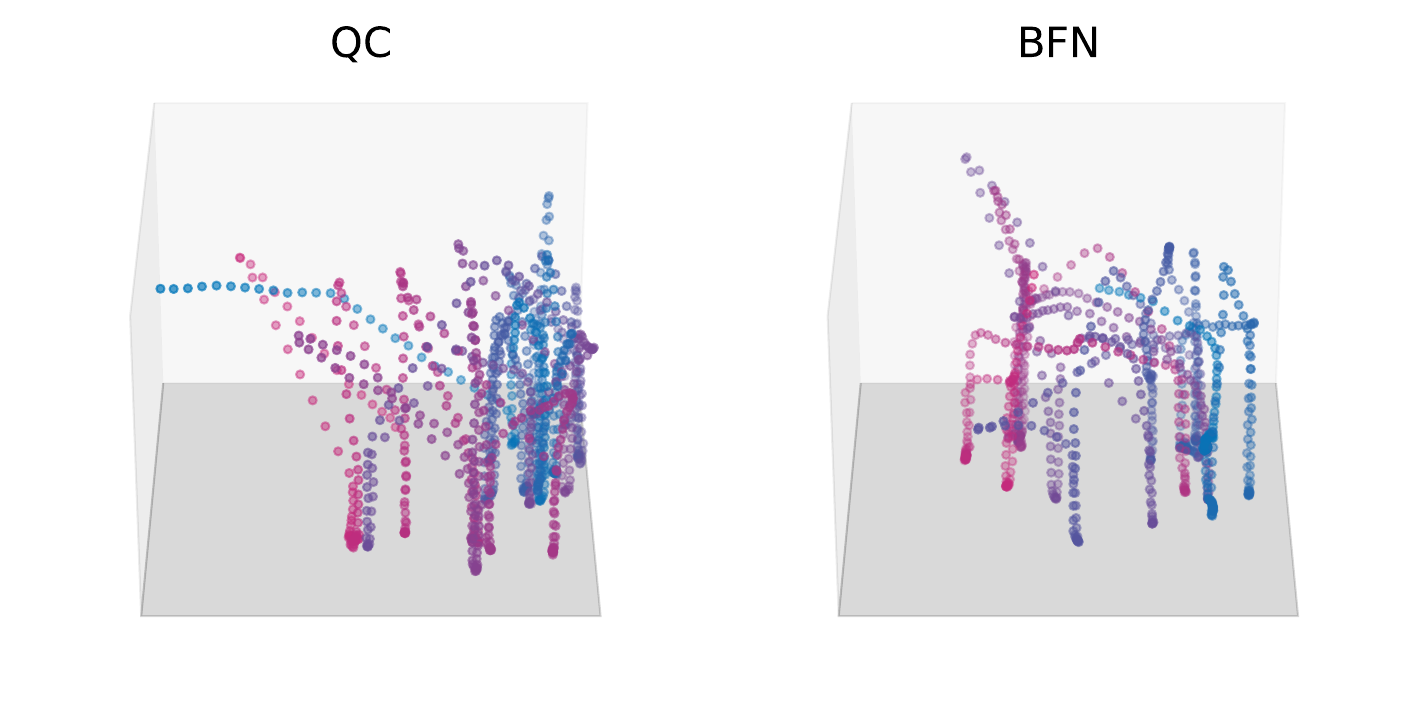}
\includegraphics[width=0.32\linewidth]{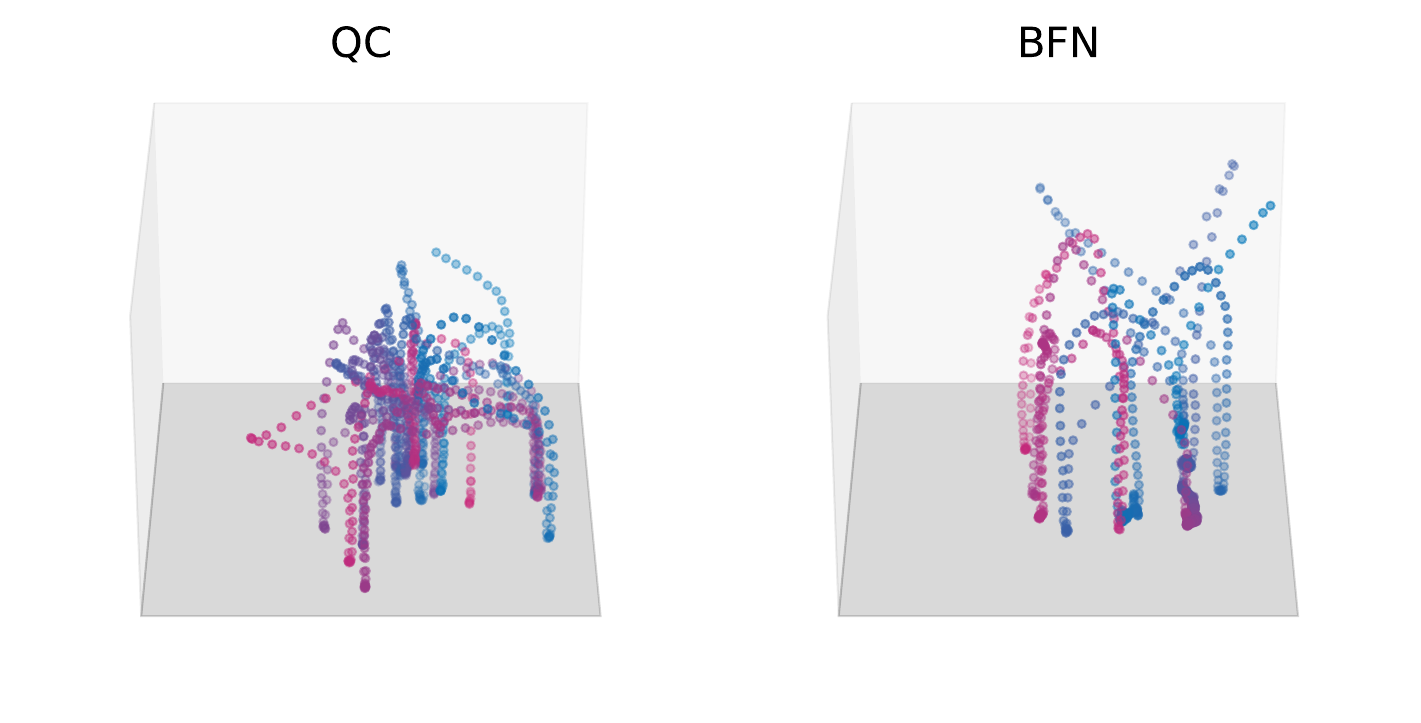}
\includegraphics[width=0.32\linewidth]{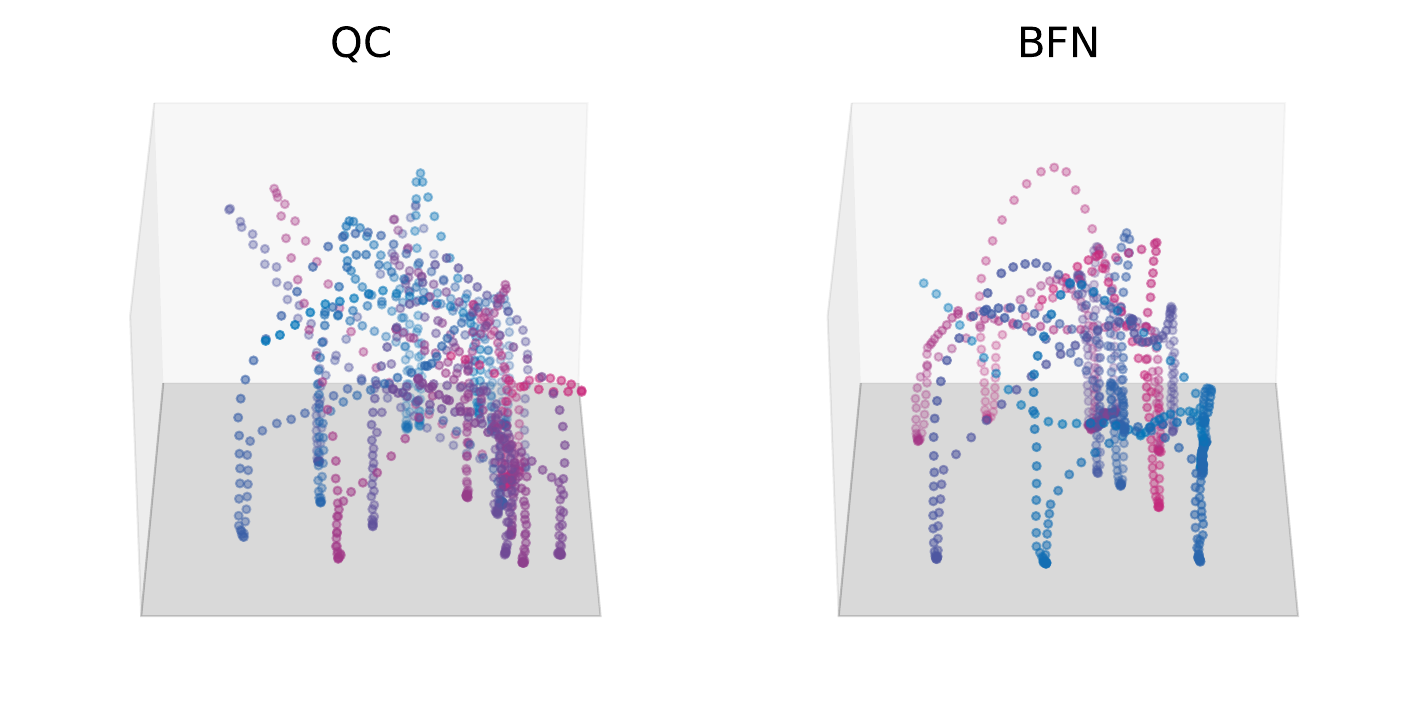}
\caption{\textbf{End-effector trajectory early in the training.} Each subplot above shows the trajectory for a consecutive of 1000 time steps. We include up to Step 9000.}
\label{fig:traj-viz-early}
\end{figure}

\begin{figure}[H]
    \centering
\includegraphics[width=0.32\linewidth]{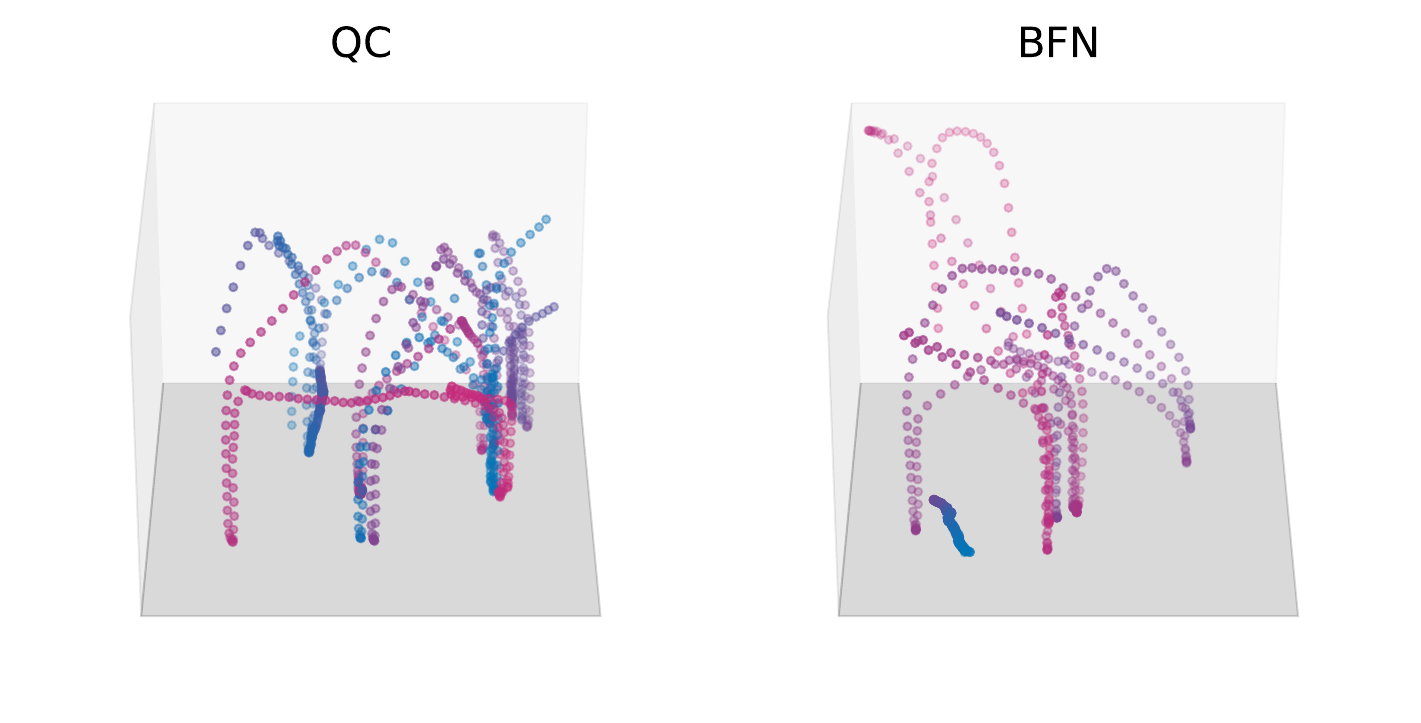}
\includegraphics[width=0.32\linewidth]{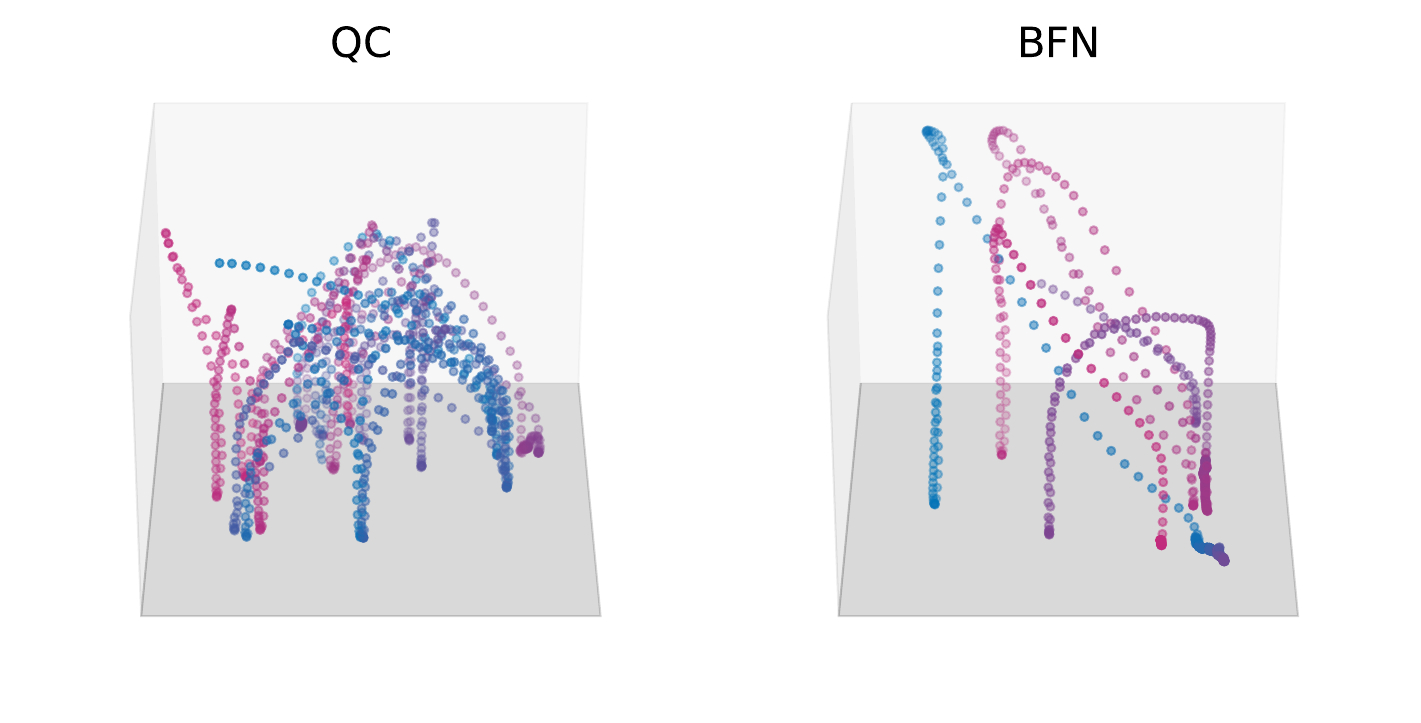}
\includegraphics[width=0.32\linewidth]{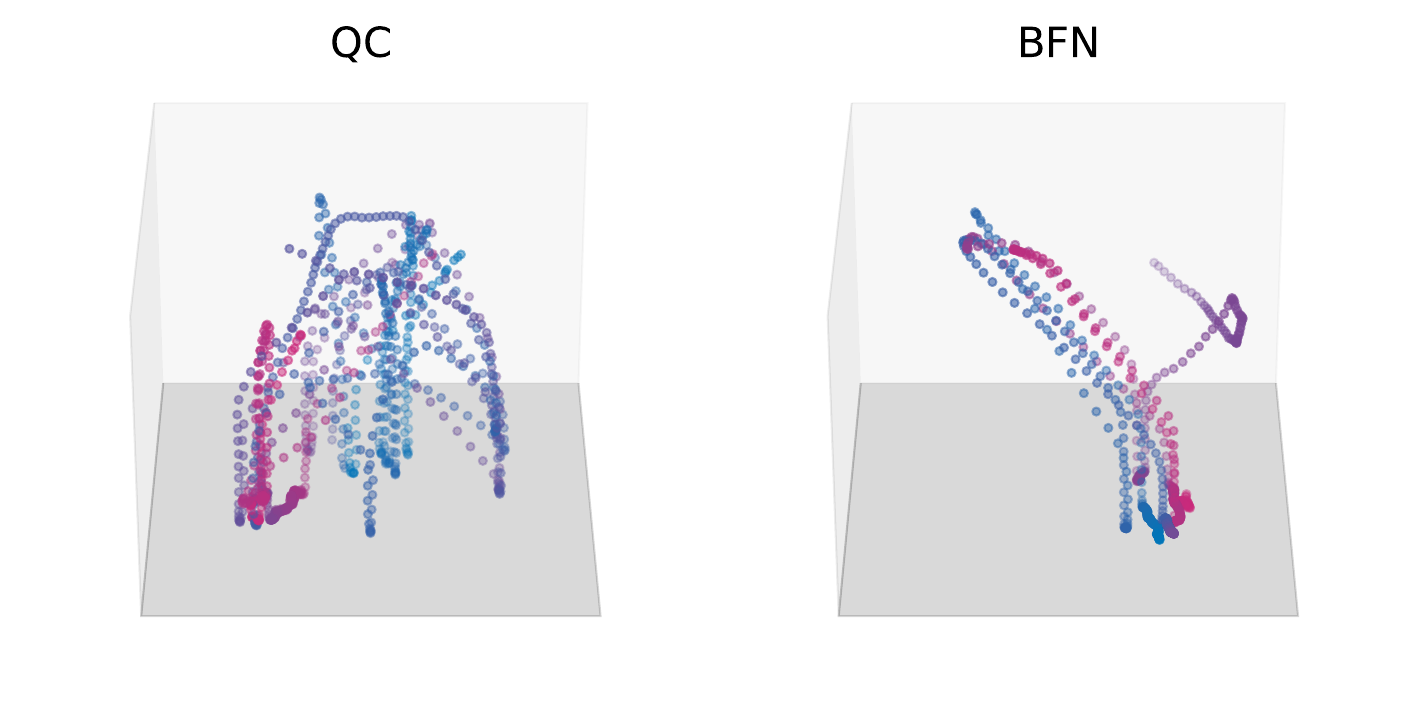}
\includegraphics[width=0.32\linewidth]{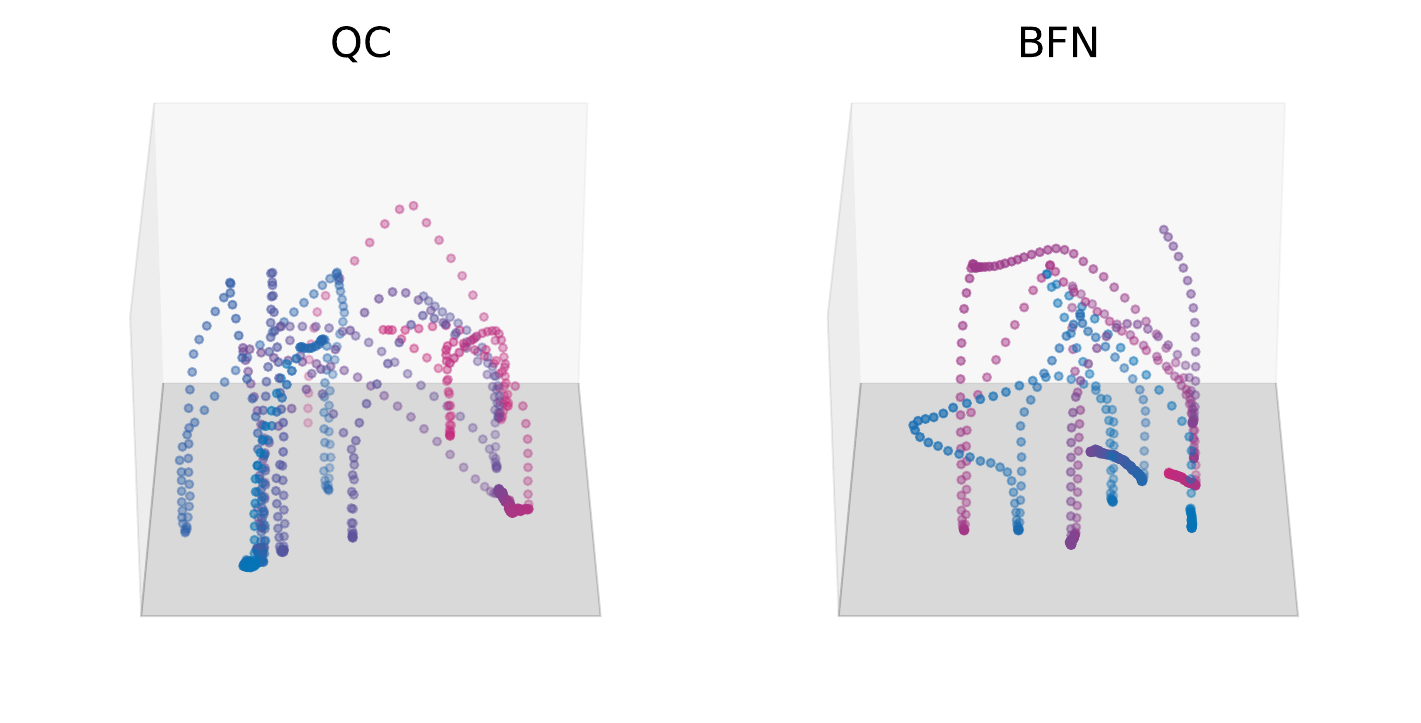}
\includegraphics[width=0.32\linewidth]{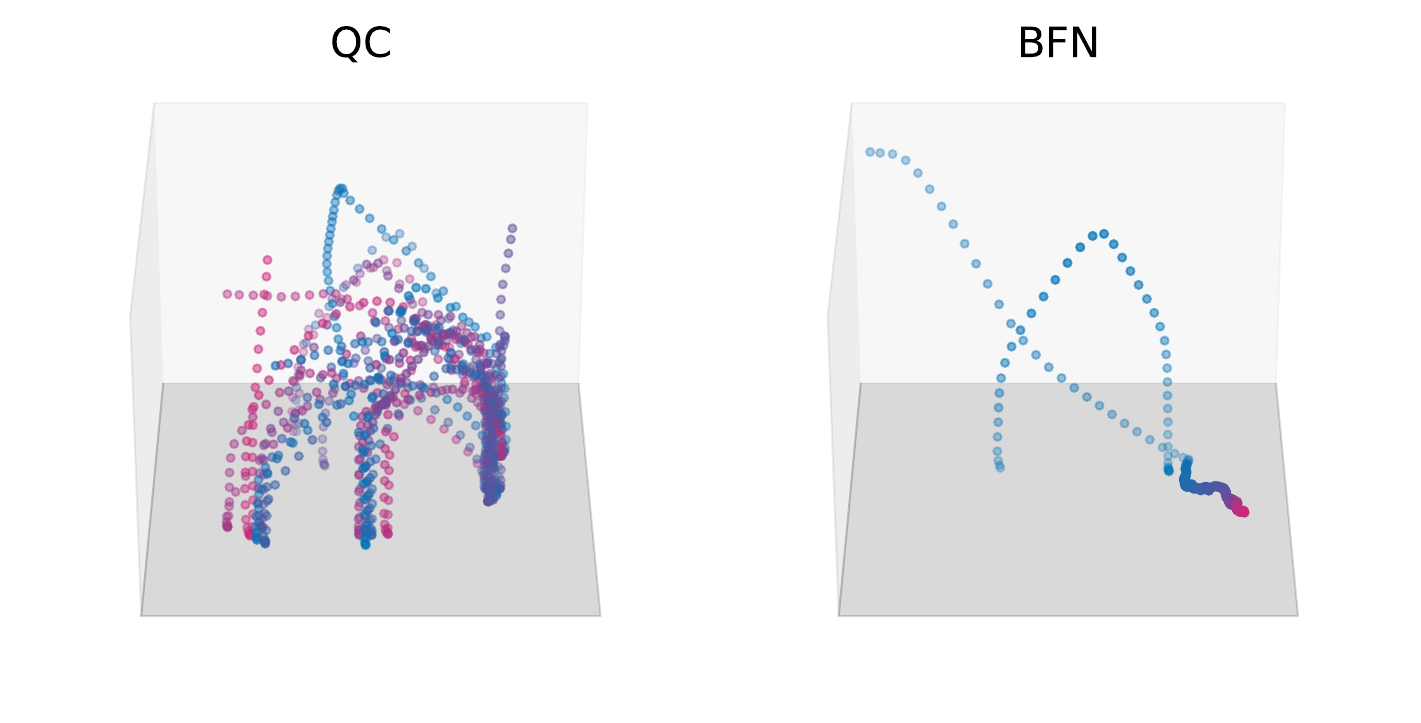}
\includegraphics[width=0.32\linewidth]{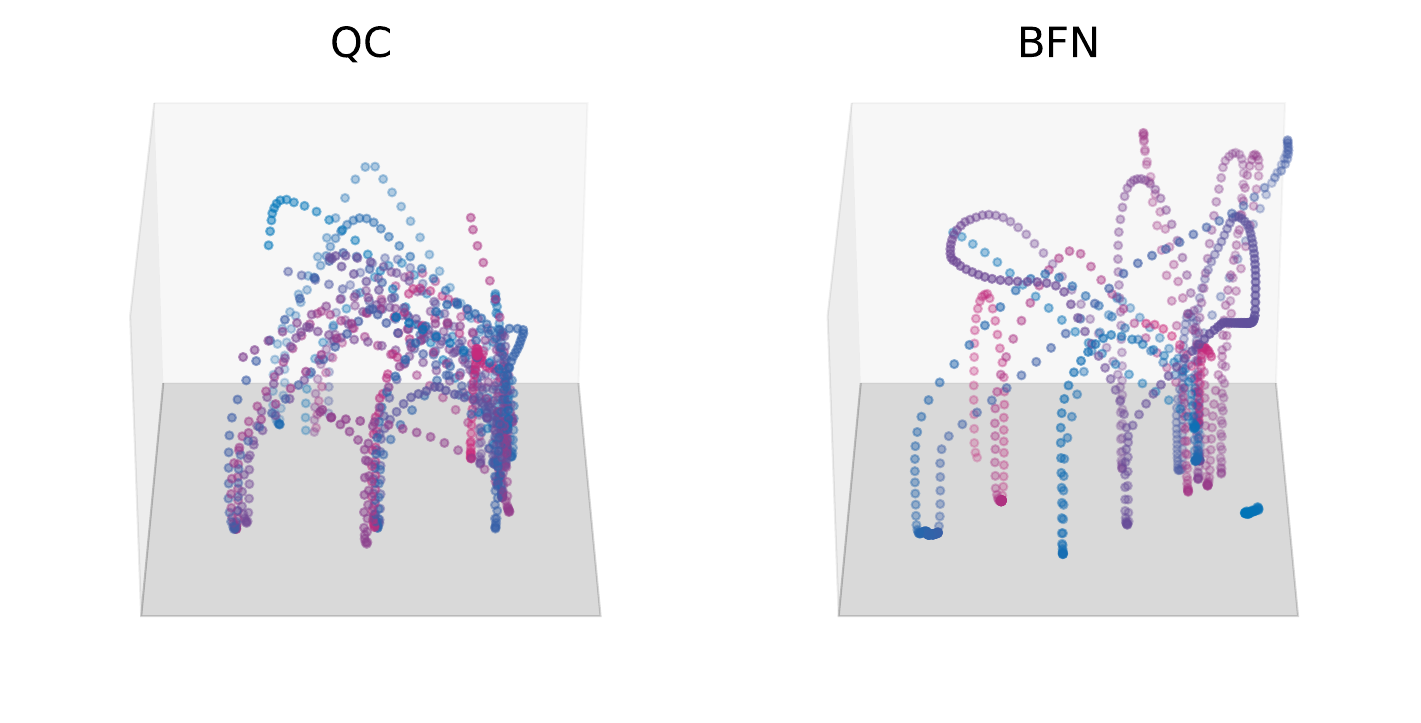}
\includegraphics[width=0.32\linewidth]{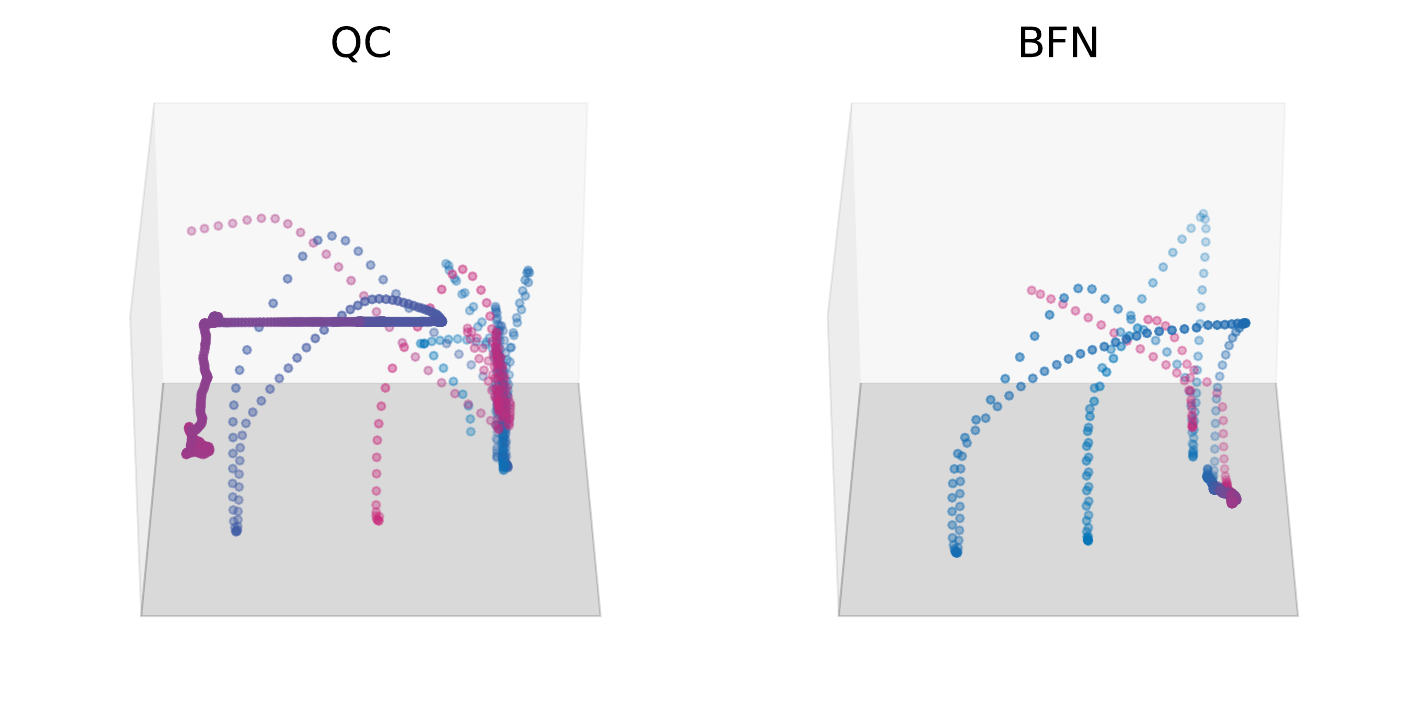}
\includegraphics[width=0.32\linewidth]{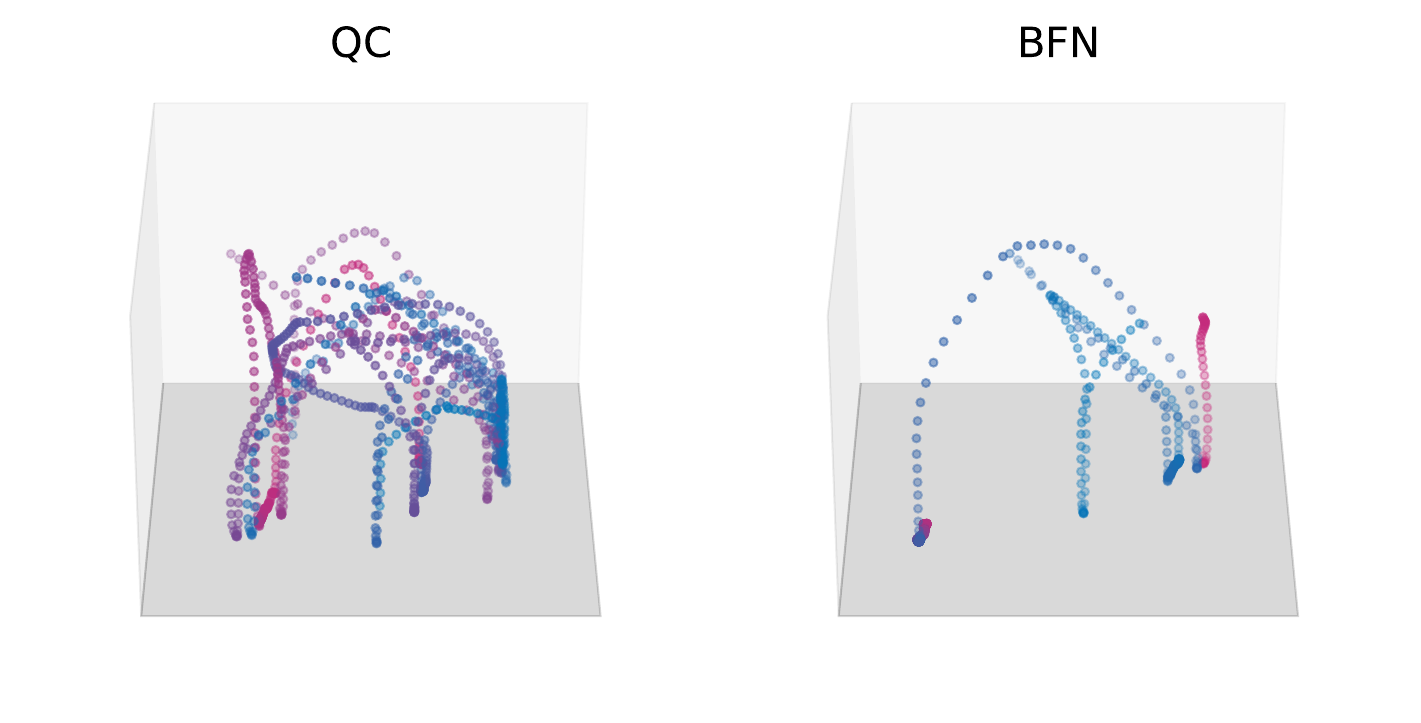}
\includegraphics[width=0.32\linewidth]{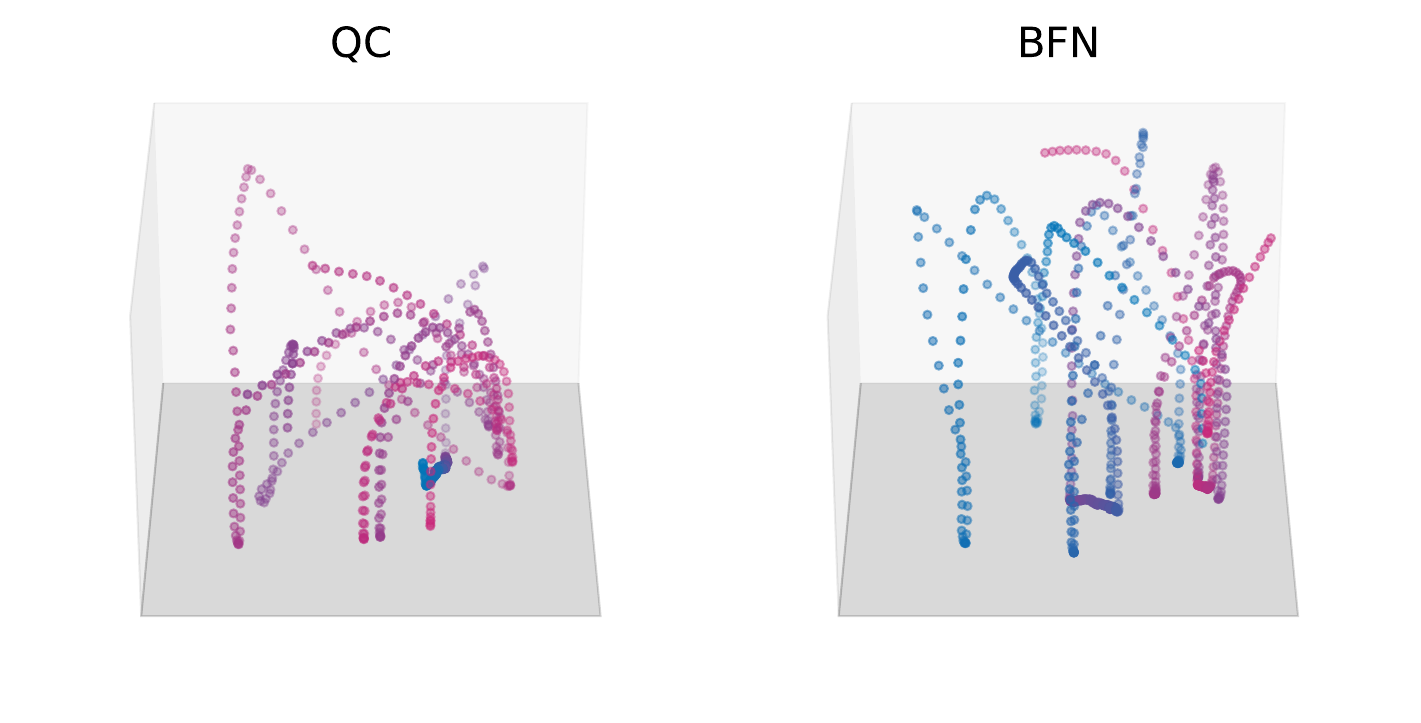}
\caption{\textbf{End-effector trajectory visualization late in the training.} Each subplot above shows the trajectory for a consecutive of 1000 time steps (from Step 900000 to Step 909000).}
\label{fig:traj-viz-late}
\end{figure}

We provide more examples of the trajectory rollouts from \hourpurple{QC} and \hourblue{BFN} over the course of online training on \texttt{cube-triple-task3}. In \cref{fig:traj-viz-early}, we show the first 9000 time steps (broken down into 9 subplots where each visualizes 1000 time steps). In \cref{fig:traj-viz-late}, we show another 9000 time steps but late in the training (from environment step $9\times10^5$). The first example is the same as the one used in \cref{fig:eef}.

\subsection{OGBench results by individual task}

\begin{figure}[H]
    \centering
\includegraphics[width=\linewidth]{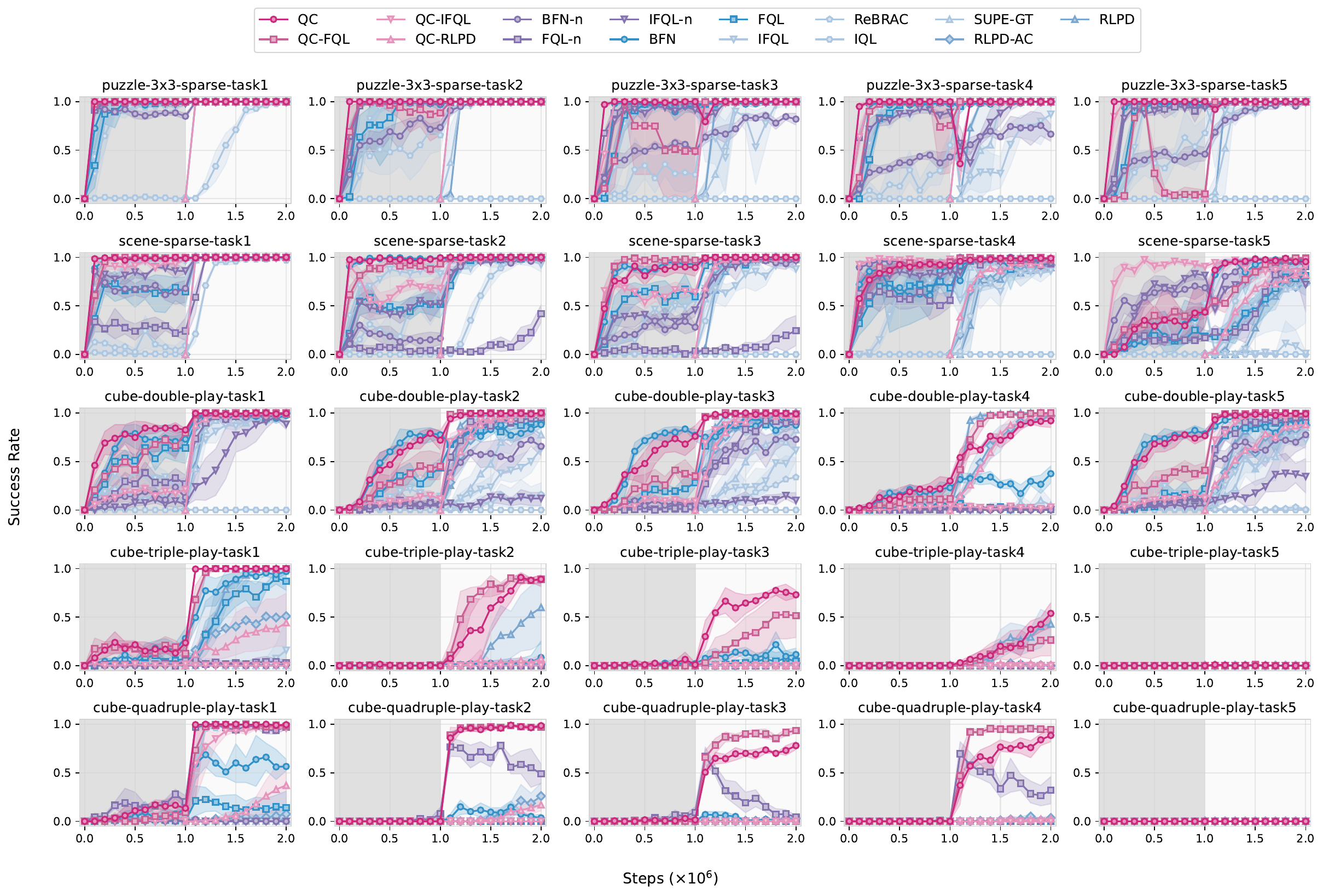} %
    \caption{\textbf{Complete OGBench offline-to-online RL results by task. 4 seeds. 95\% confidence interval.}}
    \label{fig:main-all-individual}
\end{figure}

\textbf{Main results by task.} \cref{tab:main-o2o} and \cref{fig:main-all-individual} shows the performance breakdown for all methods. 

\begin{table}[t]
    \centering

\makebox[\textwidth]{\scalebox{0.50}{
\begin{tabular}{cl>{\columncolor{ourlightblue}}c>{\columncolor{ourlightblue}}c>{\columncolor{ourlightblue}}c>{\columncolor{ourlightblue}}c>{\columncolor{ourlightblue}}c>{\columncolor{ourlightblue}}c>{\columncolor{ourlightblue}}c>{\columncolor{ourlightblue}}c>{\columncolor{ourlightmiddle}}c>{\columncolor{ourlightmiddle}}c>{\columncolor{ourlightmiddle}}c>{\columncolor{ourlightpurple}}c>{\columncolor{ourlightpurple}}c>{\columncolor{ourlightpurple}}c>{\columncolor{ourlightpurple}}c}
\toprule
 &  & \texttt{RLPD} & \texttt{RLPD-AC} & \texttt{SUPE-GT} & \texttt{ IQL} & \texttt{ ReBRAC} & \texttt{IFQL} & \texttt{FQL} & \texttt{ BFN} & \texttt{IFQL-n} & \texttt{FQL-n} & \texttt{BFN-n} & \texttt{QC-RLPD} & \texttt{QC-IFQL} & \texttt{\textbf{QC-FQL}} & \texttt{\textbf{QC}} \\
\bottomrule
\end{tabular}
}}
\vspace{1mm}
\caption{\footnotesize \textbf{Complete OGBench offline-to-online RL results.} 4 seeds. 95\% confidence interval.}
\label{tab:full-main-o2o}
\end{table}
 
\begin{figure}[t]
    \centering
\includegraphics[width=\linewidth]{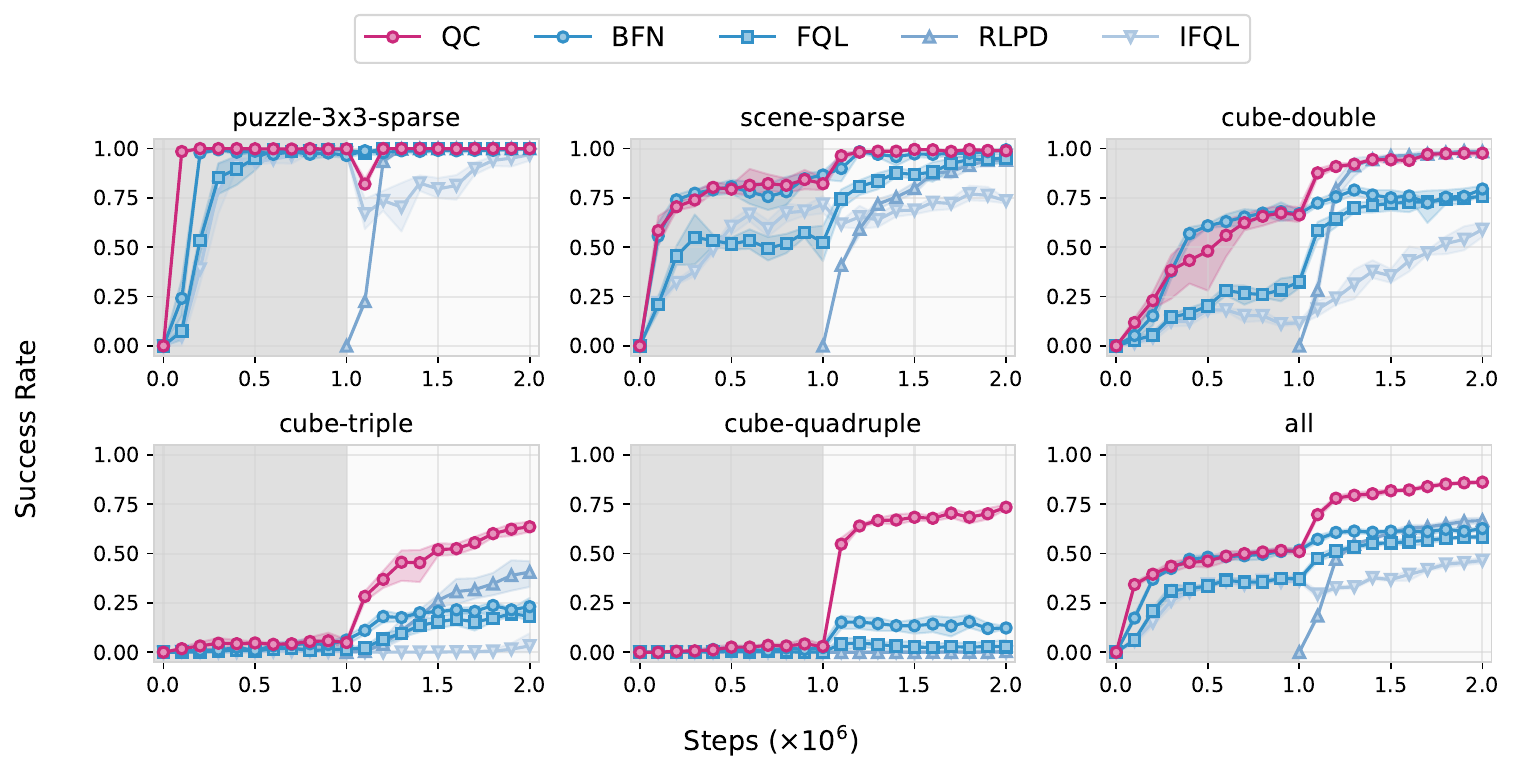}
\includegraphics[width=\linewidth]{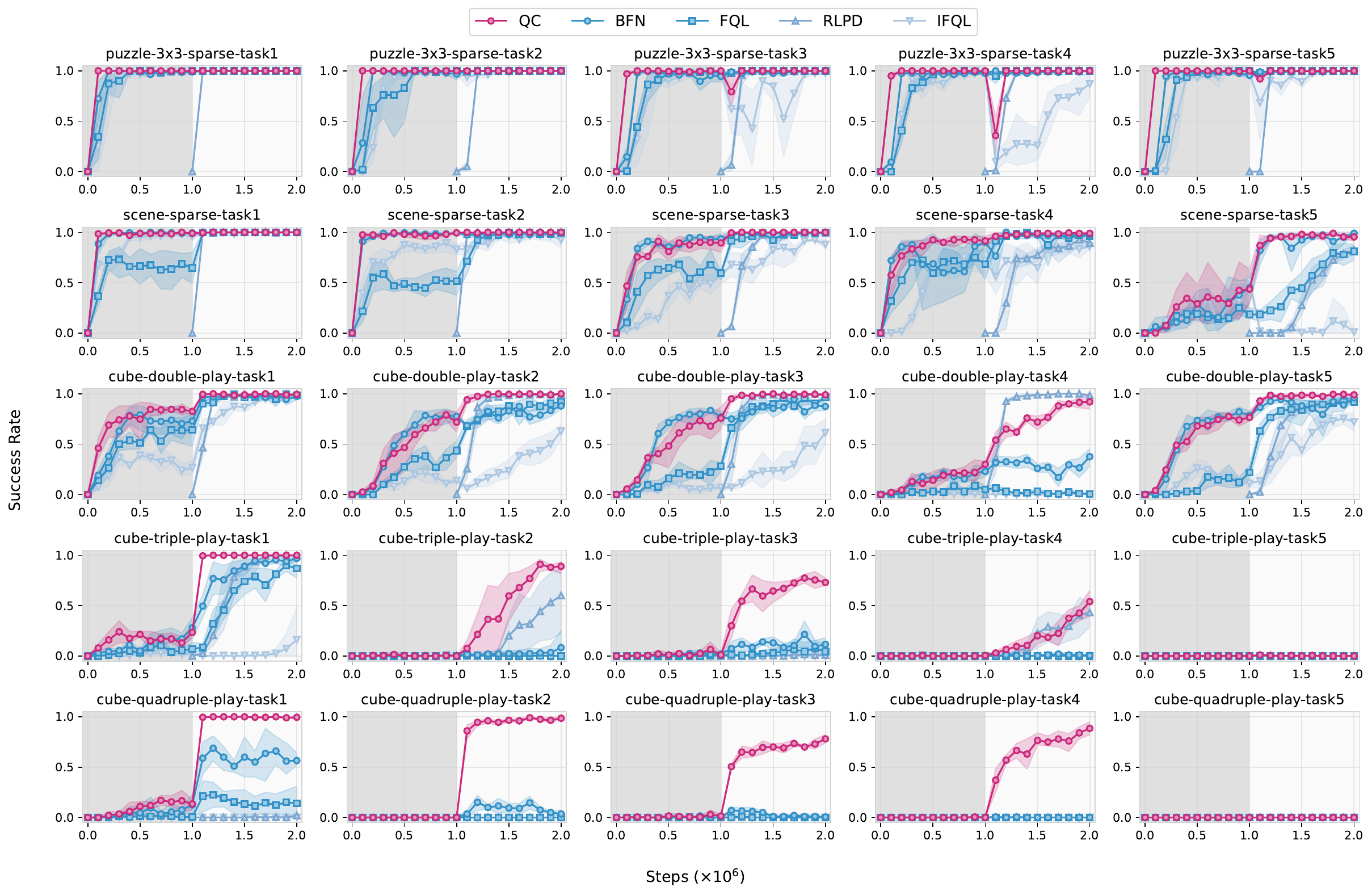} %
    \caption{\textbf{Full OGBench results by task (selected baselines).} \emph{Top:} summary plots by domain; \emph{Bottom:} individual plots by task.}
    \label{fig:main-all-individual-selected}
\end{figure}

\textbf{Selected baselines.} \cref{fig:main-all-individual-selected} shows the results for selected baselines (\emph{e.g.}, BFN, FQL, RLPD and RLPD-AC).

\begin{figure}[t]
    \centering
\includegraphics[width=\linewidth]{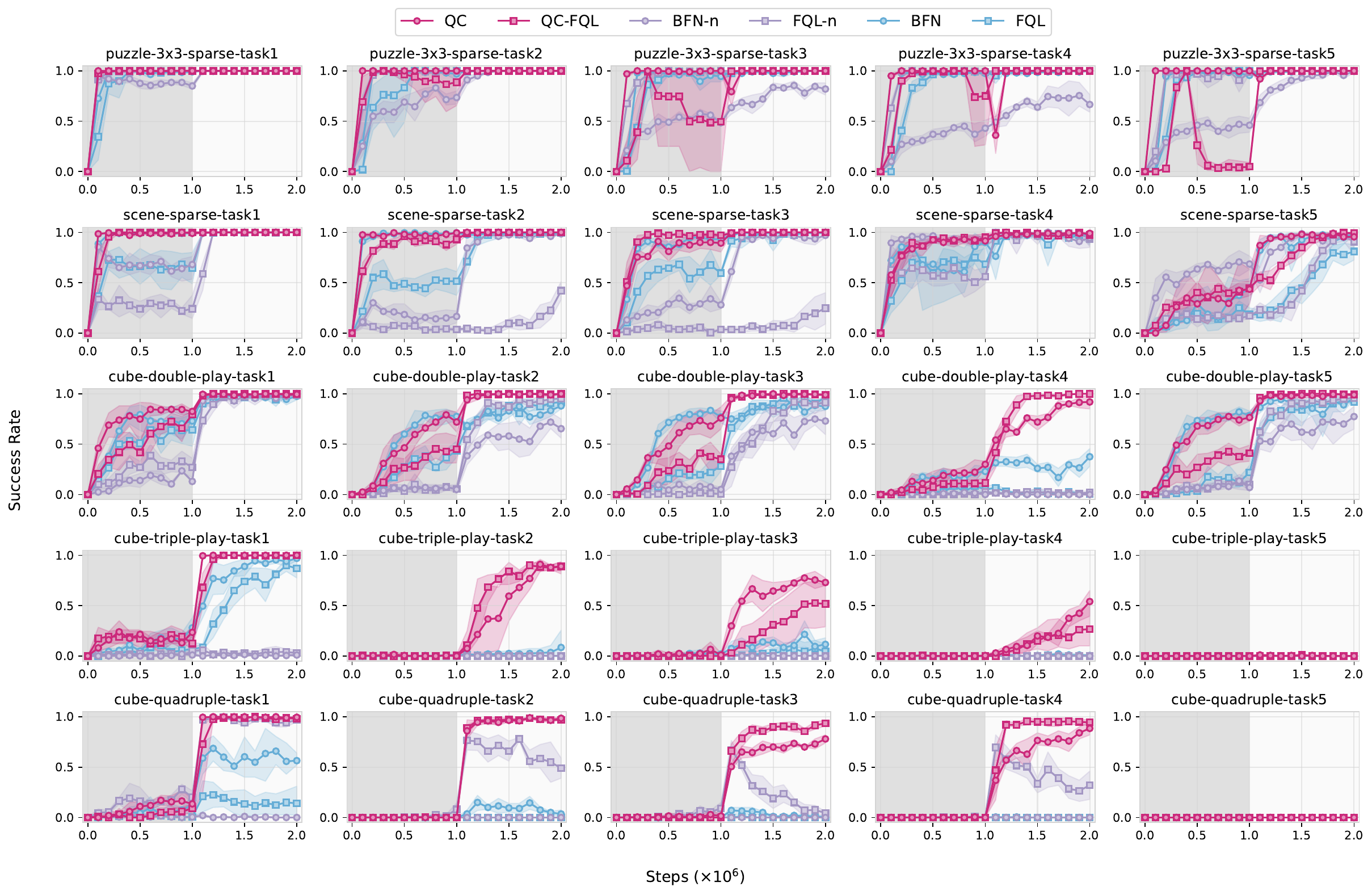} %
    \caption{\textbf{Full OGBench results by task (selected baselines).}}
    \label{fig:ablation-all-individual}
\end{figure}

\textbf{$n$-step return ablation results by task.} \cref{fig:ablation-all-individual} shows the performance breakdown for \cref{fig:ablation}.

\textbf{Q-chunking with Gaussian policies.}
The following plot shows the performance breakdown for \cref{fig:rlpd-all}. In addition, we include a new method for comparison, \hourpurple{QC-RLPD}, where we add a behavior cloning loss to \hourblue{RLPD-AC} (\hourblue{RLPD} with action chunking).
\begin{figure}[H]
    \centering
\includegraphics[width=\linewidth]{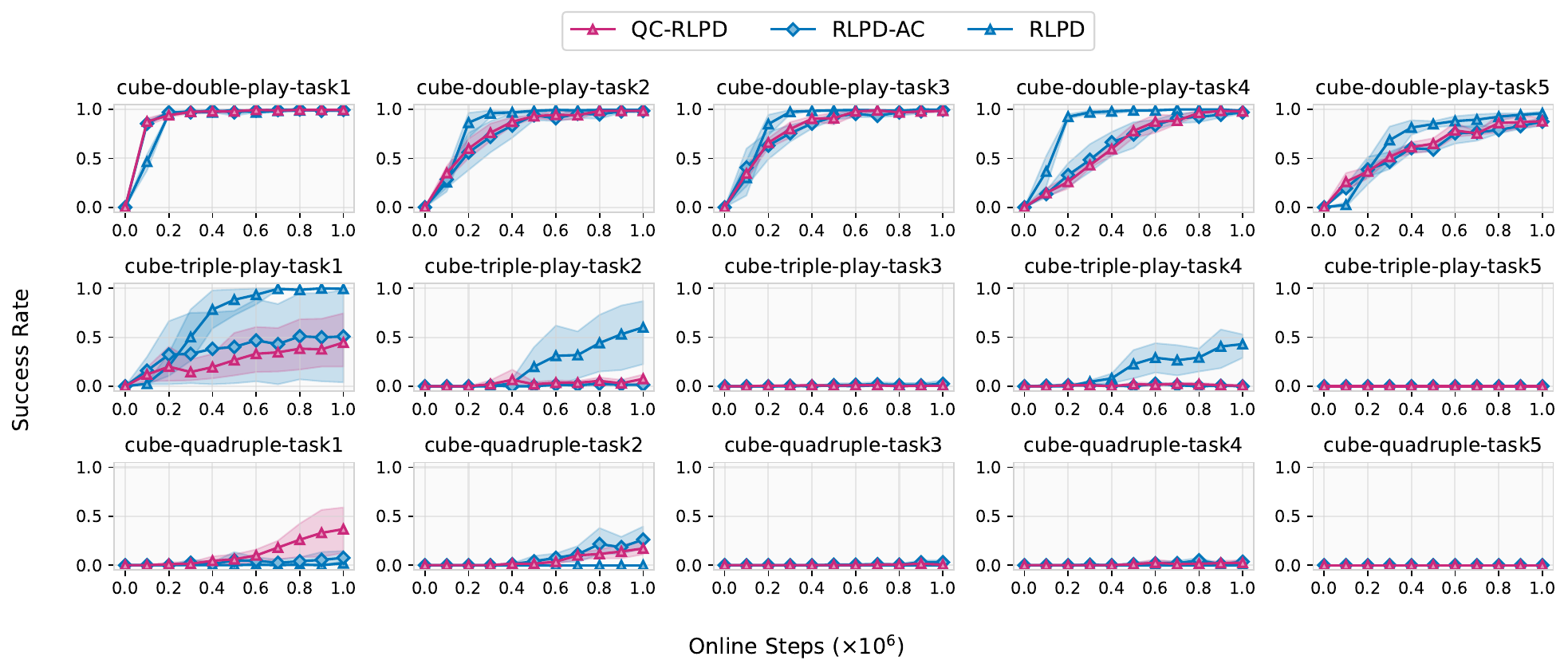} %
    \caption{\textbf{Full \hourblue{RLPD} results by task.} \hourpurple{QC-RLPD} is \hourblue{RLPD-AC} (\hourblue{RLPD} on the temporally extended action space) where we additionally add a fixed behavior cloning coefficient of $0.01$.}
    \label{fig:rlpd-all-ind}
\end{figure}

\begin{figure}[H]
    \centering
\includegraphics[width=\linewidth]{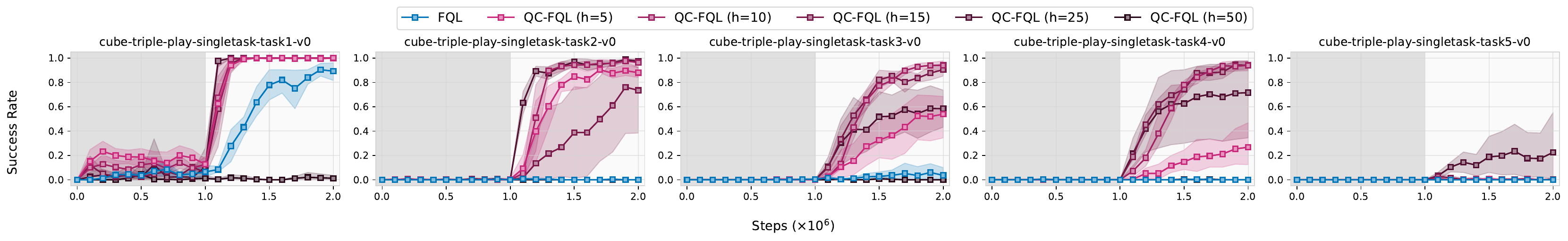} %
    \caption{\textbf{Action chunking size ablation on \texttt{cube-triple-*}.} Increasing the action chunking size generally helps until $h=25$. Surprisingly, $h=25$ is the only chunk size where \hourpurple{QC-FQL} achieves non-trivial success on the hardest \texttt{cube-triple-play-task5} (none of other methods can).}
    \label{fig:c3-ac-ablations}
\end{figure}

\begin{figure}[H]
    \centering
\includegraphics[width=\linewidth]{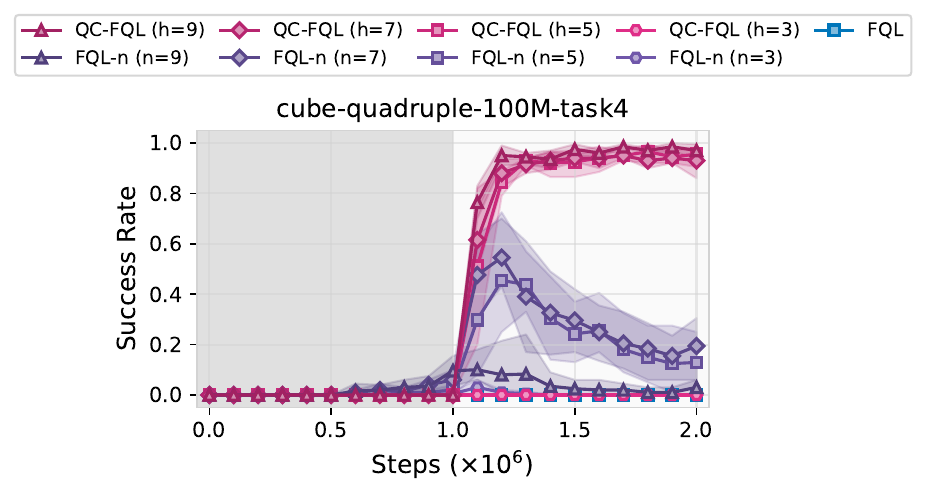} %
    \caption{\textbf{Action chunking size ablation on \texttt{cube-quadruple-play-task4}.} The $n$-step return baselines (\emph{i.e.}, \hourmiddle{FQL-n}) can achieve good initial success during online learning but quickly collapse. In contrast, our method (\emph{i.e.}, \hourpurple{QC-FQL}) does not suffer from such collapse and solves the task consistently across reasonably large chunk sizes (\emph{e.g.}, $h\in\{5, 7, 9\}$). The results are over $5$ seeds.}
    \label{fig:c4-ac-ablations}
\end{figure}

\textbf{Action chunking size ablations.}
\cref{fig:c3-ac-ablations} includes individual task results for the action chunking size ablation study. We also include results on the hardest \texttt{cube-quadruple} task in \cref{fig:c4-ac-ablations}.


\subsection{Robomimic ablation results}
\begin{figure}[H]
    \centering
\includegraphics[width=\linewidth]{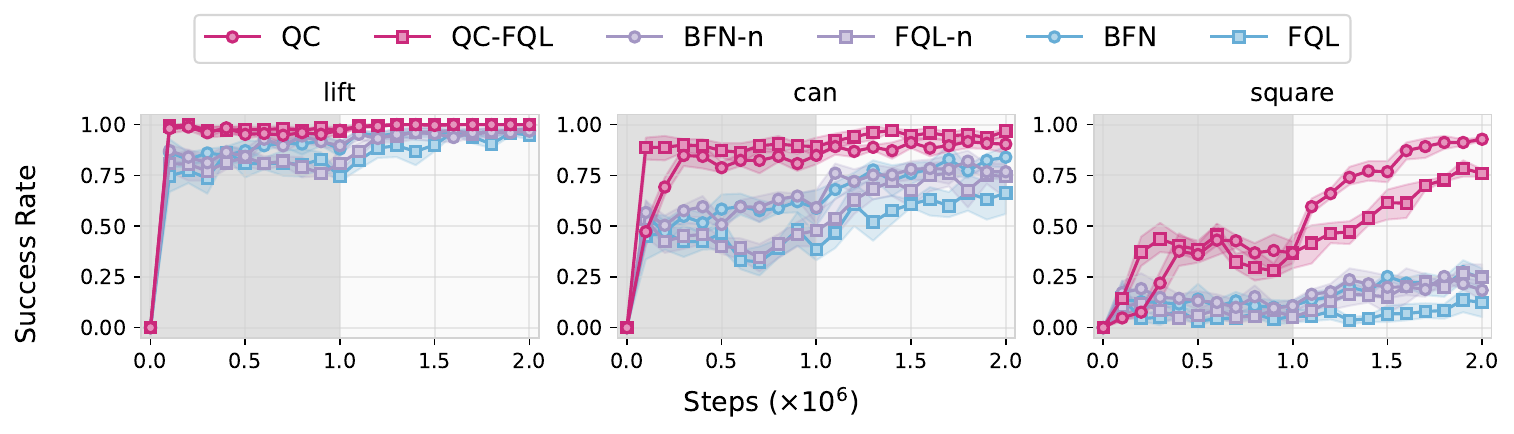}  %
    \caption{\textbf{Full robomimic ablation by task.} For each method on each task, we use 5 seeds.}
    \label{fig:full-robomimic}
\end{figure}

\cref{fig:full-robomimic} shows the performance of \hourpurple{QC}, \hourpurple{QC-FQL}, \hourmiddle{BFN-n}, \hourmiddle{FQL-n}, \hourblue{BFN}, \hourblue{FQL} our three robomimic tasks. This plot shows the performance breakdown for \cref{fig:ablation} (right). 

\subsection{How resource efficient is Q-chunking?}
\begin{figure}[H]
    \centering
\includegraphics[width=0.48\linewidth]{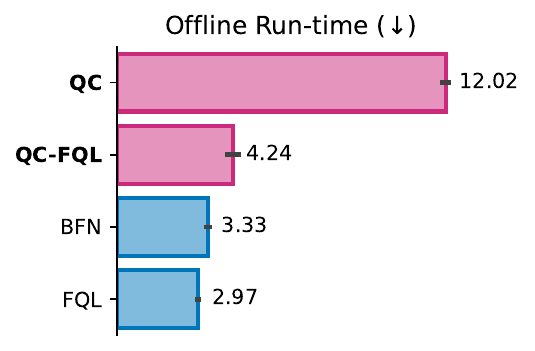}
\label{fig:run-time-offline}
\hfill
\includegraphics[width=0.48\linewidth]{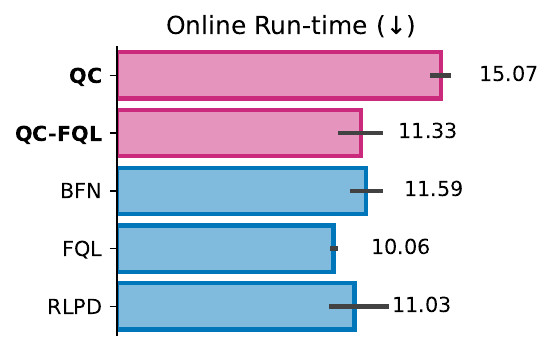}
\caption{\footnotesize \textbf{How long does each method take for one step in milliseconds.} Left: offline. Right: online (one agent training step and an environment step). The runtime is measured using the default hyperparameters in our paper on \texttt{cube-triple-task1} on a single RTX-A5000.}
\label{fig:run-time}
\end{figure}

In \cref{fig:run-time}, we report the runtime for our approach and our baselines on a representative task \texttt{cube-triple-task1}. In general, \hourpurple{QC-FQL} has a comparable run-time as our baselines (e.g., \hourblue{FQL} and \hourblue{RLPD}) for both offline and online. \hourpurple{QC} is slower for offline training as it requires sampling 32 actions for each training example for the agent update (\hourblue{BFN} is faster because it only needs to sample 4 actions). For online training, we are doing one gradient update per environment step, and it makes \hourpurple{QC} only around 50\% more expensive than other methods.

Finally, we include the parameter count of each of these method on the representative domain \texttt{cube-triple-*} in \cref{tab:pcnt} below (assuming $h=5$ for all Q-chunking methods).
\begin{table}[h]
    \centering
    \begin{tabular}{@{}cc@{}}
    \toprule
            Methods                        & Parameter Count (in millions) \\
    \midrule
    QC                             & $\approx 4.2$ \\
    BFN                            & $\approx 4.1$  \\
    QC-FQL                         & $\approx 5.0$ \\
    FQL                          & $\approx 4.9$  \\
    RLPD                          & $\approx 17.2$  \\
    \bottomrule
    \end{tabular}
    \vspace{2mm}
\caption{\footnotesize \textbf{Parameter count for each method.} RLPD has a much larger parameter count because it uses $K=10$ critic networks whereas all the other methods and baselines use only $K=2$.}
    \label{tab:pcnt}
\end{table}

\end{document}